%% file: acl_latex.tex
\documentclass[11pt]{article}

\usepackage[final]{acl}

\usepackage{times}
\usepackage{latexsym}

\usepackage[T1]{fontenc}

\usepackage[utf8]{inputenc}

\usepackage{microtype}

\usepackage{inconsolata}

\usepackage{graphicx}
\usepackage{pdfpages}

\usepackage{amsmath}
\usepackage{url}            
\usepackage{tabularx}
\usepackage{booktabs}
\usepackage{array}
\usepackage{amsfonts}       
\usepackage{nicefrac}       
\usepackage{microtype}      
\usepackage{xcolor}         
\definecolor{ForestGreen}{RGB}{34,139,34}
\definecolor{BrickRed}{RGB}{203,65,84}

\usepackage{multirow}
\usepackage{makecell}
\usepackage{adjustbox}
\usepackage{colortbl}
\usepackage{graphicx}
\usepackage{wrapfig}
\usepackage{capt-of}
\usepackage{placeins} 
\usepackage{listings} 
\usepackage{caption}
\usepackage{subcaption}
\usepackage{tocloft}

\usepackage{minted} 

\usepackage{enumitem}

\newcommand{\dUp}[1]{\scriptsize{\color{ForestGreen}$\uparrow$#1}}
\newcommand{\dDown}[1]{\scriptsize{\color{BrickRed}$\downarrow$#1}}

\newcommand{\fpDown}[1]{\,\scriptsize{\color{ForestGreen}$\downarrow$#1}}

\usepackage[normalem]{ulem}

\lstdefinestyle{neuripsPrompt}{
  basicstyle=\ttfamily\footnotesize,
  breaklines=true, breakatwhitespace=true, columns=fullflexible,
  keepspaces=true, showstringspaces=false, upquote=true,
  frame=liens, framerule=1pt, rulecolor=\color{black!20}, 
  xleftmargin=1ex, xrightmargin=1ex, aboveskip=0.5\baselineskip, belowskip=0.5\baselineskip
}

\lstdefinestyle{jsonExample}{
  language=json,
  numbers=none,
  numberstyle=\tiny\color{lstnum},
  stepnumber=1,
  numbersep=6pt,
  frame=none,
  backgroundcolor=\color{gray!3},
  xleftmargin=1ex, xrightmargin=1ex,
  aboveskip=0.5\baselineskip, belowskip=0.5\baselineskip
}

\newcounter{promptctr}
\newcounter{jsonctr}

\lstnewenvironment{Prompt}[1][]%
{%
  \stepcounter{promptctr}%
  \makeatletter

  \makeatother
  \lstset{style=neuripsPrompt,#1}
}%
{}

\lstnewenvironment{JSON}[1][]%
{%
  \stepcounter{jsonctr}%
  \makeatletter
  \makeatother
  \lstset{style=jsonExample,#1}%
}%
{}

\usepackage{etoolbox}
\apptocmd{\thebibliography}{\small}{}{}
\usepackage[table]{xcolor}

\definecolor{effsmall}{RGB}{255, 230, 153}   
\definecolor{effmedium}{RGB}{255, 183, 77}   
\definecolor{efflarge}{RGB}{255, 123, 45}   
\newcommand{\best}[1]{\textbf{#1}}
\newcommand{\second}[1]{\underline{#1}}
\newcommand{\sym}[1]{\ifmmode^{#1}\else\(^{#1}\)\fi}

\newcommand{\stat}[2]{%
  \ifstrequal{#2}{s}{\cellcolor{effsmall!35}}{}%
  \ifstrequal{#2}{m}{\cellcolor{effmedium!35}}{}%
  \ifstrequal{#2}{l}{\cellcolor{efflarge!35}}{}%
  #1$_{\textstyle #2}$%
}

\title{Do Vision--Language Models Understand Visual Persuasiveness?\\A Diagnosis via Visual Persuasive Factors}

\author{
  Gyuwon Park\textsuperscript{1}
  \and
  Hyounghun Kim\textsuperscript{1,2}
  \\
  \textsuperscript{1}Graduate School of Artificial Intelligence, POSTECH \\
  \textsuperscript{2}Department of Computer Science and Engineering, POSTECH \\
  \texttt{\{gyuwonpark,h.kim\}@postech.ac.kr}
}

\begin{document}
\maketitle
\addtocontents{toc}{\protect\setcounter{tocdepth}{-10}}

\begin{abstract}
Visual persuasion uses images to shape cognition, emotion, and behavior, with its effects depending on both visual attributes and semantic context.
Despite recent progress, it remains unclear whether Vision--Language Models (VLMs) understand visual persuasiveness.
This motivates us to ask: can VLMs assess whether an image persuasively supports an intended message, which visual factors shape this judgment, and do they align with human judgments?
Through empirical analyses on image-message pairs where human raters consistently agree on the persuasiveness judgment, we show that VLMs exhibit a recall-oriented bias: they over-predict images as persuasive while achieving high recall.
We introduce \textbf{Visual Persuasive Factors (VPFs)}, a taxonomy informed by cognitive psychology for quantifying visual cues that shape persuasive judgments.
Our factor-level analysis reveals that VPFs distinguish human persuasiveness judgments, whereas VLMs only partially reproduce these patterns---often generating false positives by treating persuasion-relevant cues as sufficient evidence.
Building on this insight, we evaluate VPF-guided interventions and find that properly framed VPF knowledge can improve performance, but merely specifying visual cues or adding step-by-step reasoning is insufficient.
By analyzing model rationales at the level of functional reasoning steps, we further identify a central bottleneck in connecting object identification to semantic message alignment.\footnote{Our code is available at \url{https://github.com/gyuwon12/visual-persuasive-factors}}

\end{abstract}

\input{latex/sections/01_introduction}
\input{latex/sections/02_related_work}

\input{latex/sections/03_VPFs}
\input{latex/sections/04_analysis_VPFs}
\input{latex/sections/05_rationale}
\input{latex/sections/06_conclusion}

\section*{Limitations}
The primary objective of this study is to establish a foundational diagnostic framework for assessing the interpretive capabilities of VLMs regarding visual persuasiveness through the proposed VPFs.
Accordingly, VPFs are proposed as a behavioral diagnostic framework for VLMs in the PVP-based setting, rather than as an exhaustive model of visual persuasion in general.
In light of this goal, we acknowledge several constraints in our current scope.
First, by focusing on high-agreement scenarios to limit variability in human judgments, our study did not explicitly model individual personality traits, which can significantly influence perceptions of persuasion.
A comprehensive understanding would require integrating these personal factors into the analysis. 
Future research will therefore aim to integrate the proposed VPF framework with user-centric modeling and psychographic profiles to investigate how VLMs can navigate the inherent subjectivity and individual variance in persuasive communication.
Second, our VPF analysis primarily examines factors at the level of individual perceptual, compositional, and semantic cues. 
However, visual persuasion can emerge from interactions among multiple factors, such as how object semantics, spatial composition, color, and message framing jointly shape judgment. 
Future work will extend VPFs beyond level-wise analysis toward modeling cross-factor interactions and combined cue configurations.

\section*{Ethical Considerations}
We acknowledge the dual-use concerns inherent in persuasion research. A deeper understanding of the factors driving visual persuasion could potentially be misused to create more effective disinformation or manipulative propaganda. However, we also believe this research has significant positive societal potential, such as in designing more effective public health and safety communications (\emph{e.g.}, ``always turn off the stove''). Our contribution is focused on the fundamental analysis of VLMs' cognitive alignment and its current limitations, not on building persuasive applications. We hope this research provides a foundation for developing VLMs that can better understand complex human cognitive processes in a responsible manner.

Furthermore, while we report effect sizes based on standard criteria for formal consistency~\cite{cohen1988}, we emphasize that these effects can have a much larger impact in real-world settings than the reported metrics suggest.
As argued in the field of psychology by \citet{effect_size_in_psychology}, effects typically categorized as ``small'' by conventional standards can be highly consequential when they operate repeatedly, affect large populations, or accumulate over time. 
In real-world visual persuasion settings, even small effects may become consequential when repeatedly deployed at scale. 
This motivates caution in interpreting and applying VPF-based analyses, especially in systems that could be used to generate or optimize persuasive visual content.

\section*{Acknowledgments}
We thank the reviewers and Action Editor for their constructive feedback and valuable comments. We also thank Daehyun Kim for his helpful support. This work was supported by Institute of Information \& Communications Technology Planning \& Evaluation (IITP) grant funded by the Korea government (MSIT) (No. RS-2019-II191906, Artificial Intelligence Graduate School Program (POSTECH)).

\bibliography{custom}

\clearpage
\appendix

\input{latex/sections/appendix}

\end{document}

%% file: latex/sections/01_introduction.tex
\section{Introduction}
Visual persuasion refers to the use of images and visual elements to influence cognition, emotion, and behavior, and it plays a central role in advertising, political communication, and social media \citep{chandler2011dictionary, visual_persuasion_cvpr, pvp}. 
Visuals are not just decorative—they reliably shape attitudes and behavior, as classic persuasion theory~\citep{ELM} and public-health evidence on pictorial warnings have shown~\cite{cigarette, brewer2016effect}.
Visual elements are therefore central to persuasion, but their effects unfold through their interaction with semantic context, persuasion strategies, and individual psychological dispositions~\citep{PersuasionStrategiesAds, pvp}.
However, existing work provides limited insight into how well VLMs actually understand visual persuasiveness in this broader context---why certain images are persuasive, which visual factors contribute to their effects, and how these factors differ from those underlying human judgments remain unclear.

\begin{figure*}[t]
    \centering
    \setlength{\abovecaptionskip}{5pt}
    \setlength{\belowcaptionskip}{0pt}
    \includegraphics[width=\textwidth, trim=0 4 0 2, clip]{figures/Main_figure.pdf}
    \caption{
    (Left) VLM persuasiveness judgments are evaluated against high-consensus human labels, revealing a recall-oriented bias.
    (Right) Visual Persuasive Factors span perceptual, compositional, and semantic cues, enabling factor-level analysis of visual persuasion.
    This setting and VPF construction are detailed in \S\ref{sec:vlm_eval}--\S\ref{sec:human_survey}.
    }
    \label{fig:vpf_overview}
    \vspace{-1.4em}
\end{figure*}

In this work, we address this gap by introducing a framework for examining the role of persuasion-relevant visual factors and comparing their associations with VLM and human judgments.
To isolate the effect of visual factors, we control for the inherent subjectivity of persuasiveness judgments~\citep{nature2024_persuasive_ai}.
Specifically, we use the Personalized Visual Persuasion (PVP) dataset~\citep{pvp}, which provides image-message pairs with multiple human persuasiveness ratings, and retain 562 clearly separable persuasive (High) and non-persuasive (Low) binary cases with almost-perfect agreement~\citep{kappa_score}.
We then evaluate whether VLMs can assess how persuasively an image supports a given target message, as shown in Fig.~\ref{fig:vpf_overview} (Left).
Surprisingly, we find that even in these seemingly straightforward scenarios, VLMs struggle to exhibit genuine understanding of visual persuasiveness; they tend to over-predict persuasiveness, revealing a pronounced \textit{recall-oriented bias}. 
Together with insights from the Elaboration Likelihood Model~\citep{ELM}, we hypothesize that VLM limitations stem from how models identify, weight, and contextualize visual and semantic cues that contribute to persuasiveness.

To make this hypothesis operational, we propose \textbf{Visual Persuasive Factors (VPFs)}, a novel structured taxonomy and behavioral diagnostic framework informed by cognitive psychology~\cite{Cross_Cultural_color, layout1, keyobj} for measuring cues that shape persuasiveness judgments in image-message pairs, as illustrated in Fig.~\ref{fig:vpf_overview} (Right).
VPFs span three levels---perceptual, compositional, and semantic---covering color and brightness; saliency and message-relevant region placement with respect to the center and rule of thirds; and the presence of key object, human, and text.
A validation survey with 141 annotators, who are distinct from the original PVP raters, provides evidence for the validity of these factors.
Using VPFs, we compare human ground-truth persuasion labels with VLM predictions, revealing three key insights:
\begin{itemize}[leftmargin=13pt, itemsep=0pt, topsep=2pt, parsep=0pt]
    \item VPFs differentiate high- and low-persuasiveness cases based on interpretable visual and semantic dimensions.
    \item VLMs only partially reproduce human-aligned VPF patterns, capturing some persuasive signals while showing inconsistent weighting across models and distorted associations with factors such as human and text presence.
    \item VLM false positives arise from overgeneralizing persuasion-relevant cues: among ground-truth Low items, models often predict High when images contain plausible persuasive cues---\emph{e.g.}, key object presence---mistaking them for sufficient evidence of persuasiveness.
\end{itemize}

Leveraging these insights, we further examine whether VPF-guided interventions can reduce the human--VLM gap in persuasive judgment.
Focusing on key object presence as a semantic cue that reliably separates human judgments yet also contributes to VLM false positives, we test whether knowledge injection and reasoning scaffolds can help VLMs use persuasion-relevant cues more appropriately.
We find that VLMs can improve when such cues are explicitly framed as auxiliary cues to be interpreted in context rather than as fixed evidence for persuasiveness.
However, naive cue provision or explicit reasoning~\cite{cot} is not sufficient for robust improvement: direct grounding often degrades performance, and reasoning with explicit cues can even amplify the recall-oriented bias, suggesting a failure to reliably connect visual cues to the final persuasive judgment.
To better understand this limitation, we extract and decompose model-generated rationales into three functional steps---(i) object identification, (ii) semantic message support, and (iii) final judgment connection---and ablate each step to examine how it affects model predictions.
This step-wise ablation shows that the difficulty lies in linking object-level and semantic evidence to communicative intent.

Our contributions are threefold:
(1) we propose Visual Persuasive Factors (VPFs), a taxonomy informed by cognitive psychology, as a structured framework for visual persuasive understanding; 
(2) we use VPFs as a cognitive factor lens to compare human-labeled and VLM-predicted persuasiveness, showing that VPFs distinguish human judgments while VLMs only partially reproduce these human-aligned patterns and overgeneralize persuasion-relevant cues; and
(3) we conduct VPF-guided intervention and reasoning analyses, revealing that semantic grounding alone does not close the gap and that the key bottleneck lies in connecting visual evidence to communicative intent.

%% file: latex/sections/02_related_work.tex
\section{Related Work}
\subsection{AI-based Persuasion}
Research on computational persuasion began in text-only dialogue settings, with corpora such as Persuasion for Good~\citep{persuasion4good} enabling strategy-based persuasive response generation; later work incorporated affect, user traits, and LLM-generated messages to study empathy, personalization, and human judgments of persuasiveness~\citep{naacl2022_emotion_persuasion,nature2024_persuasive_ai,nature2025_personalized_persuasion}.
In visual persuasion, 
early work used syntactic visual cues to infer communicative intents in persuasive images~\citep{visual_persuasion_cvpr}, while recent datasets such as PVP pair 28,454 persuasive images with 596 messages annotated for nine persuasion strategies~\citep{pvp}.
While these studies establish visual persuasion as a computational task, they leave open how VLM judgments relate to the visual factors that humans use to assess persuasiveness.
Recent evaluation protocols measure persuasiveness more directly, including human-in-the-loop and controlled prompting setups for assessing model persuasion and safety implications~\citep{anthropic_persuasion, arxiv2024_persuasion_eval_2410_02653}. 
Our study extends this paradigm to the visual domain by using factor-level analysis to compare model and human decisions.

\subsection{Cognitive Foundations for Visual Persuasive Factors}
\label{sec:rel_cognitive_}
Our taxonomy draws on cognitive psychology and communication theory. 
The Elaboration Likelihood Model~\citep{ELM} posits two paths to persuasion: a central route operating via message-relevant elaboration, and a peripheral route relying on immediate or heuristic cues.
The VPFs can operate through either route, depending on how it is processed.

Numerous behavioral studies demonstrate that color is a fundamental component of visual perception and persuasion~\cite{Cross_Cultural_color, role_of_color, cognitive_color, red_and_blue}, and is especially influential in advertisements and campaigns~\cite{Marketers_color, color_health, Red_Hot_Reactance}.
Beyond color, composition and visual hierarchy guide attention, as shown by center bias in natural scenes~\citep{centerbias2, centerbias1} and the rule-of-thirds in photographic composition~\citep{rule_of_thrid_definition,rule_of_third_for_photography}.
Advertising and eye-tracking studies further show that visual layout shapes gaze allocation and brand responses, highlighting composition as an attention-guiding factor in persuasive media~\citep{layout1, layout2}.
At the semantic level, persuasiveness depends on whether image objects align with the intended argument and form a coherent narrative, requiring models to select relevant objects and connect them to an argument structure~\cite{selective}. 
Such core objects may include products~\cite{keyobj}, text~\cite{image_text_2, image_text}, or people~\cite{human_face, human_presence}.

\subsection{High-Level Vision--Language Reasoning}
Recent foundation models~\cite{openai2025gpt5, comanici2025gemini, deepseek} have advanced rapidly on complex tasks, while chain-of-thought prompting improves reasoning in arithmetic, commonsense, and symbolic settings~\cite{cot, wang2023selfconsistency, yasunaga2024analogical}. 
VLMs extend this paradigm by coupling LLMs with visual encoders for multimodal inputs~\cite{liu2023llava, liu2024llavanext, bai2025qwen25vltechnicalreport, gemmateam2025gemma3technicalreport}.
Recent VLM research has moved beyond object recognition toward abstract and symbolic reasoning, including visual metaphor understanding, where models must bridge perception and semantics~\citep{metaclue, metamorph}.

%% file: latex/sections/03_VPFs.tex
\section{From VLM Persuasive Understanding to Visual Persuasive Factors}
The first goal of our study is to examine whether VLMs possess a meaningful notion of visual persuasive understanding.
In the following sections, we establish the evaluation setting using a high-agreement subset (\S\ref{sec:vlm_eval}), introduce our novel taxonomy Visual Persuasive Factors (\S\ref{sec:vpf_definition}), describe how factors are extracted and measured (\S\ref{sec:vpf_extraction}), and validate them through an independent survey (\S\ref{sec:human_survey}).

\subsection{Evaluating VLMs' Understanding of Visual Persuasiveness}
\label{sec:vlm_eval}
To assess whether VLMs understand visual persuasion, we derive a high-consensus subset from the PVP dataset~\cite{pvp}. 
We bin the original 0--10 persuasiveness ratings into Low (0--2), Mid (3--7), and High (8--10) bands, and retain 562 image-message pairs with four-annotator agreement in the Low or High band, \emph{i.e.}, \textit{almost-perfect} agreement under~\citet{kappa_score}. 
We then frame persuasion as a binary judgment task (image $+$ message $\rightarrow$ \{High, Low\} treated as ground truth; details in Appendix~\ref{app:dataset_details}).
Within this high-agreement subset, we find limited evidence that VPF effects are broadly moderated by personality.\footnote{PVP provides 20 viewer-level psychological dimensions: 5 Big Five traits~\citep{big5}, 10 Schwartz values~\citep{schwartz2012refining}, and 5 moral foundations~\citep{moral}. We test their interactions with 11 VPF measures ($20 \times 11=220$ tests); after FDR correction, only one interaction \emph{Tradition}$\times$T3 saliency remains significant.}
Hence, this subset comprises persuasive and non-persuasive cases that are relatively clear to viewers
Yet VLMs~\cite{openai2025gpt5, gemmateam2025gemma3technicalreport, bai2025qwen25vltechnicalreport} consistently show poor performance and a recall-oriented bias: 
they predict High too often, achieving near-perfect recall with inflated false positives, as shown in Tab.~\ref{tab:main_result} baseline.\footnote{Although we evaluate LLaVA variants~\cite{liu2023llava, liu2024llavanext}, we exclude them from the main analysis due to significant instruction-following failures and poor output stability.}

\subsection{Defining Visual Persuasive Factors}
\label{sec:vpf_definition}
Building on prior findings in \S\ref{sec:rel_cognitive_}, we define Visual Persuasive Factors as a three-level decomposition of persuasive content in image-message pairs. 

\begin{itemize}
[leftmargin=*, itemsep=0pt, topsep=2pt, parsep=0pt]
    \item \textbf{Perceptual} factors capture immediate visual impressions, including colorfulness, color entropy, representative color congruence, and brightness.
    \item \textbf{Compositional} factors describe the spatial organization of persuasive cues, including how visual saliency and message-relevant regions are distributed with respect to center bias and rule-of-thirds structure.
    \item \textbf{Semantic} factors capture various objects, including whether a key object, human, or text is present and whether they provide evidence that modulates the intended persuasive appeal.
\end{itemize}

Our VPFs are not intended to be exhaustive; they organize visually measurable cues into three diagnostic levels for VLM evaluation.

\subsection{VPF Feature Extraction}
\label{sec:vpf_extraction}
\paragraph{Perceptual Level.}
We characterize perceptual factors in two complementary ways. First, we extract image-intrinsic properties that can be directly quantified, including \emph{colorfulness, color entropy, and brightness}. We use the \citet{colofulness_hasler} colorfulness metric to measure color intensity, compute color entropy in CIELAB space to capture color diversity, and use the $L^*$ channel to quantify brightness~\cite{CIELAB}.

Second, we measure whether perceptual-level properties align with semantic framing.
The intuition is that persuasiveness increases when cues implied by message intent are reflected in the image's visual attributes (\emph{e.g.}, green color for environmental messages).
For color, we first define a discrete eight-color affective palette inspired by~\citet{Cross_Cultural_color} and then extract a representative image color via K-means clustering and map it to one palette category.
Separately, given the message and persuasion premise, GPT-5 infers the top-3 expected colors from the same palette.\footnote{A premise is a short textual cue specifying the persuasion strategy for each image variation in PVP's one-message-to-many-images structure. An example is in Fig.~\ref{fig:vpf_overview}.} 
Representative Color Congruence is the hit@3 overlap between the image's representative color and the inferred top-3 expected colors.

Similarly, for brightness, we score whether image brightness matches the implied persuasion strategy (\emph{i.e.}, premise) by aligning $L^*$ values with strategy polarity (positive, negative, or neutral), following psychological patterns such as brighter images for positive strategies and darker images for negative ones~\citep{Cross_Cultural_color}. 
In summary, we report normalized values for three color metrics (colorfulness, entropy, and representative color congruence), alongside raw brightness ($L^*$) and its strategy-specific alignment scores for positive- and negative-strategy subsets.\footnote{Since there were too few neutral cases, we only report the metrics for positive and negative cases.}
Full metric details and examples are in the Appendix~\ref{app:vpf_low}.

\paragraph{Compositional Level.}
To measure compositional factors, we focus on two canonical layout principles: the \emph{Center Bias}~\citep{centerbias1, centerbias2} and the \emph{Rule of Thirds} (RoT)~\citep{rule_of_thrid_definition}. 
We evaluate them using two representations: log-probability saliency maps from DeepGaze IIE~\cite{DeepGazeIIE}, which capture visually attended regions, and Message-Relevant Region (MRR), defined as Grounded-SAM-2~\citep{grounded-sam} segmentation masks obtained from open-vocabulary queries derived from the message. 
For both representations, we compute Center Bias Index (CBI) and Thirds Hotspot Coverage (T3), which measure the proportion of total saliency mass or mask area contained in predefined circular regions around the image center and the four rule-of-thirds intersections, respectively.
Higher values indicate stronger spatial concentration in those regions. Full metric details and examples are in Appendix~\ref{app:vpf_mid}.

For perceptual and compositional factors, we compare each feature \(f\) between persuasive and non-persuasive groups using
\begingroup
\setlength{\abovedisplayskip}{4pt}
\setlength{\belowdisplayskip}{4pt}
\setlength{\abovedisplayshortskip}{2pt}
\setlength{\belowdisplayshortskip}{4pt}
\begin{equation}
\Delta_f^{(Z)} =
\mathbb{E}[f \mid Z=\mathrm{High}]
-
\mathbb{E}[f \mid Z=\mathrm{Low}],
\label{eq:factor_delta}
\end{equation}
\endgroup
where \(Z \in \{Y, \hat{Y}\}\) denotes the ground-truth or the predicted label, respectively.

\paragraph{Semantic Level.}
We extract three factors from image content: \emph{Key Object Presence}, \emph{Human Presence}, and \emph{Text Presence}.
We operationalize all three factors in a simple binary form, focusing on whether each cue is present in the image. 
Key objects are extracted by identifying candidate nouns from the message and verifying whether their corresponding visual instances are depicted in the image; humans are detected using standard object detection; and text is manually verified for readable elements such as slogans or call-to-action phrases.
In our curated dataset, the presence rates of these semantic factors are 56.58\% for key objects, 71.00\% for humans, and 22.06\% for text.
To evaluate their association, we fit logistic regressions with each semantic factor as a binary presence indicator and either ground-truth labels or model predictions as the outcome, reporting odds ratios (ORs).
Full details of the pipeline are in Appendix~\ref{app:high_level_details}.

\subsection{Validation Survey}
\label{sec:human_survey}
We conduct an independent validation survey to assess the proposed VPFs by directly measuring their perceived influence on visual persuasive understanding. 
Annotators were recruited via Amazon Mechanical Turk.\footnote{https://www.mturk.com}
They were first given concise definitions of all factors, then shown image-message pairs and asked to rate the influence of each factor using a 5-point Likert scale (1: not at all influential, 5: extremely influential).
Each item was annotated by three annotators, with strict quality controls to remove low-quality responses.

\input{latex/sections/vpfs_main_table}

\paragraph{Results.}
If VPFs capture meaningful dimensions of persuasive evidence, they should be perceived as more relevant in High items, where visual evidence more effectively supports the target message.
Consistent with this expectation, High items receive significantly higher factor-importance ratings than Low items across all nine factors (rank-based AUC = 0.649--0.805, all $q < 10^{-7}$), supporting the construct validity of the VPF taxonomy. 
Because the validation survey and VLM evaluation differ in task format and available context, we avoid direct one-to-one comparisons between survey ratings and VLM predictions. 
Full details are in Appendix~\ref{app:human_survey_details}.

\begin{figure}[t]
    \centering
    \setlength{\abovecaptionskip}{3pt}
    \setlength{\belowcaptionskip}{0pt}
    \includegraphics[width=\linewidth, trim=0 4 0 2, clip]{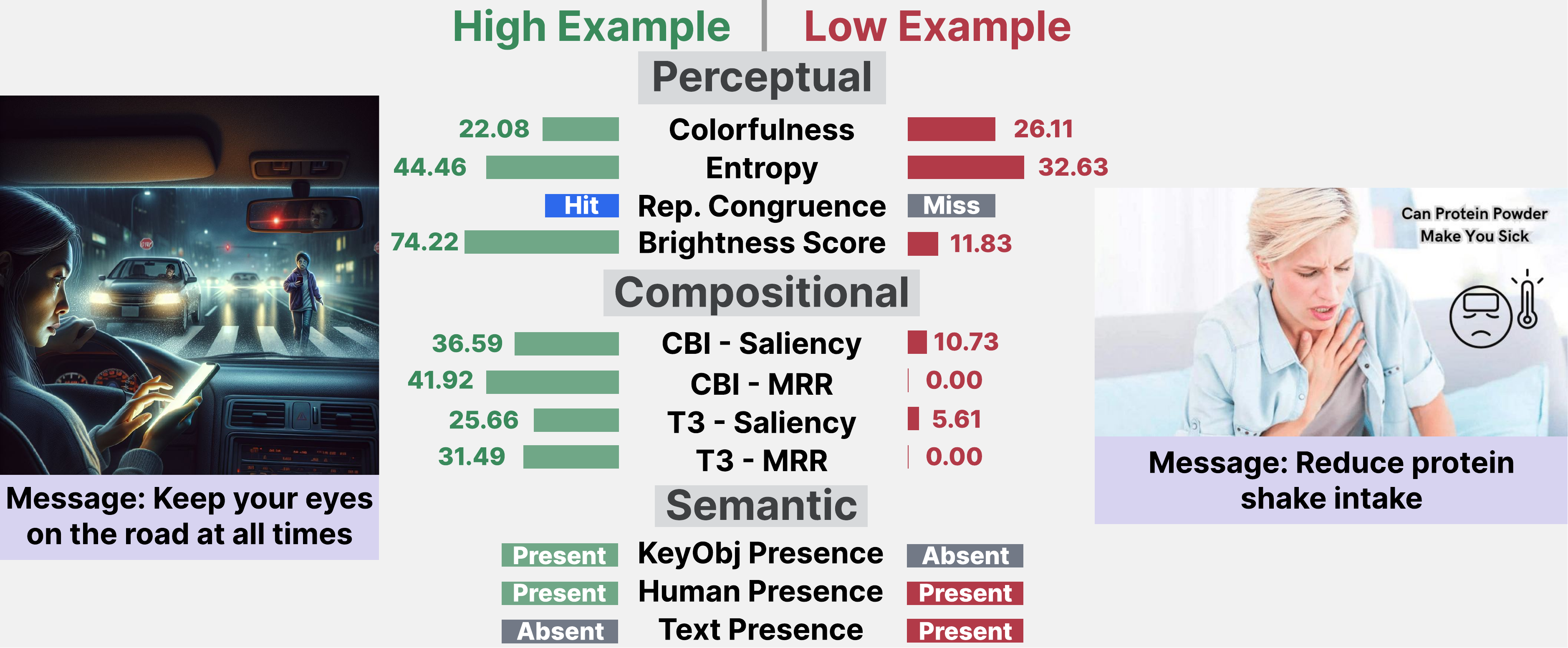}
    \caption{Qualitative comparison of VPF patterns between High- and Low-persuasiveness cases.}
    \label{fig:vpf_qualitative_comparison}
    \vspace{-1.4em}
\end{figure}

%% file: latex/sections/vpfs_main_table.tex
\begin{table*}[t]
\centering
\renewcommand{\arraystretch}{1.12}
\setlength{\tabcolsep}{3pt}

\setlength{\abovecaptionskip}{5pt}
\setlength{\belowcaptionskip}{0pt}
\begin{adjustbox}{max width=\linewidth}
\begin{tabular}{l l ccc ccc cc cc ccc}
\toprule
\multirow{3}{*}{\textbf{Group}} 
& \multirow{3}{*}{\textbf{Row}}
& \multicolumn{6}{c}{\textbf{Perceptual (\%)}}
& \multicolumn{4}{c}{\textbf{Compositional (\%)}}
& \multicolumn{3}{c}{\textbf{Semantic (OR)}} \\
\cmidrule(lr){3-8}\cmidrule(lr){9-12}\cmidrule(lr){13-15}

&
& \multicolumn{3}{c}{Color}
& \multicolumn{3}{c}{Brightness}
& \multicolumn{2}{c}{CBI}
& \multicolumn{2}{c}{T3}
& \multirow{2}{*}{KeyObj}
& \multirow{2}{*}{Human}
& \multirow{2}{*}{Text} \\
\cmidrule(lr){3-5}\cmidrule(lr){6-8}\cmidrule(lr){9-10}\cmidrule(lr){11-12}

&
& Colorfulness & Entropy & \makecell{Rep.\\Congruence}
& Raw & Pos & Neg
& Saliency & MRR
& Saliency & MRR
& & & \\
\midrule

\rowcolor{gray!15!violet!10}\multicolumn{15}{l}{\textbf{Panel A: Ground Truth}} \\
\addlinespace[2pt]
\raisebox{-0.7ex}{\includegraphics[width=1.3em]{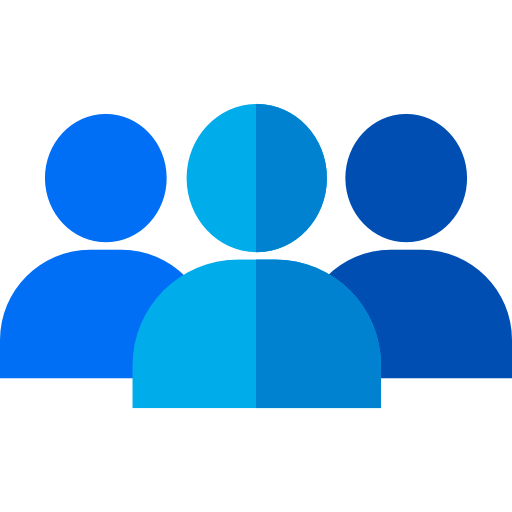}}\ Human
& $\Delta_f^{(Y)}$
& -3.94\stat{\sym{**}}{s}
&  4.57\stat{\sym{***}}{s}
& 15.09\stat{\sym{***}}{s}
& -0.34\stat{}{n}
& -5.11\stat{}{s}
& 5.91\stat{\sym{**}}{s}
& 2.65\stat{\sym{**}}{s}
& -2.83\stat{}{n}
& 3.19\stat{\sym{***}}{s}
& 3.31\stat{\sym{**}}{s}
& 3.36\stat{\sym{***}}{m}
& 1.25\stat{}{n}
& 0.60\stat{\sym{*}}{s} \\
\addlinespace[2pt]

\rowcolor{gray!15!violet!10}\multicolumn{15}{l}{\textbf{Panel B: VLM Predictions}} \\
\addlinespace[2pt]
\raisebox{-0.6ex}{\includegraphics[width=1.3em]{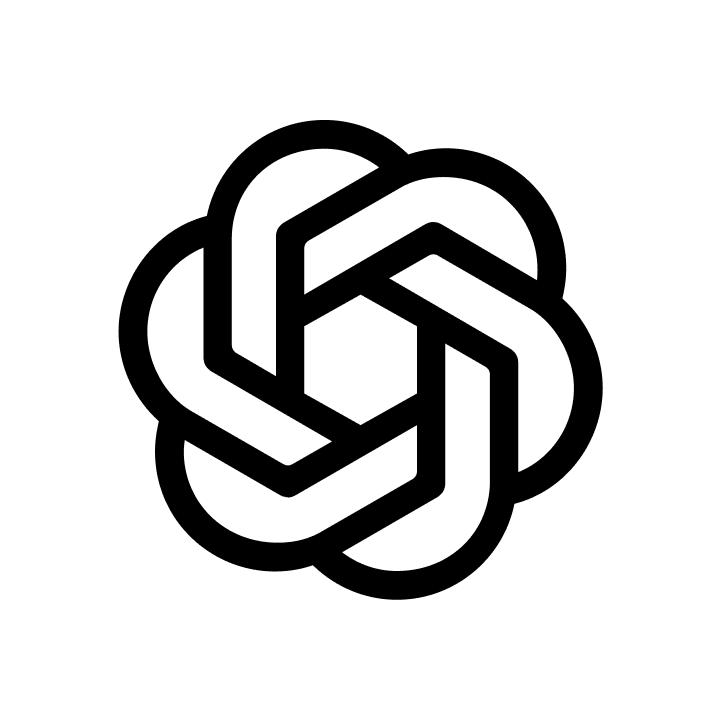}}\ GPT-5
& $\Delta_f^{(\hat{Y})}$
& -1.66\stat{}{n}
& 4.82\stat{\sym{***}}{s}
& 12.34\stat{\sym{***}}{s}
& -1.69\stat{}{n}
& -5.74\stat{\sym{*}}{s}
& 4.19\stat{\sym{*}}{s}
& 3.27\stat{\sym{***}}{s}
& 2.45\stat{}{n}
& 2.58\stat{\sym{***}}{s}
& 1.23\stat{}{n}
& 2.87\stat{\sym{***}}{m}
& 1.56\stat{\sym{*}}{s}
& 0.94\stat{}{n} \\
\addlinespace[2pt]

\raisebox{-0.6ex}{\includegraphics[width=1.3em]{icons/gpt.png}}\ GPT-5-mini
& $\Delta_f^{(\hat{Y})}$
& -3.51\stat{\sym{*}}{s}
& -3.68\stat{\sym{**}}{s}
& 16.19\stat{\sym{***}}{s}
& -1.84\stat{}{n}
& -6.92\stat{\sym{*}}{s}
& 6.67\stat{\sym{**}}{s}
& 2.90\stat{\sym{***}}{s}
& 1.46\stat{}{n}
& 2.65\stat{\sym{***}}{s}
& 0.67\stat{}{n}
& 2.68\stat{\sym{***}}{m}
& 1.58\stat{\sym{*}}{s}
& 1.35\stat{}{n} \\
\addlinespace[2pt]

\raisebox{-0.6ex}{\includegraphics[width=1.3em]{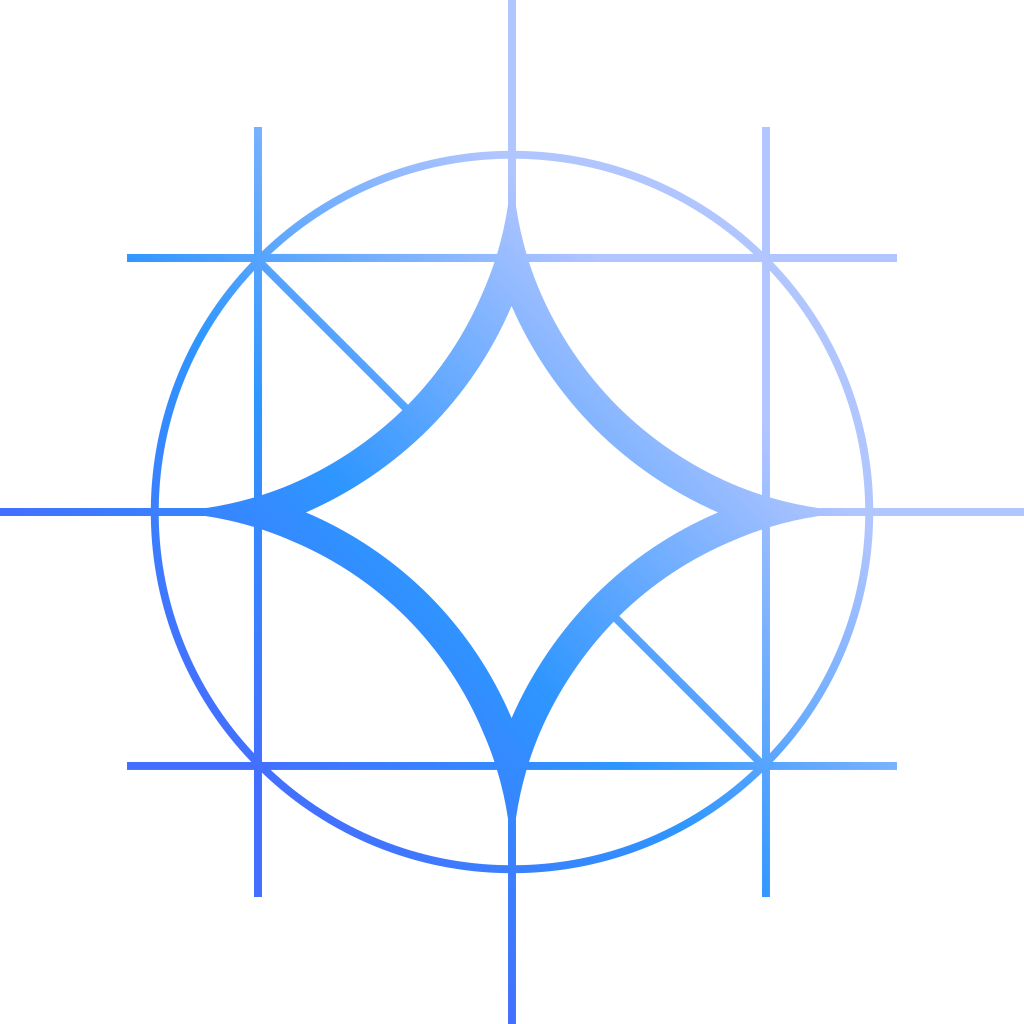}}\ Gemma3
& $\Delta_f^{(\hat{Y})}$
& -2.87\stat{}{n}
& 3.08\stat{\sym{**}}{s}
& 15.04\stat{\sym{***}}{s}
& -0.67\stat{}{n}
& -6.30\stat{\sym{*}}{s}
& 7.99\stat{\sym{***}}{m}
& 7.01\stat{\sym{***}}{m}
& 4.73\stat{\sym{*}}{s}
& 4.97\stat{\sym{***}}{m}
& 2.78\stat{\sym{*}}{s}
& 2.03\stat{\sym{***}}{s}
& 1.88\stat{\sym{**}}{s}
& 1.37\stat{}{n} \\
\addlinespace[2pt]

\raisebox{-0.6ex}{\includegraphics[width=1.1em]{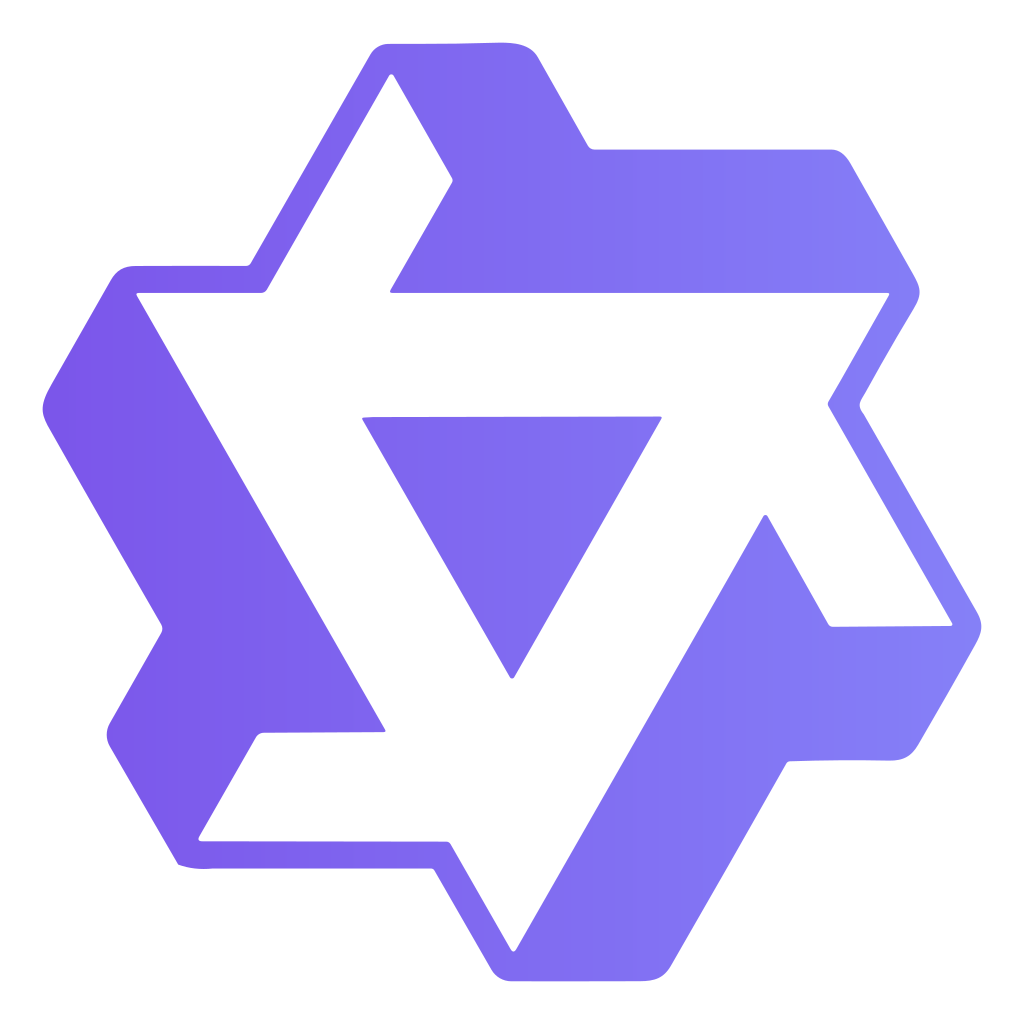}}\ Qwen2.5-VL
& $\Delta_f^{(\hat{Y})}$
& -2.51\stat{}{n}
& 4.99\stat{\sym{***}}{s}
& 14.41\stat{\sym{**}}{s}
& 0.86\stat{}{n}
& -5.35\stat{}{s}
& 8.78\stat{\sym{***}}{m}
& 6.15\stat{\sym{***}}{m}
& 1.60\stat{}{n}
& 5.49\stat{\sym{***}}{l}
& 4.73\stat{\sym{***}}{s}
& 2.73\stat{\sym{***}}{m}
& 2.48\stat{\sym{***}}{m}
& 1.47\stat{}{s} \\
\addlinespace[2pt]

\rowcolor{gray!15!violet!10}\multicolumn{15}{l}{\textbf{Panel C: VLM False Positives among Ground-Truth Low Items}} \\
\addlinespace[2pt]
\raisebox{-0.6ex}{\includegraphics[width=1.3em]{icons/gpt.png}}\ GPT-5
& $\Delta_f^{(FP-TN)}$
& 0.31\stat{}{n}
& 3.28\stat{\sym{*}}{s}
& 4.60\stat{}{n}
& -2.83\stat{}{n}
& -6.10\stat{}{s}
& 0.51\stat{}{n}
& 0.42\stat{}{n}
& 0.26\stat{}{n}
& 1.31\stat{}{n}
& -0.66\stat{}{n}
& 2.18\stat{\sym{***}}{s}
& 1.72\stat{\sym{*}}{s}
& 1.19\stat{}{n} \\
\addlinespace[2pt]

\raisebox{-0.6ex}{\includegraphics[width=1.3em]{icons/gpt.png}}\ GPT-5-mini
& $\Delta_f^{(FP-TN)}$
& -2.05\stat{}{n}
& 2.29\stat{}{n}
& 7.80\stat{}{n}
& -2.53\stat{}{n}
& -5.99\stat{}{s}
& 3.81\stat{}{s}
& 1.06\stat{}{n}
& -0.90\stat{}{n}
& 1.78\stat{\sym{*}}{s}
& 1.28\stat{}{n}
& 2.02\stat{\sym{**}}{s}
& 1.53\stat{}{s}
& 1.79\stat{\sym{*}}{s} \\
\addlinespace[2pt]

\raisebox{-0.6ex}{\includegraphics[width=1.3em]{icons/gemma-color.png}}\ Gemma3
& $\Delta_f^{(FP-TN)}$
& -0.93\stat{}{n}
& 2.90\stat{}{s}
& 6.00\stat{}{n}
& -1.90\stat{}{n}
& -6.69\stat{\sym{*}}{s}
& 5.99\stat{\sym{*}}{s}
& 2.23\stat{\sym{*}}{s}
& 1.64\stat{}{n}
& 2.47\stat{\sym{**}}{s}
& 0.85\stat{}{n}
& 1.41\stat{}{n}
& 1.96\stat{\sym{**}}{s}
& 1.63\stat{\sym{*}}{s} \\
\addlinespace[2pt]

\raisebox{-0.6ex}{\includegraphics[width=1.1em]{icons/qwen-color.png}}\ Qwen2.5-VL
& $\Delta_f^{(FP-TN)}$
& -1.73\stat{}{n}
& 4.20\stat{\sym{**}}{s}
& 14.60\stat{\sym{**}}{s}
& 0.31\stat{}{n}
& -5.35\stat{}{s}
& 8.08\stat{\sym{**}}{s}
& 3.60\stat{\sym{**}}{s}
& 0.49\stat{}{n}
& 3.42\stat{\sym{***}}{m}
& -0.21\stat{}{n}
& 1.99\stat{\sym{**}}{s}
& 2.17\stat{\sym{***}}{s}
& 1.56\stat{}{n} \\
\addlinespace[2pt]

\bottomrule
\end{tabular}
\end{adjustbox}
\caption{
Human--VLM comparison summary across VPFs.
Panel A shows that VPFs separate human-labeled High and Low items, while Panel B shows that VLM predictions partially reproduce this human-aligned pattern.
Panel C further analyzes VLM false positives among ground-truth Low items, revealing that models often associate persuasion-relevant cues with erroneous High predictions.
\textbf{Overall, VLMs appear sensitive to several persuasive cues, but their false positives suggest an overgeneralization from persuasive cues to High persuasiveness.}
}

\label{tab:human_vlm_levels}

\footnotesize
\vspace{-1.4em}
\end{table*}

%% file: latex/sections/04_analysis_VPFs.tex
\section{Analysis of Visual Persuasive Factors}
\label{sec:vpf_analysis}
Our main human--VLM comparison results are shown in Tab.~\ref{tab:human_vlm_levels} with qualitative examples in Fig.~\ref{fig:vpf_qualitative_comparison}. 
To facilitate interpretation of Tab.~\ref{tab:human_vlm_levels}, we report stars\footnote{Stars indicate $p$-values: \sym{***} $p<.001$, \sym{**} $p<.01$, \sym{*} $p<.05$ (two-sided).} and standard effect sizes~\citep{cohen1988} for how strongly each factor separates High- from Low-persuasiveness groups: large$_l$ ($\geq 0.8$), medium$_m$ ($\geq 0.5$), small$_s$ ($\geq 0.2$), and negligible$_n$ ($< 0.2$). 

\paragraph{VPFs differentiate High- and Low-persuasiveness cases along interpretable visual and semantic dimensions.}
To test whether VPFs capture factors that distinguish human judgments, we first examine Panel A of Tab.~\ref{tab:human_vlm_levels}.
Panel A shows that the proposed VPFs meaningfully separate human-labeled High and Low items: High items show stronger VPF signals across perceptual, compositional, and semantic levels---\emph{e.g.}, representative color congruence, T3 saliency, and key object presence---whereas cues such as colorfulness and text presence are more associated with Low items.

\paragraph{VLMs partially follow, but do not fully reproduce, human-aligned VPF patterns.}
We compare Panel A and Panel B of Tab.~\ref{tab:human_vlm_levels}.
Panel B shows that VLM predictions broadly track part of this pattern: for example, representative color congruence and key object presence remain consistently associated with predicted High labels across models, with small-to-medium effect sizes. 
However, this alignment is only partial. 
Several factors, including human presence, CBI saliency, T3 saliency, and brightness score under negative strategies, show stronger associations with VLM predictions than with human judgments.
The brightness results provide a useful side observation: while raw brightness is not informative, the brightness score under negative strategies shows meaningful associations, suggesting that the persuasion strategy is also a meaningful factor to consider in visual persuasive understanding.
Conversely, text presence is 
associated with human-labeled Low persuasiveness but tends to be neutral or High for VLMs, and colorfulness shows a negative human association that models reflect only weakly and inconsistently.
Overall, VLMs appear sensitive to several human-aligned VPFs, but they do not consistently weight or contextualize these factors in the same way as humans.

\paragraph{VLM false positives arise from overgeneralizing persuasion-relevant cues.}
Motivated by the recall-oriented bias observed in \S\ref{sec:vlm_eval}, we further analyze VLM false positives through the lens of VPFs.
Panel C of Tab.~\ref{tab:human_vlm_levels} focuses on human-labeled Low items and compares cases that VLMs incorrectly predict as High against those correctly predicted as Low. 
This comparison reveals that some factors that look human-aligned in Panel B can also drive erroneous High predictions. 
For example, key object presence is associated with human High labels and VLM High predictions in Panels A and B, but it is also significantly associated with false positives in Panel C. 
Moreover, human and text presence are more strongly tied to VLM false positives than to human persuasiveness judgments, suggesting that models overgeneralize semantic presence cues as persuasive evidence. A similar trend also appears for some perceptual and compositional cues, with factors such as T3 saliency, color entropy, and brightness score under negative strategies elevated among false positives. Since all items in this comparison are human-labeled Low, these results indicate that VLMs often mistake plausible persuasive visual cues for sufficient evidence of persuasiveness, rather than evaluating whether those cues genuinely support the target message.

\input{latex/sections/prediction_signal}

\paragraph{VPF findings generalize and provide complementary predictive signals.}
Across progressively broader agreement levels (with at least minimal annotator agreement) and diverse model scales, families, and reasoning variants, the recall-oriented bias, partial reproduction of human-aligned VPF patterns, and false-positive cue overgeneralization persist (results in Appendix~\ref{app:generalization_results}).
Beyond this diagnostic robustness, we test whether VPFs also serve as explicit predictive signals for human persuasiveness judgments.
Using logistic regression over 100 repeated stratified 80/20 train--test splits to ensure broad sample coverage, we compare VPF-only, VLM-prediction-only, and combined VLM+VPF inputs.
As shown in Tab.~\ref{tab:vpf_prediction}, VPFs alone achieve 69.5\% accuracy and 71.8\% AUC, while adding them to VLM predictions consistently improves AUC by 6.3--10.9\%p. 
Together, these results show that VPFs support robust diagnosis across settings while providing predictive information complementary to raw VLM judgments.

\input{latex/sections/experiment_main_table}
\input{latex/sections/CI_result_table}

%% file: latex/sections/prediction_signal.tex
\begin{table}[t]
\centering
\small
\setlength{\tabcolsep}{3.5pt}
\renewcommand{\arraystretch}{1.05}
\resizebox{\columnwidth}{!}{%
\begin{tabular}{llccc}
\toprule
\textbf{VLM Source} & \textbf{Input Features}
& \textbf{Acc. (\%)} & \textbf{AUC (\%)}
& \textbf{$\Delta$AUC (pp)} \\
\midrule
--            & VPFs                 & 69.5 & 71.8 & -- \\
\midrule
\raisebox{-0.6ex}{\includegraphics[width=1.3em]{icons/gpt.png}}\ GPT-5         & VLM pred.       & 72.0 & 78.8 & -- \\
\raisebox{-0.6ex}{\includegraphics[width=1.3em]{icons/gpt.png}}\ GPT-5         & VLM pred. + VPFs & 75.9 & 85.1 & \textbf{+6.3} \\
\addlinespace[2pt]
\raisebox{-0.6ex}{\includegraphics[width=1.3em]{icons/gpt.png}}\ GPT-5-mini    & VLM pred.       & 66.7 & 70.5 & -- \\
\raisebox{-0.6ex}{\includegraphics[width=1.3em]{icons/gpt.png}}\ GPT-5-mini    & VLM pred. + VPFs & 71.3 & 79.6 & \textbf{+9.1} \\
\addlinespace[2pt]
\raisebox{-0.6ex}{\includegraphics[width=1.3em]{icons/gemma-color.png}}\ Gemma3        & VLM pred.       & 66.4 & 71.0 & -- \\
\raisebox{-0.6ex}{\includegraphics[width=1.3em]{icons/gemma-color.png}}\ Gemma3        & VLM pred. + VPFs & 73.1 & 79.9 & \textbf{+8.9} \\
\addlinespace[2pt]
\raisebox{-0.6ex}{\includegraphics[width=1.3em]{icons/qwen-color.png}}\ Qwen2.5-VL    & VLM pred.       & 66.7 & 64.5 & -- \\
\raisebox{-0.6ex}{\includegraphics[width=1.3em]{icons/qwen-color.png}}\ Qwen2.5-VL    & VLM pred. + VPFs & 70.2 & 75.4 & \textbf{+10.9} \\
\bottomrule
\end{tabular}%
}
\caption{VPF-aware prediction of human ground-truth labels. VPFs carry predictive signal and provide additive gains beyond VLM predictions in logistic regression.}
\label{tab:vpf_prediction}
\vspace{-1.4em}
\end{table}

%% file: latex/sections/experiment_main_table.tex
\begin{table*}[t]
\centering
\renewcommand{\arraystretch}{1.1}
\setlength{\tabcolsep}{2.5pt}
\setlength{\abovecaptionskip}{5pt}
\setlength{\belowcaptionskip}{0pt}
\begin{adjustbox}{max width=\linewidth}
\small
\begin{tabular}{l@{\hskip 4pt}cccc@{\hskip 8pt}cccc@{\hskip 8pt}cccc}
\toprule

\rowcolor{violet!15}
\multicolumn{13}{l}{\textbf{Non-Aware Key Object}} \\
\addlinespace[3pt]

\textbf{Model} &
\multicolumn{4}{c}{\textbf{Baseline}} &
\multicolumn{4}{c}{\textbf{Chain-of-Thoughts}} &
\multicolumn{4}{c}{\textbf{Knowledge-Conditioned Chain}} \\
\cmidrule(lr){2-5}\cmidrule(lr){6-9}\cmidrule(lr){10-13}
& Acc. & Prec. & Rec. & F1 
& Acc. & Prec. ($\Delta$) & Rec. ($\Delta$) & F1 ($\Delta$) 
& Acc. & Prec. ($\Delta$) & Rec. ($\Delta$) & F1 ($\Delta$) \\
\midrule
\raisebox{-0.6ex}{\includegraphics[width=1.3em]{icons/gpt.png}}\ GPT-5       
& 69.57 & 51.01 & 98.88 & 67.30 
& 70.11 & 51.52 \dUp{0.51} & 94.94 \dDown{3.94} & 66.80 \dDown{0.50}   
& 65.48 & 47.81 \dDown{3.20} & 98.31 \dDown{0.57} & 64.34 \dDown{2.96} \\

\raisebox{-0.6ex}{\includegraphics[width=1.3em]{icons/gpt.png}}\ GPT-5-mini  
& 60.14 & 44.13 & 97.19 & 60.70 
& 63.88 & 46.63 \dUp{2.50} & 97.19 {\scriptsize(0.00)} & 63.02 \dUp{2.32}    
& 63.35 & 46.30 \dUp{2.17} & 98.31 \dUp{1.12} & 62.95 \dUp{2.25}   \\

\raisebox{-0.6ex}{\includegraphics[width=1.3em]{icons/gemma-color.png}}\ Gemma3      
& 63.88 & 46.33 & 88.76 & 60.89 
& 51.25 & 38.84 \dDown{7.49} & 93.82 \dUp{5.06} & 54.93 \dDown{5.96}  
& 39.15 & 33.98 \dDown{12.35} & 97.75 \dUp{8.99} & 50.43 \dDown{10.46}\\

\raisebox{-0.6ex}{\includegraphics[width=1.1em]{icons/qwen-color.png}}\ Qwen2.5-VL  
& 54.98 & 40.51 & 89.89 & 55.85 
& 68.09 & 49.82 \dUp{9.31} & 76.40 \dDown{13.49} & 60.31 \dUp{4.46}    
& 57.65 & 42.46 \dUp{1.95} & 94.94 \dUp{5.05} & 58.68 \dUp{2.83}   \\

\midrule

\rowcolor{violet!15}
\multicolumn{13}{l}{\textbf{Aware Key Object}} \\
\addlinespace[3pt]

\textbf{Model} &
\multicolumn{4}{c}{\textbf{Aligned Key Object Context}} &
\multicolumn{4}{c}{\textbf{Key Object-Aware CoT}} &
\multicolumn{4}{c}{\textbf{Rationale Injection}} \\
\cmidrule(lr){2-5}\cmidrule(lr){6-9}\cmidrule(lr){10-13}
& Acc. & Prec. ($\Delta$) & Rec. ($\Delta$) & F1 ($\Delta$) 
& Acc. & Prec. ($\Delta$) & Rec. ($\Delta$) & F1 ($\Delta$) 
& Acc. & Prec. ($\Delta$) & Rec. ($\Delta$) & F1 ($\Delta$) \\
\midrule
\raisebox{-0.6ex}{\includegraphics[width=1.3em]{icons/gpt.png}}\ GPT-5       
& 69.93 & 51.44 \dUp{0.43} & 90.45 \dDown{8.43} & 65.58 \dDown{1.72} 
& 63.70 & 46.41 \dDown{4.60} & 94.38 \dDown{4.50} & 62.22 \dDown{5.08} 
& 69.40 & 50.89 \dDown{0.12} & 96.07 \dDown{2.81} & 66.54 \dDown{0.86} \\

\raisebox{-0.6ex}{\includegraphics[width=1.3em]{icons/gpt.png}}\ GPT-5-mini  
& 58.72 & 43.11 \dDown{1.02} & 94.94 \dDown{2.25} & 59.30 \dDown{1.40} 
& 54.98 & 41.18 \dDown{2.95} & 98.31 \dUp{1.12} & 58.04 \dDown{2.66} 
& 64.06 & 46.79 \dUp{2.66} & 98.31 \dUp{1.12} & 63.41 \dUp{2.71} \\

\raisebox{-0.6ex}{\includegraphics[width=1.3em]{icons/gemma-color.png}}\ Gemma3       
& 62.28 & 45.06 \dDown{1.27} & 87.08 \dDown{1.68} & 59.39 \dDown{1.50} 
& 46.44 & 36.89 \dDown{9.44} & 97.19 \dUp{8.43} & 53.48 \dDown{7.41} 
& 59.79 & 43.65 \dDown{2.68} & 92.70 \dUp{3.94} & 59.35 \dDown{1.54} \\

\raisebox{-0.6ex}{\includegraphics[width=1.1em]{icons/qwen-color.png}}\ Qwen2.5-VL  
& 53.02 & 39.56 \dDown{0.95} & 91.57 \dUp{1.68} & 55.25 \dDown{0.60} 
& 62.46 & 45.22 \dUp{4.71} & 87.64 \dDown{2.25} & 59.66 \dUp{3.81} 
& 53.91 & 39.37 \dDown{1.14} & 84.27 \dDown{5.62} & 53.67 \dDown{2.18} \\

\bottomrule
\end{tabular}
\end{adjustbox}
\caption{Results of semantic grounding and reasoning scaffolds on VLM performance. $\Delta$ next to Precision, Recall, and F1 shows the absolute percentage-point change relative to the model's own baseline scores ($\uparrow/\downarrow$ for direction).}
\label{tab:main_result}
\vspace{-1.0em}
\end{table*}

%% file: latex/sections/CI_result_table.tex
\begin{table}[t]
\centering
\renewcommand{\arraystretch}{1.08}
\setlength{\tabcolsep}{5.8pt}
\setlength{\abovecaptionskip}{5pt}
\setlength{\belowcaptionskip}{0pt}
\begin{adjustbox}{max width=\linewidth}
\small
\begin{tabular}{lccccc}
\toprule
\textbf{Injection Type} &
\textbf{$\Delta$FP Rate} &
\textbf{$\Delta$Prec.} &
\textbf{$\Delta$Rec.} &
\textbf{$\Delta$F1} &
\makecell{\textbf{No. of Models}\\\textbf{with F1 Gain}} \\
\midrule

\rowcolor{blue!8}
\multicolumn{6}{l}{\textbf{Panel A. Factor-wise Cognitive Injection}} \\

Color
& \fpDown{5.60} & \dUp{1.98} & \dDown{2.81} & \dUp{1.06} & 3/4 \\

Brightness
& \fpDown{5.01} & \dUp{2.15} & \best{\dDown{0.70}} & \dUp{1.76} & 3/4 \\

Saliency-Center
& \fpDown{5.66} & \dUp{2.10} & \dDown{2.39} & \dUp{1.30} & 3/4 \\

MRR-Center
& \fpDown{5.34} & \dUp{2.12} & \dDown{1.54} & \dUp{1.53} & 3/4 \\

Saliency-RoT
& \second{\fpDown{7.68}} & \second{\dUp{3.25}} & \dDown{2.25} & \best{\dUp{2.32}} & 4/4 \\

MRR-RoT
& \fpDown{5.08} & \dUp{1.94} & \dDown{1.54} & \dUp{1.38} & 3/4 \\

KeyObj
& \fpDown{3.91} & \dUp{1.46} & \dDown{1.69} & \dUp{0.88} & 4/4 \\

Human
& \best{\fpDown{9.38}} & \best{\dUp{3.61}} & \dDown{4.35} & \second{\dUp{1.95}} & 4/4 \\

Text
& \fpDown{5.27} & \dUp{2.13} & \second{\dDown{1.40}} & \dUp{1.52} & 3/4 \\

\midrule
\rowcolor{green!8}
\multicolumn{6}{l}{\textbf{Panel B. All-factor Cognitive Injection}} \\

All
& \fpDown{1.03} & \dUp{0.42} & \dDown{0.28} & \dUp{0.31} & 2/4 \\

\bottomrule
\end{tabular}
\end{adjustbox}
\caption{Average false-positive mitigation and performance effects of cognitive injection across four VLMs relative to each model's baseline in Tab.~\ref{tab:main_result}. All $\Delta$ values are computed as $\text{injection} - \text{baseline}$.}
\label{tab:ci_fp_mitigation_summary}
\vspace{-1.4em}
\end{table}

%% file: latex/sections/05_rationale.tex
\section{VPF-Guided Knowledge Injection and Reasoning}
The analysis in \S\ref{sec:vpf_analysis} motivates two questions.
First, if VLMs are already sensitive to VPFs but overuse them as generic positive signals, can explicit VPF knowledge help models treat these cues as context-dependent evidence rather than sufficient evidence of persuasiveness?
Second, can reasoning scaffolds help models integrate available visual evidence into a persuasive judgment?
We refer to the latter ability as \emph{persuasive integration}: the ability to use visual cues as evidence for or against the communicative intent of an image-message pair.
We test this via VPF-guided interventions, focusing on key object cues because they reliably separate human judgments yet also contribute to VLM false positives.

\subsection{Task Description}
\label{sec:keyobj_injection}
\paragraph{Testing cue use through knowledge injection.}
To test whether explicit VPF knowledge helps models avoid treating persuasion-relevant cues as determinative of persuasiveness, we consider two injection variants: Factor-aware Cognitive Injection and Aligned Key Object Context. 
(i) Factor-aware Cognitive Injection provides the model with explicit knowledge of the proposed VPFs during evaluation, either one factor at a time or all nine factors together, while presenting them only as context-dependent auxiliary cues and not prescribing any fixed effect on the persuasive judgment.
(ii) Aligned Key Object Context supplies compact structured information about which detected key object candidates are semantically aligned with the message (\emph{e.g.}, ``devices'' for the message ``remove unnecessary electronic devices''). If no key object is found, no additional context is provided.

\paragraph{Testing persuasive integration through reasoning scaffolds.}
To examine whether structured reasoning can improve persuasive judgment, we consider three scaffold variants: (i) general Chain-of-Thought (CoT), which appends the instruction ``\emph{Let's think step by step}''
to encourage explicit stepwise reasoning~\citep{cot}; (ii) Knowledge-Conditioned Chain, a lightweight reasoning scaffold that explicitly prompts the model to (a) identify candidate key objects, (b) assess their alignment with the message, and (c) make a final persuasiveness judgment; and (iii) Key Object-Aware CoT, which uses the same step-by-step prompting while additionally informing the model of the relevant key object cue (\emph{i.e.}, Aligned Key Object Context setting). 
Both task prompts are in Appendix~\ref{app:pr_sec}.

\subsection{Results}
\label{sec:grounding_reasoning_result}

\paragraph{Explicit VPF knowledge reveals cue-framing sensitivity.}
In Tab.~\ref{tab:ci_fp_mitigation_summary}, Factor-aware Cognitive Injection yields a general trend of lower false-positive rates and higher F1 scores across models for most VPFs (full results are in Appendix~\ref{app:knolwedge_injection_and_reasoning}).
This suggests that making VPFs explicit as context-dependent supporting cues can mitigate the recall-oriented bias by discouraging models from treating persuasion-relevant cues as sufficient evidence of persuasiveness. Thus, the bias is not merely a fixed tendency; VLM judgments are sensitive to how persuasive cues are framed.
The contrast with Aligned Key Object Context in Tab.~\ref{tab:main_result} makes this clearer.
The consistent F1 drops show that direct key object context does not resolve the bias, aligning with the earlier finding that key object cues are associated with VLM false positives.
This indicates that direct cue specification, without contextual framing, is insufficient; VLMs still lack a stable criterion for when such cues should count as persuasive evidence.

\paragraph{Reasoning scaffolds expose the integration bottleneck.}
Reasoning-based experiments show that explicit self-reasoning is not, by itself, a robust remedy for persuasive judgment. General CoT and Knowledge-Conditioned Chain yield mixed effects, indicating that simply prompting the model to reason step by step does not consistently improve performance. 
More importantly, when VLMs reason with explicit key object cues, the recall-oriented bias can be amplified, as shown by Key Object-Aware CoT (\emph{e.g.}, Gemma3: Prec. $-9.44$, Rec. $+8.43$, F1 $-7.41$).
Together with the knowledge injection results, this suggests that the remaining bottleneck lies in persuasive integration.
This motivates a finer-grained diagnosis of where the integration process breaks down.

\begin{figure*}[t]
    \centering
    \setlength{\abovecaptionskip}{3pt}
    \setlength{\belowcaptionskip}{0pt}
    \includegraphics[width=\linewidth, trim=0 4 0 2, clip]{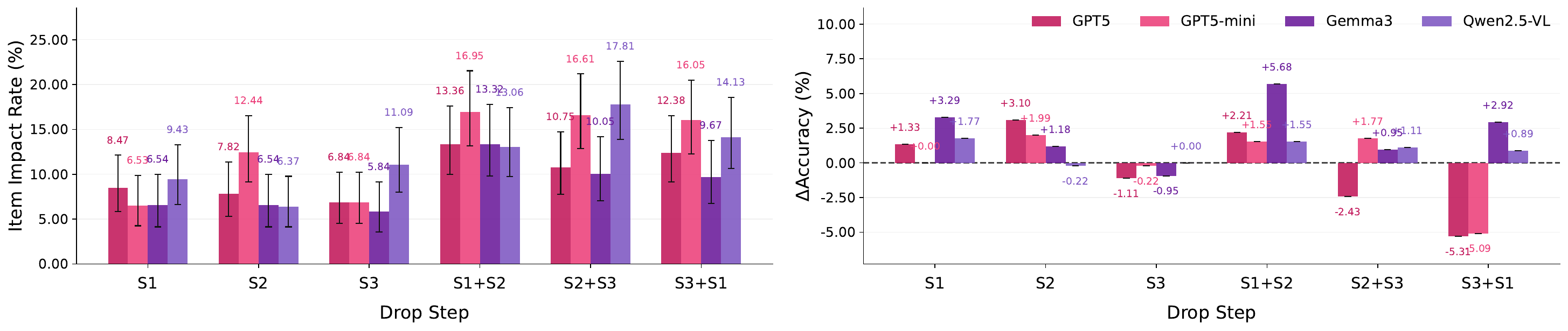}
    \caption{(Left) Item impact rate, the proportion of predictions changed by each single- or pair-step ablation. (Right) Accuracy change, measuring the corresponding change in binary prediction accuracy.}
    \label{fig:step_ablation}
    \vspace{-1.4em}
\end{figure*}

\subsection{Diagnosing Persuasive Integration through Rationale Analysis}
We analyze model rationales by decomposing them into functional steps to identify which components are most critical for persuasive integration. 
Prior work treats rationales as supporting evidence for model predictions and evaluates their alignment and faithfulness~\citep{lei-etal-2016-rationalizing,deyoung-etal-2020-eraser}. 
Building on this view, we use rationales as a diagnostic proxy for examining how models use evidence and integrate it into persuasive judgments.

\paragraph{Step-wise rationale ablation.}
For each detected key object from the VPF pipeline (\S\ref{sec:vpf_extraction}), we prompt the model to generate a per-object rationale explaining whether and how that object contributes to persuasiveness (details in Appendix~\ref{app:rationale_gen}).
We decompose each rationale into three functional steps: $S1$, which identifies the relevant object; $S2$, which explains how that evidence semantically supports or undermines the message; and $S3$, which connects the preceding analysis to the final persuasive judgment.
The functional spans are tagged by GPT-5 under a text-preserving constraint.
To ensure semantic parsing reliability, we report results only on the 92.04\% of rationale segmentations verified as correct by human evaluators (details in Appendix~\ref{app:stepwise_ablation}).

We then assess both individual-step contributions and their interactions through single- and pair-step ablations, selectively removing one or two tagged spans and re-evaluating the model on the same item relative to the Rationale Injection output in Tab.~\ref{tab:main_result}.
The full rationale injection prediction serves as the baseline for comparison.
For each ablation, we measure (i) the item impact rate, defined as the rate at which predictions change, and (ii) the change in prediction accuracy against the ground-truth label. 
In addition, as a robustness check against incidental sentence-form effects, we shuffle the rationale steps while preserving their content and find that model behavior remains 93.79--96.45\% consistent, indicating that it is driven more by the rationale content itself.
Full parsing procedure and details are in Appendix~\ref{app:stepwise_ablation}.

\paragraph{Results.}
Fig.~\ref{fig:step_ablation} (Left) shows that all three rationale steps meaningfully affect model behavior. 
Removing a single step changes predictions for a non-trivial portion of samples, and pair-step ablations all produce double-digit item impact rates across models.
Thus, earlier steps are not simply ignored; rather, each part of the rationale actively intervenes in the model's decision process.
However, Fig.~\ref{fig:step_ablation} (Right) reveals an asymmetry in how these components affect correctness.
While ablating $S1$ or $S2$ often leaves accuracy unchanged or even improves it, removing $S3$ alone tends to hurt performance more directly. 
The clearest pattern appears in pair-step ablations: dropping $S1{+}S2$ improves accuracy across all models.
This sharpens the integration-failure account. 
A central weakness appears to lie in how object identification ($S1$) and semantic message alignment ($S2$) are connected to the final persuasive claim.
In other words, VLMs do not reliably know how that evidence should count for or against the communicative intent of the image-message pair. 
We report an additional false-positive-focused analysis in Appendix~\ref{app:step_wise_additional_result}.

%% file: latex/sections/06_conclusion.tex
\section{Conclusion}
Our work investigates visual persuasive understanding in VLMs from a cognitive psychology perspective.
We introduce Visual Persuasive Factors, a structured taxonomy for analyzing how visual cues shape persuasiveness judgments. 
We show that even in high-consensus cases, VLMs exhibit a recall-oriented bias; VPF analysis further reveals that VLMs only partially reproduce human-aligned factor patterns, and their false positives arise from conflating persuasion-relevant cues with message support.
Building on this insight, we evaluate VPF-guided interventions and find that explicit factor knowledge can help only when framed as auxiliary cues rather than fixed evidence, while naive cue grounding or step-by-step reasoning is insufficient.
Finally, rationale ablation identifies persuasive integration as a key bottleneck, highlighting that VLM-based visual persuasion evaluation requires connecting visual evidence to communicative intent.

%% file: latex/sections/appendix.tex
\section{Dataset Details}
\label{app:dataset_details}

\begin{table}[t]
\centering
\renewcommand{\arraystretch}{1.08}
\setlength{\tabcolsep}{6pt}

\begin{adjustbox}{max width=\columnwidth}
\begin{tabular}{lr}
\toprule
\textbf{Topic} & \textbf{Proportion (\%)} \\
\midrule
Safety Awareness               & 11.39 \\
Housing and Urban Development  & 8.90 \\
Exercise                       & 8.54 \\
Homeland Security              & 8.54 \\
Transportation                 & 8.36 \\
Defense                        & 7.83 \\
Treasury                       & 6.23 \\
Justice                        & 5.69 \\
Advertising                    & 5.16 \\
Privacy                        & 4.27 \\
Agriculture                    & 4.09 \\
Energy                         & 3.74 \\
Education                      & 3.74 \\
Veterans Affairs (VA)          & 3.02 \\
Labor                          & 2.49 \\
Commerce                       & 2.14 \\
Health and Human Services      & 1.96 \\
Interior                       & 1.42 \\
State                          & 1.25 \\
Cyber Etiquette                & 1.25 \\
\bottomrule
\end{tabular}
\end{adjustbox}

\caption{
Topic distribution of the 562-image high-consensus subset.
All 20 topics represented in the original PVP dataset are retained.
}
\label{tab:subset_topic_distribution}
\end{table}

\begin{table}[t]
\centering
\renewcommand{\arraystretch}{1.08}
\setlength{\tabcolsep}{8pt}

\begin{adjustbox}{max width=\columnwidth}
\begin{tabular}{lr}
\toprule
\textbf{Image Source} & \textbf{Proportion (\%)} \\
\midrule
DALL$\cdot$E-generated & 50.00 \\
Web-retrieved          & 50.00 \\
\bottomrule
\end{tabular}
\end{adjustbox}

\caption{
Image-source distribution of the 562-image high-consensus subset.
}
\label{tab:subset_source_distribution}
\end{table}

\begin{table*}[t]
\centering
\small
\renewcommand{\arraystretch}{1.12}
\setlength{\tabcolsep}{6pt}
\begin{tabular}{lcc}
\toprule
\textbf{Metric} & \textbf{Original (0--10)} & \textbf{Banded (Low/Mid/High)} \\
\midrule
Overall Fleiss' $\kappa$ & 0.0233 & 0.0629 \\
Mean per-item $\kappa$   & 0.023  & 0.063  \\
Median per-item $\kappa$ & 0.003  & 0.011  \\
\midrule
\multicolumn{3}{l}{\textit{Kappa level distribution~\cite{kappa_score} (count of images)}} \\
\quad Worse than chance  & 12{,}424 & 13{,}186 \\
\quad Slight             & 13{,}176 & 11{,}318 \\
\quad Fair               & 1{,}264  & 1{,}035 \\
\quad Moderate           & 1{,}501  & 126 \\
\quad Substantial        & 3        & 0 \\
\quad Almost perfect     & 86       & 2{,}789 \\
\midrule
\multicolumn{3}{l}{\textit{Robust subset (constructed from \textbf{banded} ``Almost perfect'')}} \\
\quad Almost-perfect images (banded) & --- & 2{,}789 \\
\quad Kept (Low/High only, dedup by image) & --- & 562 (Low: 384, High: 178) \\
\quad Mid-only excluded               & --- & 2{,}227 \\
\bottomrule
\end{tabular}

\caption{
Inter-annotator agreement statistics and construction of the high-consensus subset.
We compare the original 0--10 integer scores with a banded scheme that maps scores to \textit{Low} (0--2), \textit{Mid} (3--7), and \textit{High} (8--10).
Fleiss' $\kappa$ is reported overall together with the per-item distribution of agreement levels.
}
\label{tab:dataset_agreement}
\end{table*}

\paragraph{Subset composition.}
The resulting 562-image high-consensus subset spans all 20 topics represented in PVP and contains 184 unique messages, corresponding to an average of 3.05 images per message.
Tab.~\ref{tab:subset_topic_distribution} reports the topic distribution.
The subset also contains equal proportions of DALL$\cdot$E-generated and Web-retrieved images, as summarized in Tab.~\ref{tab:subset_source_distribution}.

\paragraph{Dataset and agreement filtering.}
We use the PVP dataset~\cite{pvp}, which is released under CC-BY-4.0.
Each message in the dataset is paired with approximately 54 images, providing multiple visual realizations for the same persuasive message.
Each image is rated by four annotators with distinct psychological profiles, who assign a persuasiveness score from 0 to 10 based on the image--message pair.
Because the four raters may agree or disagree for a given image, we analyze inter-annotator agreement and filter the dataset accordingly.
Specifically, we consider two schemes: (i) the original 0--10 integer scores and (ii) a banded scheme (\textit{Low}=0--2, \textit{Mid}=3--7, \textit{High}=8--10).
Fleiss' $\kappa$ over the full corpus is low under both schemes, while banding produces a larger set of ``Almost perfect'' agreement items (Tab.~\ref{tab:dataset_agreement}).

\paragraph{Robust high-consensus core subset.}
To obtain a subset reflecting stable human persuasiveness judgments, we select banded items with ``Almost perfect'' agreement among the four raters and retain only images whose consensus label is \textit{Low} or \textit{High}, discarding \textit{Mid}.
After deduplication by image path, this yields 562 images (384 \textit{Low}, 178 \textit{High}) from the 2{,}789 ``Almost perfect'' images, with 2{,}227 \textit{Mid}-only cases excluded.

\section{Detailed VPF and Extracted Examples}
\label{app:vpf_details}

This section provides the detailed definitions and equations for the VPF features described in Section~\ref{sec:vpf_extraction}.

\subsection{Perceptual Level}
\label{app:vpf_low}
The following section describes how we extracted features related to color and brightness and how we used them as comparison metrics. For metrics such as colorfulness and entropy, we reported the normalized values (\emph{i.e.}, those included in Tab.~\ref{tab:human_vlm_levels}) for comparison purposes.

\begin{itemize}
[leftmargin=*, itemsep=0pt, topsep=2pt, parsep=0pt]
    \item \textbf{Colorfulness:} To measure color intensity, we use the colorfulness metric of~\citet{colofulness_hasler}. 
    Given RGB channels, we first compute two opponent-color components:
    \begin{align}
        o_{rg} &= R - G, \\
        o_{yb} &= \frac{1}{2}(R + G) - B.
    \end{align}
    We then summarize their dispersion and average magnitude as
    \begin{align}
        s_o &= \sqrt{\operatorname{std}(o_{rg})^2 + \operatorname{std}(o_{yb})^2}, \\
        m_o &= \sqrt{\operatorname{mean}(o_{rg})^2 + \operatorname{mean}(o_{yb})^2},
    \end{align}
    and define the image colorfulness score as
    \begin{align}
        C_{\mathrm{colorfulness}} = s_o + 0.3\,m_o .
    \end{align}
    \noindent\textit{Interpretation}: Typical reference values are 0 (not colorful), 15 (slightly), 33 (moderately), 45 (average), 59 (quite), 82 (highly), and 109 (extremely colorful).
    
    \begin{table}[H]
    \centering
    \small
    \begin{tabular}{lccc}
    \toprule
     & \textbf{High} & \textbf{Low} & \textbf{\textbf{Effect ($d$, High--Low)}} \\
    \midrule
    \raisebox{-0.7ex}{\includegraphics[width=1.3em]{icons/audience.png}}\ Human
    & 29.28 & 33.22 & -0.23 \\
    \midrule
    \raisebox{-0.7ex}{\includegraphics[width=1.3em]{icons/gpt.png}}\ GPT-5
    & 31.33 & 32.99 & -0.10 \\
    \raisebox{-0.7ex}{\includegraphics[width=1.3em]{icons/gpt.png}}\ GPT-5-mini
    & 30.91 & 34.42 & -0.21 \\
    \raisebox{-0.7ex}{\includegraphics[width=1.3em]{icons/gemma-color.png}}\ Gemma3
    & 30.85 & 33.71 & -0.17 \\
    \raisebox{-0.7ex}{\includegraphics[width=1.3em]{icons/qwen-color.png}}\ Qwen2.5
    & 31.23 & 33.74 & -0.15 \\
    \bottomrule
    \end{tabular}
    \caption{Normalized colorfulness scores (Hasler metric) for high- and low-persuasiveness groups, with Cohen’s $d$ effect sizes.}
    \label{tab:cf_raw}
    \end{table}
    
    \item \textbf{Color Entropy:} To measure color diversity, we converted the image into the CIELAB color space and quantized the chrominance channels ($a^*, b^*$) into a 2D histogram. We then calculated the amount of information in the color distribution, $H_{\text{color}}$, using the Shannon entropy formula, where a higher value indicates a more uniform distribution of diverse colors.
    \begin{equation}
        H_{\text{color}} = - \sum_{i} p_i \log_2 p_i
    \end{equation}
    where $p_i$ is the probability of the $i$-th color bin in the $a^*b^*$ histogram.

    \begin{table}[H]
    \centering
    \small
    \begin{tabular}{lccc}
    \toprule
     & \textbf{High} & \textbf{Low} & \textbf{\textbf{Effect ($d$, High--Low)}} \\
    \midrule
    \raisebox{-0.7ex}{\includegraphics[width=1.3em]{icons/audience.png}}\ Human
    & 42.97 & 38.40 & 0.35 \\
    \midrule
    \raisebox{-0.7ex}{\includegraphics[width=1.3em]{icons/gpt.png}}\ GPT-5
    & 41.71 & 36.89 & 0.37 \\
    \raisebox{-0.7ex}{\includegraphics[width=1.3em]{icons/gpt.png}}\ GPT-5-mini
    & 40.96 & 37.28 & 0.28 \\
    \raisebox{-0.7ex}{\includegraphics[width=1.3em]{icons/gemma-color.png}}\ Gemma3
    & 41.06 & 37.98 & 0.24 \\
    \raisebox{-0.7ex}{\includegraphics[width=1.3em]{icons/qwen-color.png}}\ Qwen2.5
    & 41.33 & 36.34 & 0.39 \\
    \bottomrule
    \end{tabular}
    \caption{Normalized color entropy scores for high- and low-persuasiveness groups, with Cohen’s $d$ effect sizes.}
    \label{tab:entropy_raw}
    \end{table}

    \item \textbf{Representative Color Congruence}:
    To measure the alignment between perceptual color properties and semantic framing, we define a discrete eight-category color palette based on the affective color space introduced by~\citet{Cross_Cultural_color}. Each color is associated with a core affective profile (\emph{e.g.}, \textit{strong}, \textit{pleasant}, \textit{calm}), as summarized in Tab.~\ref{tab:color_palette}. 

    \begin{table}[htbp]
        \centering
        \small
        \begin{tabular}{llc}
        \toprule
        \textbf{Color} & \textbf{Core Affective Profile} & \textbf{Proportion (\%)} \\
        \midrule
        Red & strong, salient & 0.71 \\
        Blue & pleasant, calm & 12.81 \\
        Green & good, calm & 2.14 \\
        Yellow & positive, active & 4.27 \\
        White & pleasant, weak & 19.93 \\
        Black & bad, strong & 31.67 \\
        Grey & bad, passive & 17.97 \\
        Brown/Purple & contextual, variable & 10.50 \\
        \bottomrule
        \end{tabular}
        \caption{Discrete color palette, affective profiles based on~\citet{Cross_Cultural_color}, and representative color distributions extracted from dominant K-means clusters.}
        \label{tab:color_palette}
    \end{table}

    For the image-side representation, we extract a representative color using K-means clustering ($K=8$). 
    We select the dominant cluster, defined as the cluster with the largest pixel proportion, and map its centroid RGB value to the nearest palette category. 
    The resulting distribution of representative colors is shown in Tab.~\ref{tab:color_palette}.

    \begin{table}[htbp]
    \centering
    \small
    \begin{tabular}{lcc}
    \toprule
    \textbf{Color} & \textbf{High (\%)} & \textbf{Low (\%)} \\
    \midrule
    Black & 25.66 & 16.06 \\
    Blue & 10.11 & 21.18 \\
    Brown/Purple & 0.00 & 0.17 \\
    Green & 7.68 & 14.32 \\
    Grey & 15.36 & 11.37 \\
    Red & 22.85 & 16.67 \\
    White & 1.31 & 5.64 \\
    Yellow & 17.04 & 14.58 \\
    \bottomrule
    \end{tabular}
    \caption{Distribution of GPT-5 predicted top-3 color categories (message+ premise + affective profile) for high- and low-persuasiveness cases.}
    \label{tab:gpt_color_dist}
    \end{table}

    Motivated by our validation survey~\ref{app:human_survey_details}, where annotators selected 2.99 color categories per item on average, we infer the top-3 expected colors rather than a single color.
    For the message-side expectation, we use GPT-5 to infer three colors given the message and its persuasion premise, under three prompting conditions: (i) message only, (ii) message + premise, and (iii) message + premise with explicit affective profiles. 
    We then compute \textit{hit@3} overlap between image and message color categories. 
    As shown in Tab.~\ref{tab:color_hit}, incorporating both premise and affective profiles improves alignment, though effect sizes remain small to moderate. 
    We additionally report the distribution of GPT-5 predicted color categories (iii) in Tab.~\ref{tab:gpt_color_dist}.

    \begin{table}[htbp]
    \centering
    \small
    \begin{tabular}{lcc}
    \toprule
    \textbf{Color} & \textbf{High $\Delta$\%} & \textbf{Low $\Delta$\%} \\
    \midrule
    Red & 6.93 & 9.38 \\ 
    Blue & -3.00 & -5.56 \\
    Green & -3.93 & -6.42 \\
    Yellow & -8.43 & 4.25 \\
    White & -1.12 & -4.25 \\
    Black & 0.19 & -0.09 \\
    Grey & 9.36 & 1.30 \\
    Brown/Purple & 0.00 & 1.39 \\
    \bottomrule
    \end{tabular}
    \caption{Ablation results using affective profiles only.}
    \label{tab:color_ablation}
    \end{table}

    \begin{table*}[t]
    \centering
    \small
    \setlength{\tabcolsep}{6pt}
    \begin{tabular}{lcccc}
    \toprule
     & \textbf{Message Only} & \textbf{+ Premise} & \textbf{+ Premise + Affective Profile} & \textbf{Effect ($h$, High--Low)} \\
    \midrule
    \raisebox{-0.7ex}{\includegraphics[width=1.3em]{icons/audience.png}}\ Human (High) 
    & 0.410 & 0.500 & 0.534 & \multirow{2}{*}{0.304} \\
    \raisebox{-0.7ex}{\includegraphics[width=1.3em]{icons/audience.png}}\ Human (Low)  
    & 0.333 & 0.427 & 0.383 & \\
    \midrule
    \raisebox{-0.7ex}{\includegraphics[width=1.3em]{icons/gpt.png}}\ GPT-5      
    & 0.332 / 0.358 & 0.458 / 0.438 & 0.478 / 0.355 & 0.251 \\
    \raisebox{-0.7ex}{\includegraphics[width=1.3em]{icons/gpt.png}}\ GPT-5-mini 
    & 0.355 / 0.365 & 0.459 / 0.470 & 0.480 / 0.318 & 0.333 \\
    \raisebox{-0.7ex}{\includegraphics[width=1.3em]{icons/gemma-color.png}}\ Gemma3 
    & 0.375 / 0.330 & 0.463 / 0.430 & 0.490 / 0.339 & 0.307 \\
    \raisebox{-0.7ex}{\includegraphics[width=1.3em]{icons/qwen-color.png}}\ Qwen2.5 
    & 0.362 / 0.347 & 0.466 / 0.413 & 0.473 / 0.329 & 0.295 \\
    \bottomrule
    \end{tabular}
    \caption{Hit@3 overlap between image representative color and predicted colors under different prompting conditions. For VLMs, each cell reports prediction High / Low values. Effect sizes ($h$)~\cite{cohen1988} correspond to the High--Low difference for the \textit{+ Premise + Affective Profile} condition (OptA Hit@3). Statistical significance markers are reported in Tab.~\ref{tab:human_vlm_levels}.}
    \label{tab:color_hit}
    \end{table*}

    To examine whether the model is biased toward affective descriptors rather than color semantics, we conduct an ablation where only affective profiles are provided without explicit color categories. As shown in Tab.~\ref{tab:color_ablation}, the resulting distribution differs substantially from the color-aware setting, indicating that the model is not merely relying on affective descriptors. Instead, meaningful alignment emerges only when explicit color information is jointly considered with affective profiles.

    \item \textbf{Brightness:} To quantify perceptual lightness, we convert each image from sRGB to the CIELAB color space and use the $L^*$ channel, which approximates human luminance perception (ranging from 0 to 100). 
    The brightness metric is defined as the spatial mean of $L^*$:
    \begin{equation}
    \text{Brightness} = \frac{1}{HW}\sum_{x=1}^{H}\sum_{y=1}^{W} L^*(x,y).
    \end{equation}
    We implemented this computation using the \texttt{OpenCV} library (\texttt{cv2}), where $L^*$ is obtained by converting an image to the \texttt{LAB} color space and rescaling from the 0--255 range to the standard 0--100 scale. 
    Higher values indicate perceptually brighter images.
    
    Beyond measuring absolute brightness, we evaluate whether image brightness aligns with the semantic framing of the persuasion strategy inspired by~\citet{Cross_Cultural_color}. For each image, we first compute its brightness as the spatial mean of the $L^*$ channel in CIELAB space and normalize it to $x \in [0,1]$. 
    
    We then infer the polarity of the persuasion strategy (positive, negative, or neutral) from the message and its associated premise, using a rule-based mapping derived from the premise text. Given the normalized brightness $x$ and the inferred strategy polarity, we define a matching score that captures the degree of alignment between visual brightness and semantic intent:
    \begin{equation}
        \text{Score}(x)=
        \begin{cases}
        x & \mathrm{pos}\\
        1-x & \mathrm{neg}\\
        1-2|x-0.5| & \mathrm{neu}
        \end{cases}
    \end{equation}
        
    A higher score indicates stronger alignment between image brightness and the intended persuasion framing (\emph{e.g.}, brighter images for positive strategies and darker images for negative ones). 
 
\end{itemize}

\begin{figure*}[t]
  \centering

  \begin{minipage}[t]{0.32\linewidth}
    \includegraphics[width=\linewidth]{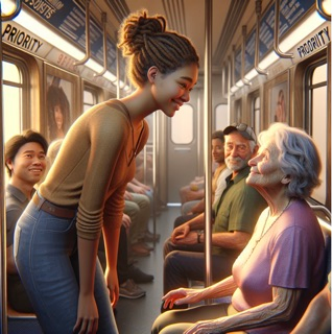}
    \captionof*{figure}{\scriptsize (a) Colorfulness average}
  \end{minipage}\hfill
  %
  \begin{minipage}[t]{0.32\linewidth}
    \includegraphics[width=\linewidth]{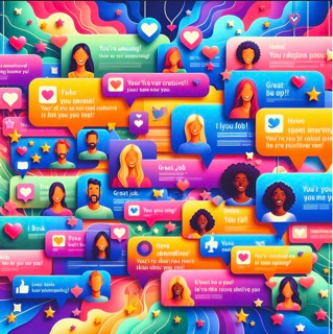}
    \captionof*{figure}{\scriptsize (b) Colorfulness high}
  \end{minipage}\hfill
  %
  \begin{minipage}[t]{0.32\linewidth}
    \includegraphics[width=\linewidth]{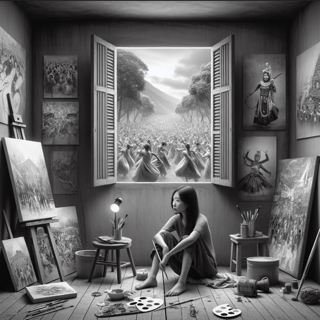}
    \captionof*{figure}{\scriptsize (c) Colorfulness low}
  \end{minipage}

  \vspace{4pt}
  \caption{Illustrative perceptual-level examples used in Appendix~\ref{app:low_example}}
  \label{fig:lowlevel_examples_row3}
\end{figure*}

\paragraph{Extraction Example.}
\label{app:low_example}
We provide illustrative examples of images with varying degrees of colorfulness in Fig.~\ref{fig:lowlevel_examples_row3}, along with corresponding color metric values—specifically Hasler's colorfulness metric (\texttt{CF\_Hasler}) and color entropy (\texttt{Entropy\_bits}).
We further provide representative color congruence and brightness score examples in Tab.~\ref{tab:rep_congruence_examples}.

\begin{table}[htbp] 
\centering 
\small
\begin{tabular}{lrr}
\toprule
Example Image & \texttt{CF\_Hasler} & \texttt{Entropy\_bits} \\
\midrule
(a) Colorfulness average & 49.37 & 6.38 \\
(b) Colorfulness high & 165.98 & 9.36 \\
(c) Colorfulness low & 1.85 & 0.95 \\ 
\bottomrule
\end{tabular}
\caption{Color metrics for the examples in Fig.~\ref{fig:lowlevel_examples_row3}.}
\label{tab:my_metrics} 
\end{table}

\clearpage

\input{latex/sections/app_cperceptual_table}

\clearpage

\subsection{Compositional Level}
\label{app:vpf_mid}
To analyze how spatial arrangement and visual hierarchy guide human attention, we utilized features based on saliency. For each image, a saliency map $S$ was extracted using the pre-trained DeepGaze IIE~\cite{DeepGazeIIE}. In addition to saliency, we construct Message-Relevant Regions (MRR) using Grounded-SAM-2~\cite{grounded-sam}. We convert each message into an open-vocabulary prompt and detect candidate regions with GroundingDINO (box threshold 0.40, text threshold 0.25), followed by mask refinement using SAM2 (hiera-large). The resulting segmentation masks are used as MRR. For cases where no regions are detected under the initial prompt, we perform a fallback by re-querying with noun-based prompts extracted from the message (up to four terms) with relaxed detection thresholds (box threshold 0.25, text threshold 0.20). In the initial step, detections are obtained for $n_{\text{high}}=130$ and $n_{\text{low}}=247$ samples; after applying the fallback to non-detected cases, coverage increases to $n_{\text{high}}=178$ and $n_{\text{low}}=360$.

The following equations define the metrics:

\begin{itemize}
[leftmargin=*, itemsep=0pt, topsep=2pt, parsep=0pt]

    \item \textbf{Center Bias Index (CBI):} The proportion of saliency mass within a central disk. Let the image size be $H{\times}W$, the center $c{=}\big(\tfrac{H-1}{2},\,\tfrac{W-1}{2}\big)$, and the radius
    \(
      r_c \;=\; \alpha \,\min(H,W)
    \)
    with $\alpha{=}0.20$.
    \begin{gather}
        \mathrm{CBI} = \sum_{(i,j) \in D(c,r_c)} S_{i,j} \\
        \text{where } D(c,r) = \{ (i,j) : \|(i,j)-c\|_2 \le r \} \nonumber
    \end{gather}

    \item \textbf{T3:} The saliency mass within the union of four disks centered at the rule-of-thirds intersection points. Let
    \(
       \mathcal{P}
       = \{(\tfrac{H}{3},\tfrac{W}{3}),\,(\tfrac{H}{3},\tfrac{2W}{3}),\,
            (\tfrac{2H}{3},\tfrac{W}{3}),\,(\tfrac{2H}{3},\tfrac{2W}{3})\}
    \)
    and
    \(
      r_t \;=\; \beta\,\min(H,W)
    \)
    with $\beta{=}0.10$.
    \begin{equation}
        \mathrm{T3}
        \;=\;
        \sum_{(i,j)\in \bigcup_{p\in \mathcal{P}} D(p,r_t)} S_{i,j}.
    \end{equation}
\end{itemize}

\begin{table}[H]
\centering
\small
\begin{tabular}{lcc|cc}
\toprule
 & \multicolumn{2}{c|}{\textbf{CBI (Saliency, \%)}} & \multicolumn{2}{c}{\textbf{T3 (Saliency, \%)}} \\
\cmidrule(lr){2-3} \cmidrule(lr){4-5}
 & High & Low & High & Low \\
\midrule
\raisebox{-0.7ex}{\includegraphics[width=1.3em]{icons/audience.png}} Human
& 34.93 & 32.28 & 20.64 & 17.45 \\
\midrule
\raisebox{-0.7ex}{\includegraphics[width=1.3em]{icons/gpt.png}} GPT-5
& 34.20 & 30.93 & 19.32 & 16.75 \\
\raisebox{-0.7ex}{\includegraphics[width=1.3em]{icons/gpt.png}} GPT-5-mini
& 33.71 & 30.82 & 19.01 & 16.36 \\
\raisebox{-0.7ex}{\includegraphics[width=1.3em]{icons/gemma-color.png}} Gemma3
& 34.40 & 27.39 & 19.38 & 14.41 \\
\raisebox{-0.7ex}{\includegraphics[width=1.3em]{icons/qwen-color.png}} Qwen2.5
& 33.68 & 27.53 & 18.97 & 13.49 \\
\bottomrule
\end{tabular}
\caption{Raw compositional scores (saliency-based).}
\label{tab:comp_saliency}
\end{table}

\begin{table}[H]
\centering
\small
\begin{tabular}{lcc|cc}
\toprule
 & \multicolumn{2}{c|}{\textbf{CBI (MRR, \%)}} & \multicolumn{2}{c}{\textbf{T3 (MRR, \%)}} \\
\cmidrule(lr){2-3} \cmidrule(lr){4-5}
 & High & Low & High & Low \\
\midrule
\raisebox{-0.7ex}{\includegraphics[width=1.3em]{icons/audience.png}} Human
& 17.27 & 20.10 & 16.81 & 13.49 \\
\midrule
\raisebox{-0.7ex}{\includegraphics[width=1.3em]{icons/gpt.png}} GPT-5
& 19.98 & 17.53 & 15.00 & 13.77 \\
\raisebox{-0.7ex}{\includegraphics[width=1.3em]{icons/gpt.png}} GPT-5-mini
& 19.46 & 18.00 & 14.73 & 14.06 \\
\raisebox{-0.7ex}{\includegraphics[width=1.3em]{icons/gemma-color.png}} Gemma3
& 20.06 & 15.33 & 15.12 & 12.34 \\
\raisebox{-0.7ex}{\includegraphics[width=1.3em]{icons/qwen-color.png}} Qwen2.5
& 19.31 & 17.71 & 15.03 & 10.30 \\
\bottomrule
\end{tabular}
\caption{Raw compositional scores (MRR-based).}
\label{tab:comp_msr}
\end{table}

\begin{figure*}[t]
  \centering

  \begin{subfigure}[t]{0.30\textwidth}
    \centering
    \includegraphics[width=\linewidth]{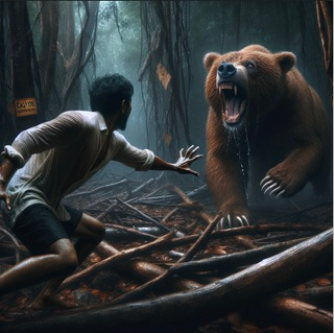}
    \vspace{2pt}
    \includegraphics[width=\linewidth]{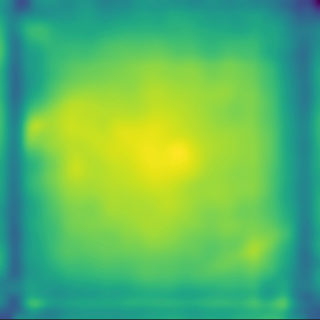}
    \caption*{(a) Example 1}
  \end{subfigure}\hfill
  %
  \begin{subfigure}[t]{0.30\textwidth}
    \centering
    \includegraphics[width=\linewidth]{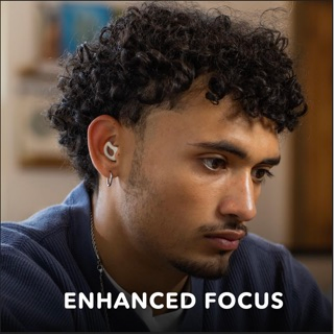}
    \vspace{2pt}
    \includegraphics[width=\linewidth]{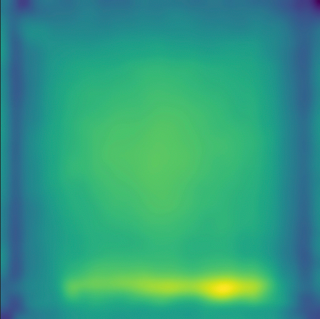}
    \caption*{(b) Example 2}
  \end{subfigure}\hfill
  %
  \begin{subfigure}[t]{0.34\textwidth}
    \centering
    \includegraphics[width=\linewidth]{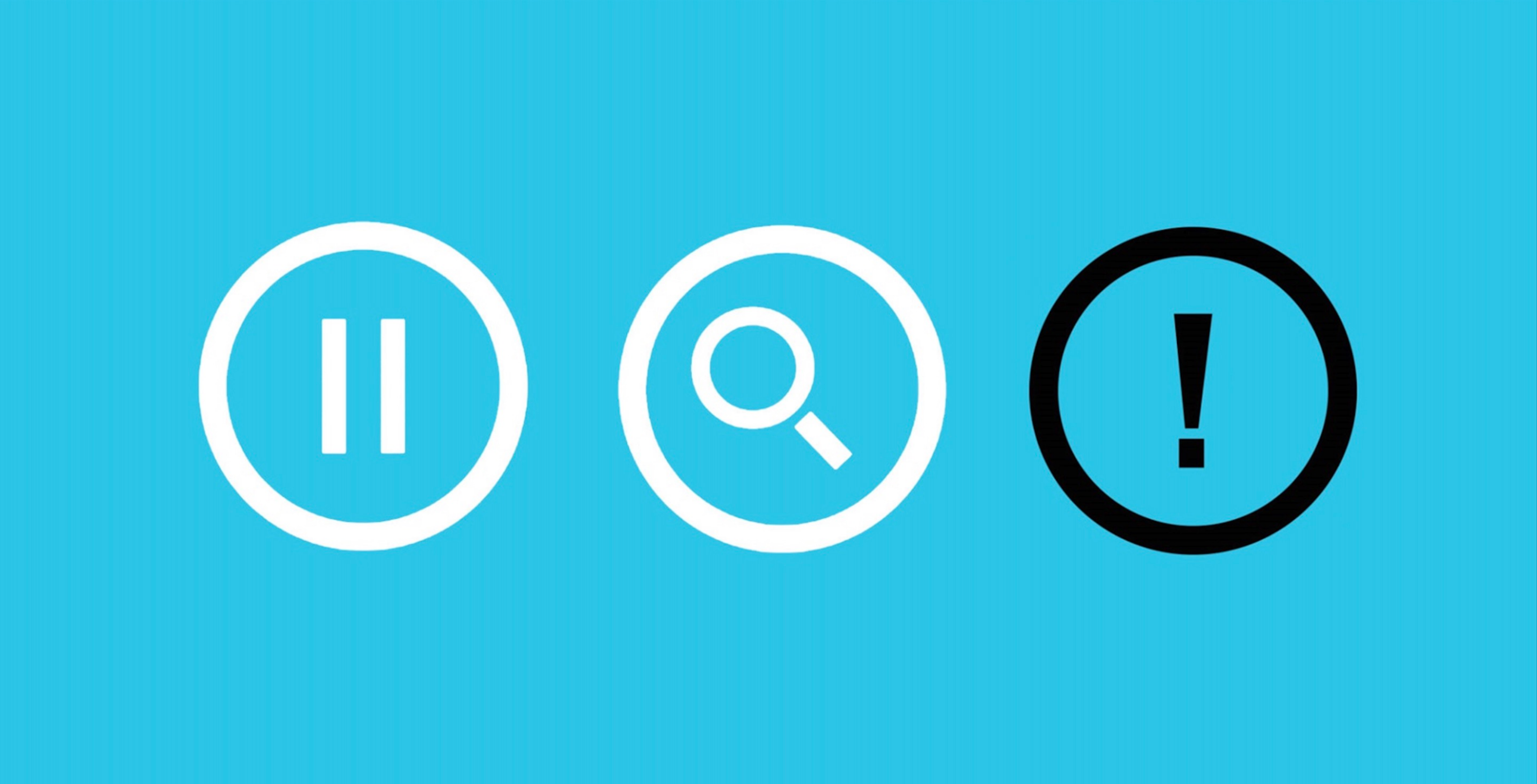}
    \vspace{2pt}
    \includegraphics[width=\linewidth]{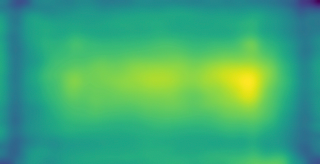}
    \caption*{(c) Example 3}
  \end{subfigure}

  \caption{Illustrative examples used in Appendix~\ref{app:vpf_mid}. 
  Each example (top) is shown with its corresponding saliency map (bottom).}
  \label{fig:midlevel_examples_row3}
\end{figure*}

\begin{figure*}[t]
  \centering

  \begin{minipage}[t]{0.32\linewidth}
    \includegraphics[width=\linewidth]{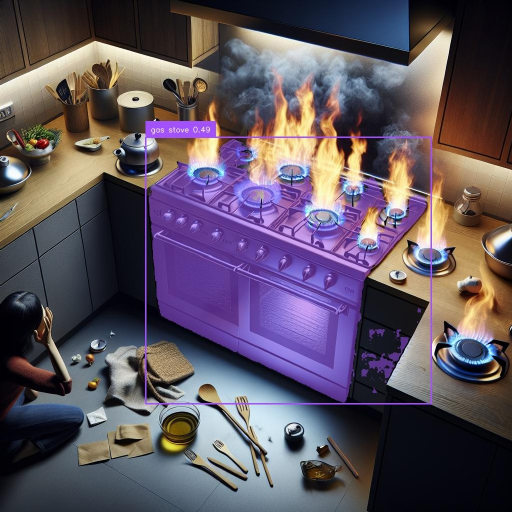}
    \captionof*{figure}{\scriptsize (a) Gas stove}
  \end{minipage}\hfill
  \begin{minipage}[t]{0.32\linewidth}
    \includegraphics[width=\linewidth]{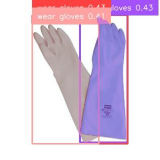}
    \captionof*{figure}{\scriptsize (b) Gloves}
  \end{minipage}\hfill
  \begin{minipage}[t]{0.32\linewidth}
    \includegraphics[width=\linewidth]{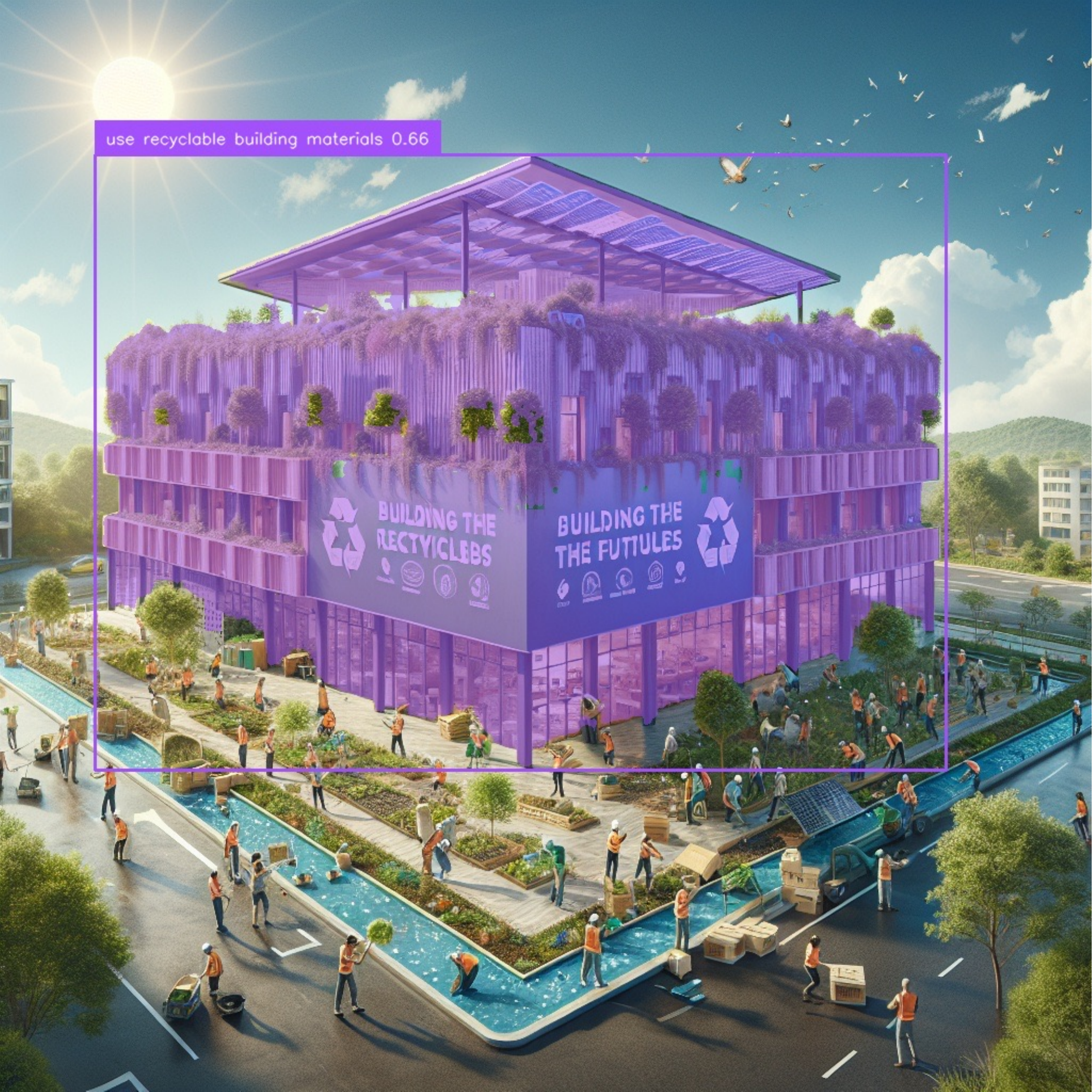}
    \captionof*{figure}{\scriptsize (c) Use recyclable building materials}
  \end{minipage}

  \vspace{4pt}
  \caption{
  Illustrative examples of message-related region extraction used for compositional-level analysis.
  Highlighted regions correspond to message-relevant segmented objects or areas used to compute MRR-based composition metrics.
  }
  \label{fig:msr_examples_row3}
\end{figure*}

\paragraph{Extraction Example.}
Fig.~\ref{fig:midlevel_examples_row3} presents three representative compositional-level examples, each showing an original image (top) and its corresponding feature map (bottom). These illustrate how the model captures compositional cues—specifically saliency prediction. The high density in the saliency map means that humans will pay more attention to it.
In addition, Fig.~\ref{fig:msr_examples_row3} shows examples of message-related region (MRR) extraction. 
The highlighted regions indicate segmented areas corresponding to message-relevant objects or regions, which are used to compute message-related region statistics such as center bias and rule-of-thirds overlap.

\clearpage

\clearpage

\subsection{Semantic Level}
\label{app:high_level_details}

\begin{figure*}[t]
    \centering
    \includegraphics[width=\linewidth]{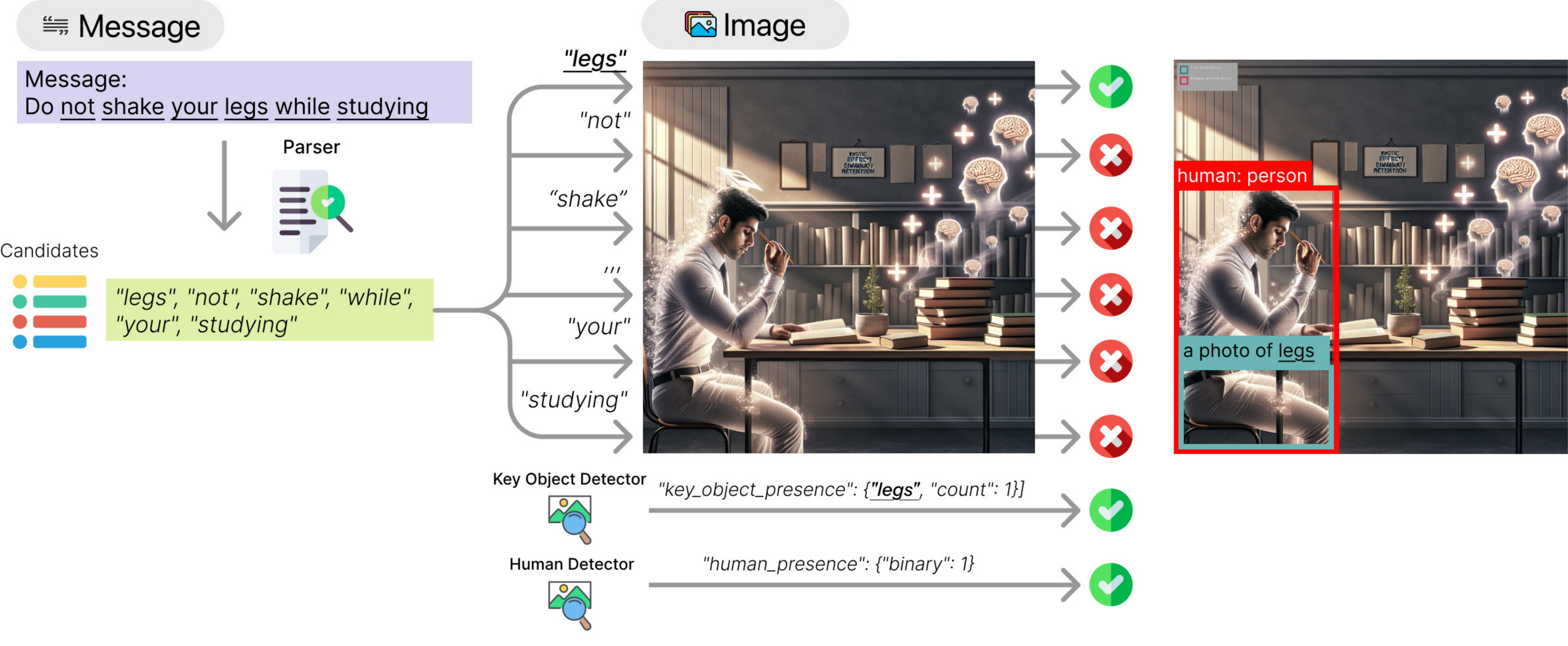}
    \caption{Overview of the semantic-level feature extraction pipeline. The pipeline detects whether semantically relevant key objects mentioned in the message are visually present in the image and identifies human presence through dedicated detectors.
    }
    \label{fig:semantic_pipeline}
\end{figure*}

\paragraph{Models Used.} These components collectively allow us to detect whether semantically referenced entities are visually grounded in the image.
Fig.~\ref{fig:semantic_pipeline} summarizes the extraction pipeline for key object and human presence detection.

\begin{itemize}
[leftmargin=*, itemsep=0pt, topsep=2pt, parsep=0pt]
    \item \textbf{Key Noun Extraction:} We used the \texttt{en\_core\_web\_sm} model as the parser from the \texttt{spaCy} library for Part-of-Speech (POS) tagging and noun phrase extraction. This lightweight model was selected for its balance between processing speed and linguistic accuracy.
    
    \item \textbf{Key Object Presence Detection:} We employed the OWL-ViT v1~\cite{owlvit_v1} model. OWL-ViT enables zero-shot object detection by aligning vision and text embeddings, allowing us to flexibly identify key visual concepts based on extracted textual nouns.
    
    \item \textbf{Human Presence Detection:} We used the pre-trained YOLOv8-nano~\cite{yolov8_ultralytics} as the primary person detector, optimized for inference efficiency on GPU. In cases where no human instance was detected, we optionally cross-validated results using the open-vocabulary detector above.

    \item \textbf{Text Presence Detection:}
    We identify text presence using a lightweight heuristic based on the dataset-provided image descriptions. Specifically, we extract candidate textual content from quoted spans (\emph{e.g.}, “...”, ‘...’) and ALL-CAPS phrases that resemble signage or logos, based on the observation that such patterns often correspond to explicitly rendered text in the associated images. This rule-based parsing yields 144 candidate instances out of 562 samples. Given the relatively small number, we manually verify each case to confirm whether the extracted text is visibly present in the image, resulting in 124 validated instances (22.06\%).
\end{itemize}

\paragraph{Metrics and Analytical Procedure.}
After extracting binary indicators of key object presence ($x_{\mathrm{obj}} \in \{0,1\}$), human presence ($x_{\mathrm{hum}} \in \{0,1\}$) and text presence ($x_{\mathrm{text}} \in \{0,1\}$) for each image, we constructed a feature matrix 
\[
X = 
\begin{bmatrix}
1 & x_{\mathrm{obj}}^{(1)} & x_{\mathrm{hum}}^{(1)} & x_{\mathrm{text}}^{(1)} \\
1 & x_{\mathrm{obj}}^{(2)} & x_{\mathrm{hum}}^{(2)} & x_{\mathrm{text}}^{(2)} \\
\vdots & \vdots & \vdots & \vdots \\
1 & x_{\mathrm{obj}}^{(n)} & x_{\mathrm{hum}}^{(n)} & x_{\mathrm{text}}^{(n)}
\end{bmatrix},
\]
where the intercept term allows for estimation of the baseline log-odds.

For each outcome variable $y \in \{0,1\}$ (\emph{e.g.}, perceived persuasiveness or GPT agreement), we fit a logistic regression model of the form:
\[
\begin{aligned}
\Pr(y=1 \mid X)
= \sigma(&\beta_0 + \beta_1 x_{\mathrm{obj}} \\
&+ \beta_2 x_{\mathrm{hum}} + \beta_3 x_{\mathrm{text}})
\end{aligned}
\]
where $\sigma(z)=\frac{1}{1+e^{-z}}$ denotes the logistic link function.
Estimated coefficients $\hat{\beta}$ were exponentiated to yield odds ratios:
\[
\mathrm{OR}_i = e^{\hat{\beta}_i},
\]
representing multiplicative changes in the odds of $y=1$ given the presence of each feature.  
Robust standard errors (HC3 covariance) were used to account for potential heteroskedasticity, and all $p$-values were adjusted for multiple comparisons using the Benjamini--Hochberg false discovery rate (FDR).

This procedure allows us to quantify how semantic alignment signals—\emph{i.e.}, the co-occurrence of mentioned and visually realized entities—contribute to persuasiveness judgments and model–human agreement.

\begin{figure*}[t]
  \centering

  \begin{subfigure}[t]{0.485\textwidth}
    \centering
    \includegraphics[width=\linewidth,height=4.2cm,keepaspectratio]{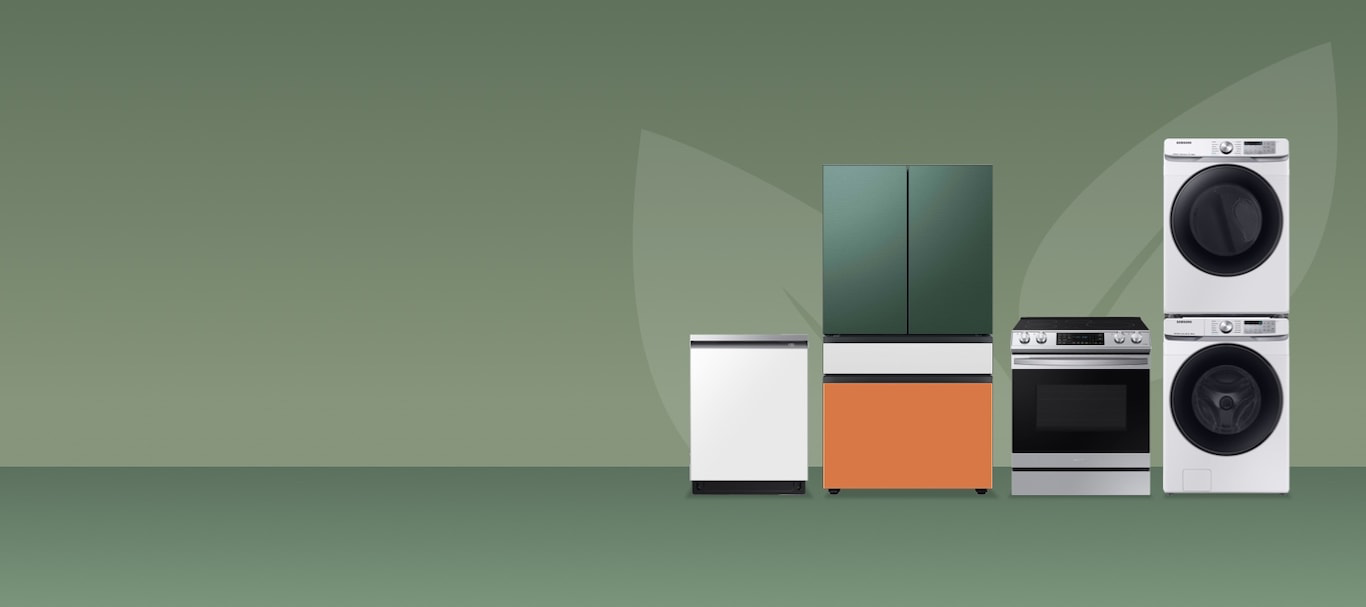}
    \caption*{(a) Example A — original}
  \end{subfigure}\hfill
  \begin{subfigure}[t]{0.485\textwidth}
    \centering
    \includegraphics[width=\linewidth,height=4.2cm,keepaspectratio]{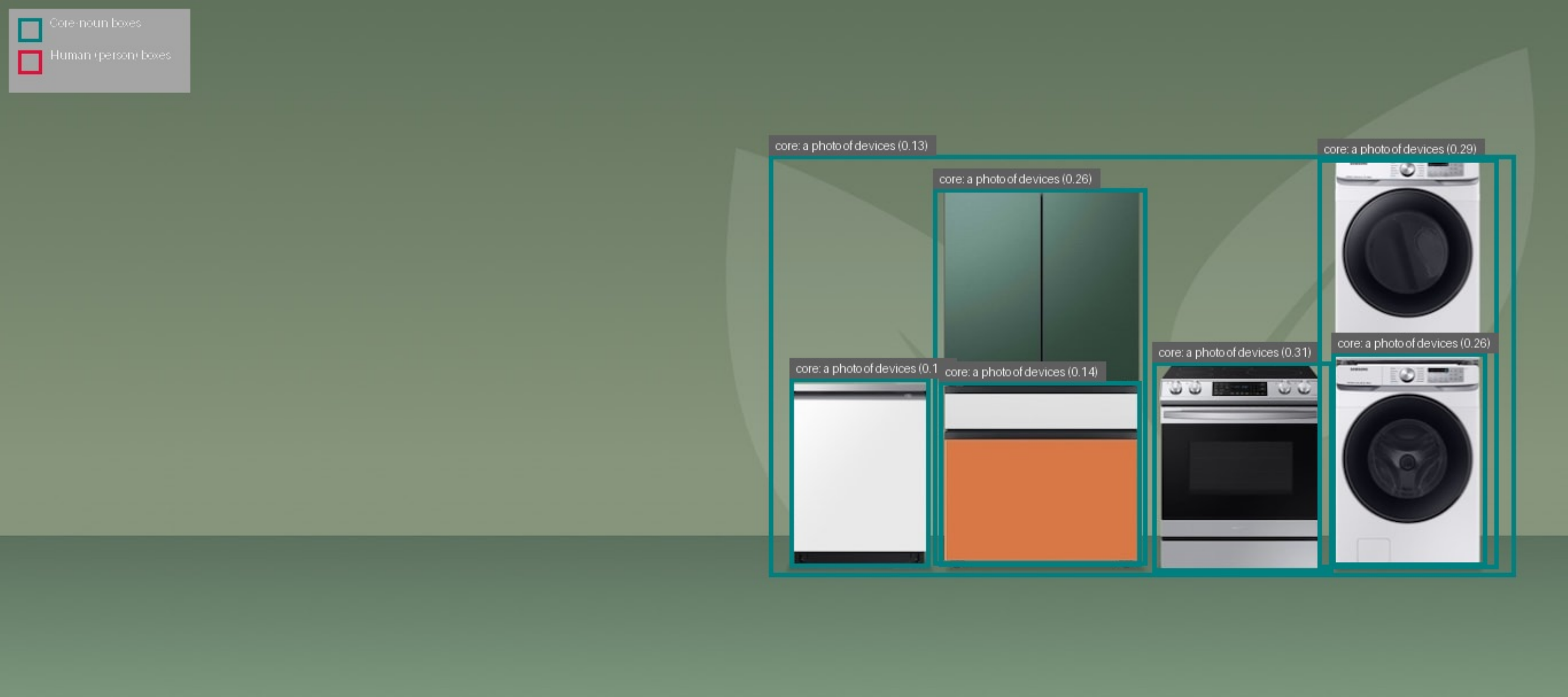}
    \caption*{(b) Example A — detection}
  \end{subfigure}

  \vspace{4pt}

  \begin{subfigure}[t]{0.485\textwidth}
    \centering
    \includegraphics[width=\linewidth,height=4.2cm,keepaspectratio]{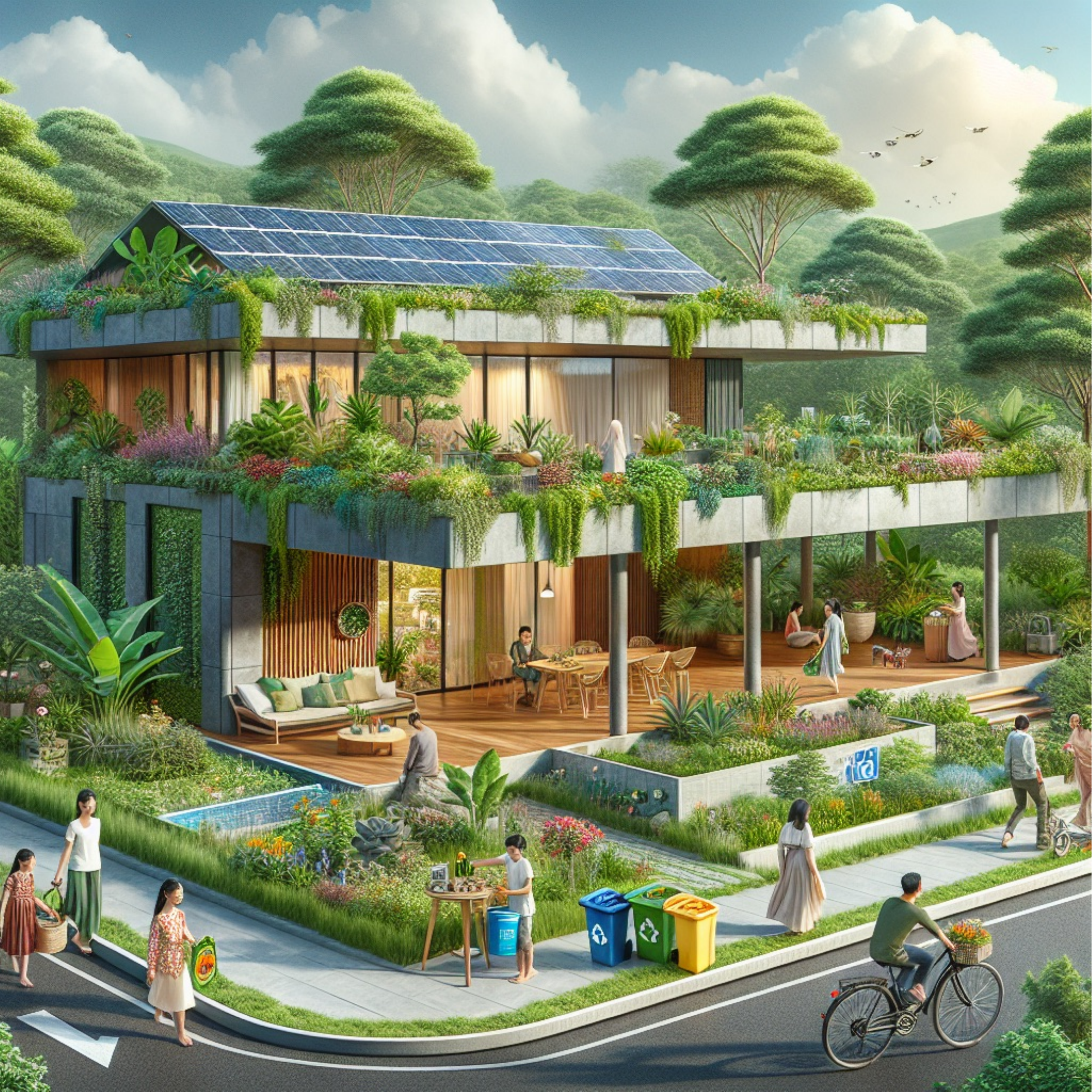}
    \caption*{(c) Example B — original}
  \end{subfigure}\hfill
  \begin{subfigure}[t]{0.485\textwidth}
    \centering
    \includegraphics[width=\linewidth,height=4.2cm,keepaspectratio]{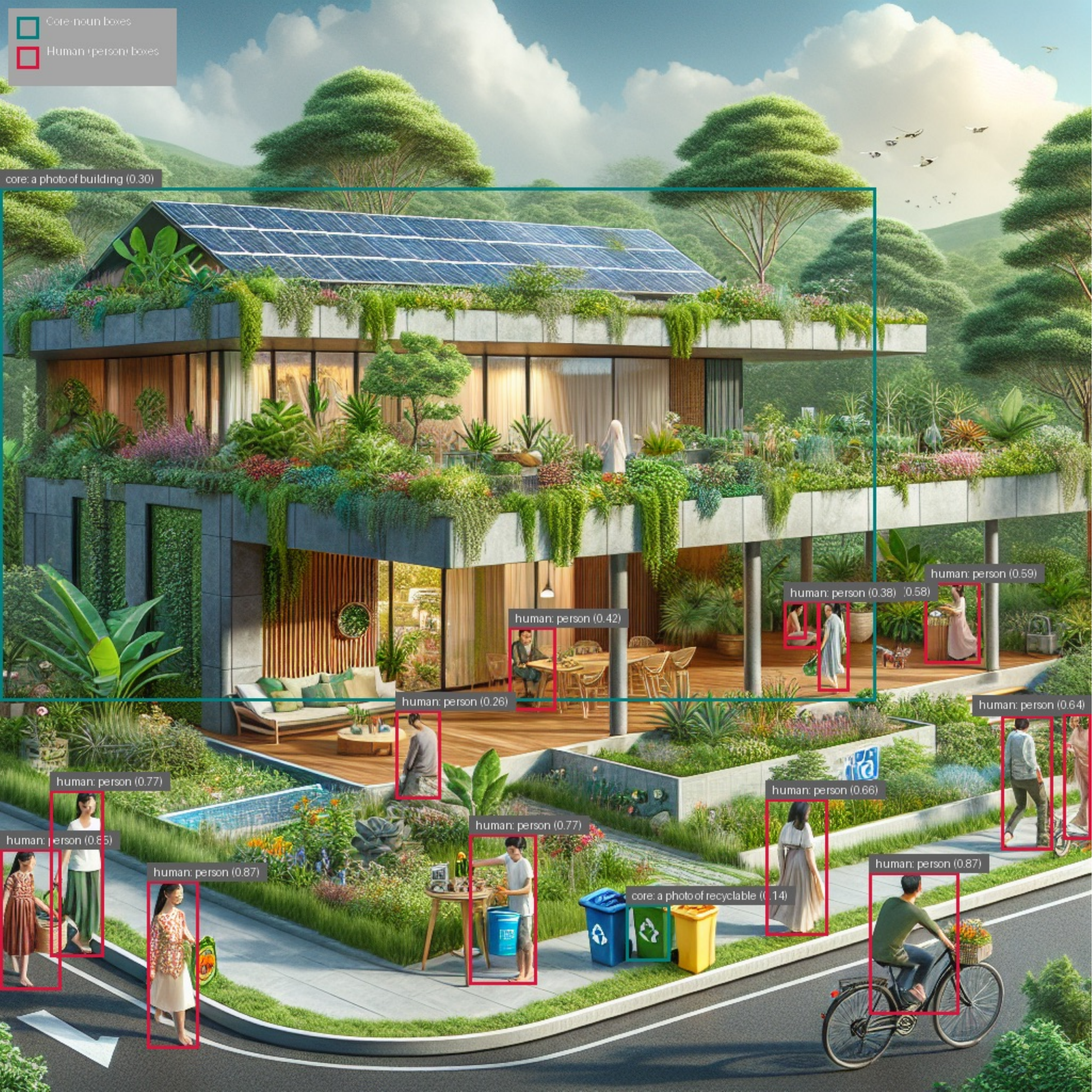}
    \caption*{(d) Example B — detection}
  \end{subfigure}

  \caption{Key object presence examples: each row shows a pair (original vs. object-detection overlay).}
  \label{fig:highlevel_pairs}
\end{figure*}

\paragraph{Extraction Example.}
To illustrate how semantic-level factors are identified and mapped in our framework, 
Fig.~\ref{fig:semantic_pipeline} presents the extraction pipeline and Fig.~\ref{fig:highlevel_pairs} presents two representative examples 
showing the original images and their corresponding object-detection overlays. 
The extracted results for key objects and text cues are detailed in Tab.~\ref{tab:keyobj_extraction_examples} and Tab.~\ref{tab:relevant_text_extraction_examples}, respectively.

\begin{table}[t]
\centering
\small
\renewcommand{\arraystretch}{1.12}
\setlength{\tabcolsep}{3pt}
\begin{tabularx}{\linewidth}{
  >{\raggedright\arraybackslash}X
  >{\centering\arraybackslash}p{0.24\linewidth}
  >{\centering\arraybackslash}p{0.11\linewidth}
}
\toprule
\textbf{Message} & \textbf{Extracted KeyObj} & \textbf{Count} \\
\midrule
Be careful when using a knife & \texttt{knife} & 2 \\
Consume enough protein & \texttt{protein} & 2 \\
Wear gloves when handling chemicals & \texttt{gloves} & 1 \\
Wear a leather jacket to look stylish & \texttt{jacket} & 2 \\
Look up videos of military dog training & \texttt{dog} & 2 \\
Throw trash in the trash can & \texttt{trash} & 1 \\
\bottomrule
\end{tabularx}
\caption{
Illustrative examples of key-object extraction.
The extracted key object is taken from \texttt{core\_noun\_presence.detail.matches}; count denotes the number of detected matching regions.
}
\label{tab:keyobj_extraction_examples}
\end{table}

\clearpage

\begin{table*}[t]
\centering
\footnotesize
\renewcommand{\arraystretch}{1.05}
\setlength{\tabcolsep}{4pt}
\begin{tabularx}{\linewidth}{
  >{\centering\arraybackslash}m{0.16\linewidth}
  >{\raggedright\arraybackslash}X
  >{\raggedright\arraybackslash}m{0.27\linewidth}
}
\toprule
\textbf{Image} & \textbf{Text Mention in Image Description} & \textbf{Extracted Text} \\
\midrule

\includegraphics[width=\linewidth]{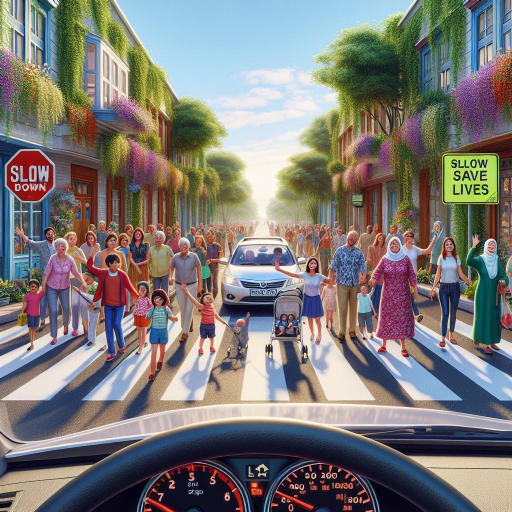}
&
\ldots signs along the crosswalk: one on the left reading ``SLOW DOWN'' and another on the right reading ``SLOW SAVE LIVES.'' \ldots
&
\texttt{SLOW DOWN}; \texttt{SLOW SAVE LIVES}
\\
\midrule

\includegraphics[width=\linewidth]{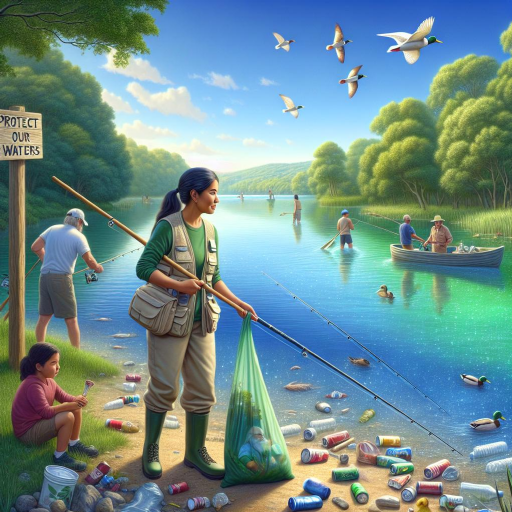}
&
\ldots A sign to the left reads ``PROTECT OUR WATERS,'' emphasizing the environmental conservation theme. \ldots
&
\texttt{PROTECT OUR WATERS}
\\
\midrule

\includegraphics[width=\linewidth]{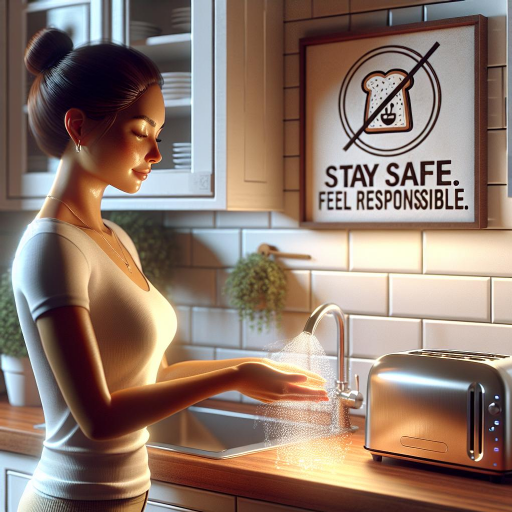}
&
\ldots above the toaster, there is a framed sign accompanied by the text ``STAY SAFE. FEEL RESPONSIBLE.'' \ldots
&
\texttt{STAY SAFE. FEEL RESPONSIBLE}
\\
\midrule

\includegraphics[width=\linewidth]{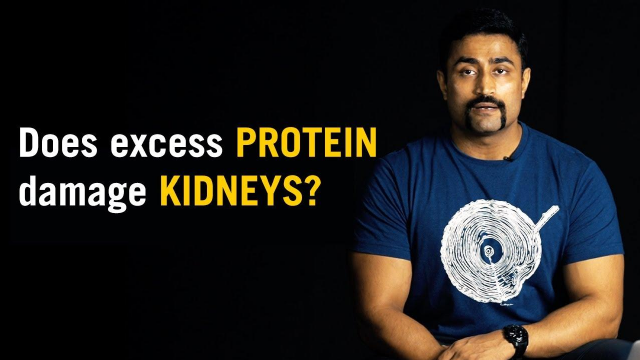}
&
\ldots the text on the left side of the image reads, ``Does excess PROTEIN damage KIDNEYS?'' \ldots
&
\texttt{Does excess PROTEIN damage KIDNEYS?}
\\

\bottomrule
\end{tabularx}
\caption{
Illustrative examples of relevant text extraction.
The extracted text corresponds to visible slogan-like or message-relevant text in the image, rather than incidental brand or location names.
}
\label{tab:relevant_text_extraction_examples}
\end{table*}

\clearpage
\section{Validation Survey Details}
\label{app:human_survey_details}
\paragraph{Annotators.}
We recruited adult annotators via Amazon Mechanical Turk (MTurk)\footnote{https://www.mturk.com} and conducted the survey through a custom web-based interface. 
The HIT title warned workers that the task could contain adult or sensitive content, and the task description described it as an academic study on message--image analysis and visual persuasion. 
Workers were required to have more than 10{,}000 approved HITs and an approval rate above 98\%. 
Each HIT contained 12 image-message items and was designed to take approximately 12--15 minutes based on repeated pilot runs of the task interface. 
Workers were compensated at \$2 per HIT, corresponding to an estimated hourly rate of approximately \$8--\$10. 
We did not collect protected demographic attributes beyond platform-provided worker identifiers used for task administration and quality control. 
The entire data collection process was conducted anonymously to protect participants' privacy; no personally identifiable information (PII) was collected or stored.
In total, 476 participants were collected, of which 141 were retained after quality filtering.

\begin{figure*}[p]
    \centering
    \includegraphics[page=1,width=\textwidth,height=0.93\textheight,keepaspectratio]{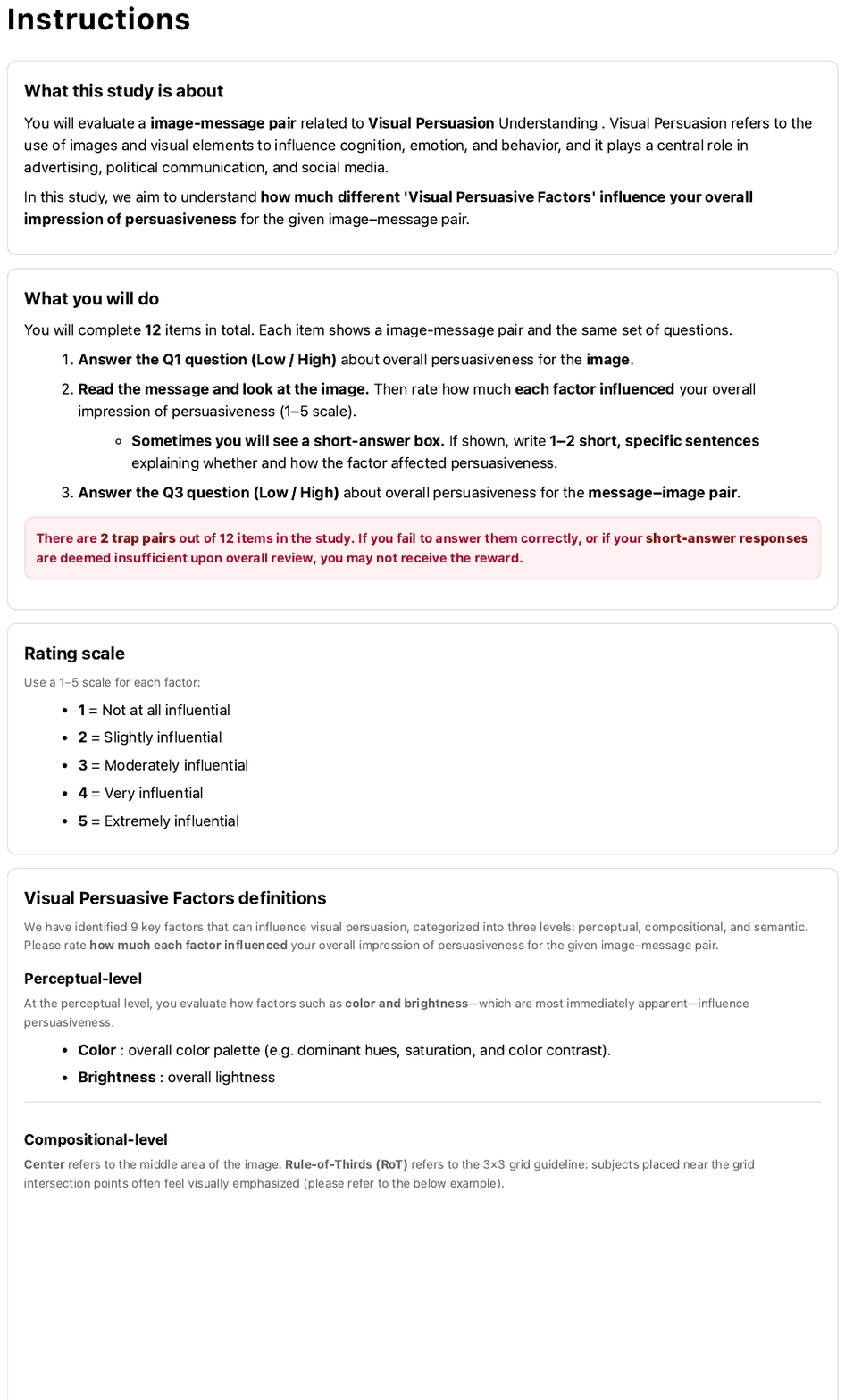}
    \caption{Instruction page 1 of the human survey interface.}
    \label{fig:survey_instruction_page1}
\end{figure*}

\begin{figure*}[p]
    \centering
    \includegraphics[page=1,width=\textwidth,height=0.93\textheight,keepaspectratio]{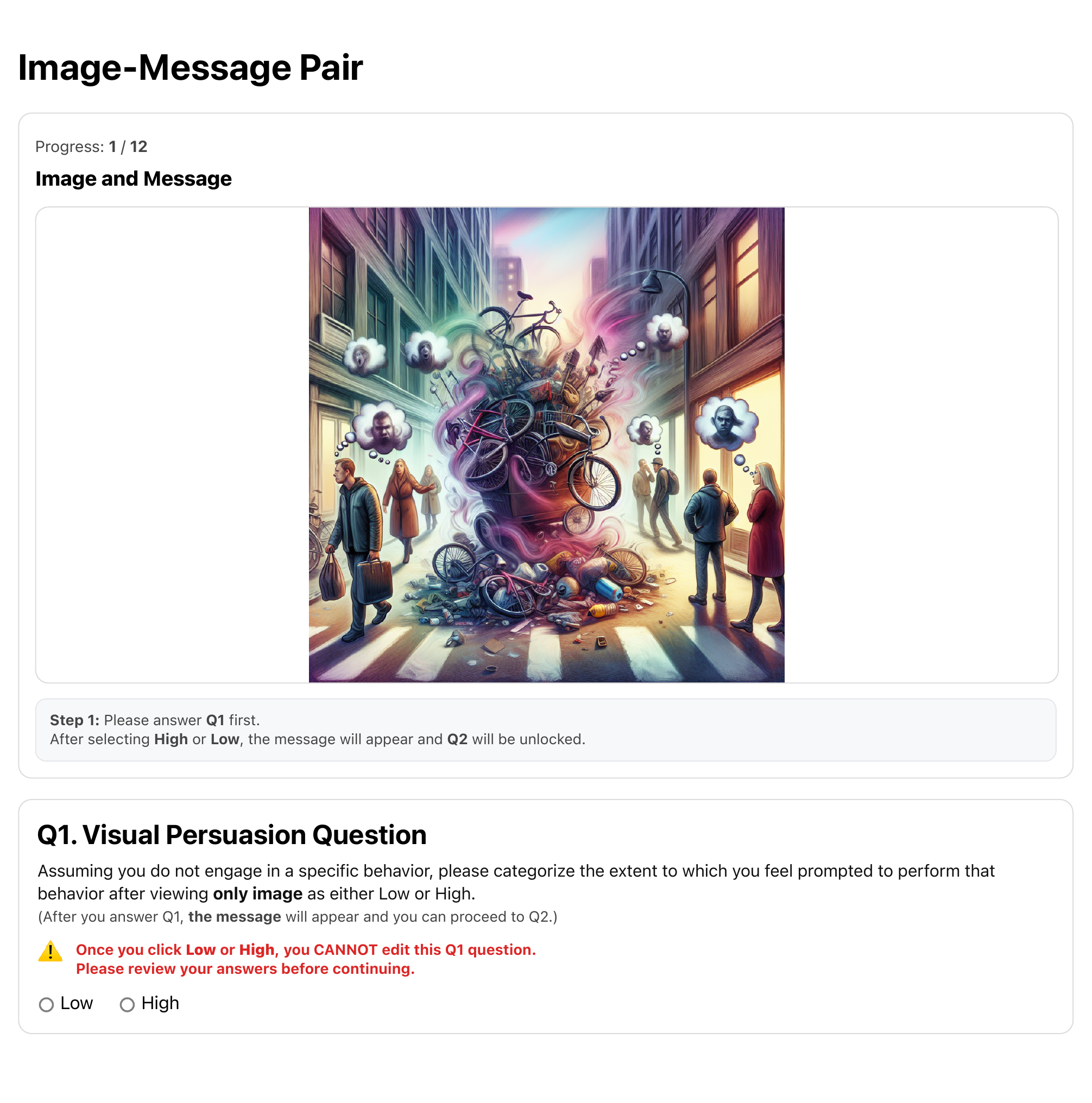}
    \caption{Step 1 of the human survey interface. Annotators first provide an image-only binary persuasiveness judgment before the message and factor-rating questions are revealed.}
    \label{fig:human_survey_task1}
\end{figure*}

\begin{figure*}[p]
    \centering
    \includegraphics[page=1,width=\textwidth,height=0.93\textheight,keepaspectratio]{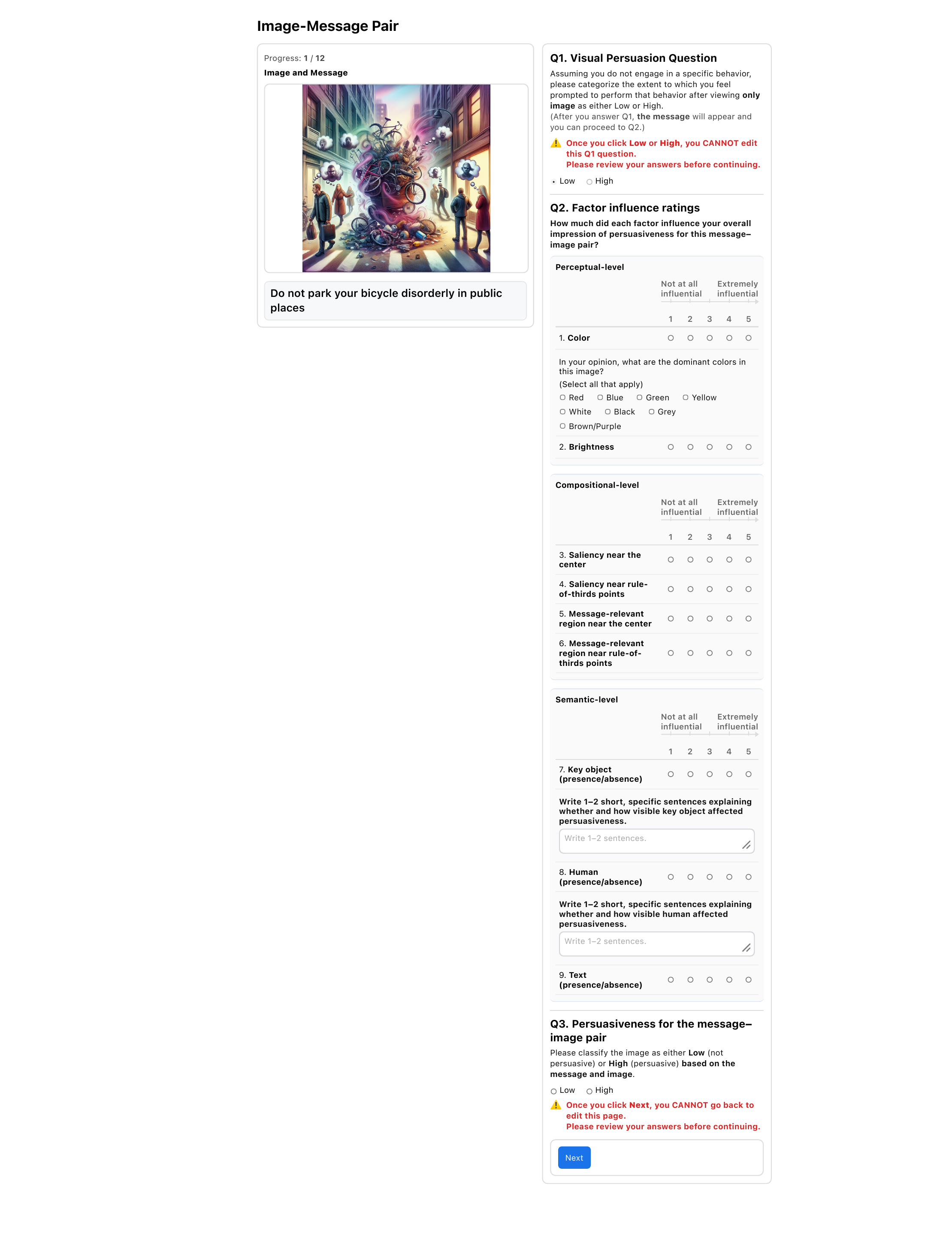}
    \caption{Step 2 of the human survey interface. After answering the image-only question, annotators rate factor-level influence, provide short explanations when applicable, and then judge persuasiveness for the full image-message pair.}
    \label{fig:human_survey_task2}
\end{figure*}

\paragraph{Task Design.}
Each trial consisted of an image-only stage followed by an image-message stage. Participants were first asked to provide a binary persuasiveness judgment (Low/High) based only on the image, using a slightly adapted version of the original dataset prompt to match the binary format. This question was included to assess alignment with the original dataset judgments, yielding approximately 80\% consistency. The instruction pages shown to annotators are provided in Fig.~\ref{fig:survey_instruction_page1} and~\ref{fig:survey_instruction_page2}, while the two-stage task interface is illustrated in Fig.~\ref{fig:human_survey_task1} and~\ref{fig:human_survey_task2}.

Subsequently, annotators rated the influence of each Visual Persuasive Factor (VPF) on a 5-point Likert scale (1: not at all influential, 2: slightly influential, 3: moderately influential, 4: very influential, 5: extremely influential). For quality check, when semantic-level factors (\emph{e.g.}, key objects, humans, or text) were present, annotators were additionally asked to provide brief explanations (1--2 sentences) describing whether and how these elements affected their judgment. Finally, annotators provided a second binary persuasiveness judgment based on the full image-message pair.

\paragraph{Annotation Protocol.}
Each item was annotated by three independent participants. The full dataset of 562 image-message pairs was covered three times, resulting in triply annotated labels for all items. To prevent redundancy and learning effects, participants were not allowed to re-annotate items they had previously seen, even if they participated in multiple survey batches. As illustrated in Fig.~\ref{fig:human_survey_task1} and~\ref{fig:human_survey_task2}, the task followed a fixed sequence within each trial: (Q1) image-only persuasiveness judgment, followed by (Q2) factor-level ratings, and (Q3) image-message persuasiveness judgment.

\paragraph{Quality Control.}
We implemented strict quality control procedures to ensure annotation reliability. As described in the instruction interface (Fig.~\ref{fig:survey_instruction_page1} and~\ref{fig:survey_instruction_page2}), each task included 2 hidden attention-check (trap) questions among the 12 items; responses from annotators who failed both checks were excluded from analysis. Participants were informed of the presence of such checks but not their locations.

In addition, all free-text responses were manually reviewed by the authors, and annotations deemed uninformative or invalid were removed through a conservative filtering process. Annotators who failed attention checks or whose responses were rejected were prevented from re-participating in subsequent survey batches. Overall, this process reduced the dataset from 476 initial participants to 141 valid annotators, yielding a high-quality subset for analysis.

\begin{table}[t]
\centering
\small
\resizebox{\linewidth}{!}{%
\begin{tabular}{llc}
\toprule
Level & Factor & Human-rated Importance AUC \\
\midrule
Perceptual & Color & 0.724 \\
Perceptual & Brightness & 0.654 \\
\midrule
Compositional & CBI (Saliency) & 0.748 \\
Compositional & CBI (MRR) & 0.784 \\
Compositional & T3 (Saliency) & 0.744 \\
Compositional & T3 (MRR) & 0.805 \\
\midrule
Semantic & Key Object & 0.755 \\
Semantic & Human & 0.709 \\
Semantic & Text & 0.649 \\
\bottomrule
\end{tabular}%
}
\caption{
Validation survey results for the proposed visual persuasive factors.
AUC denotes rank-based separation between High and Low items in human-rated factor importance.
}
\label{tab:human_survey_auc}
\end{table}

\begin{figure}[t]
    \centering
    \includegraphics[width=\linewidth]{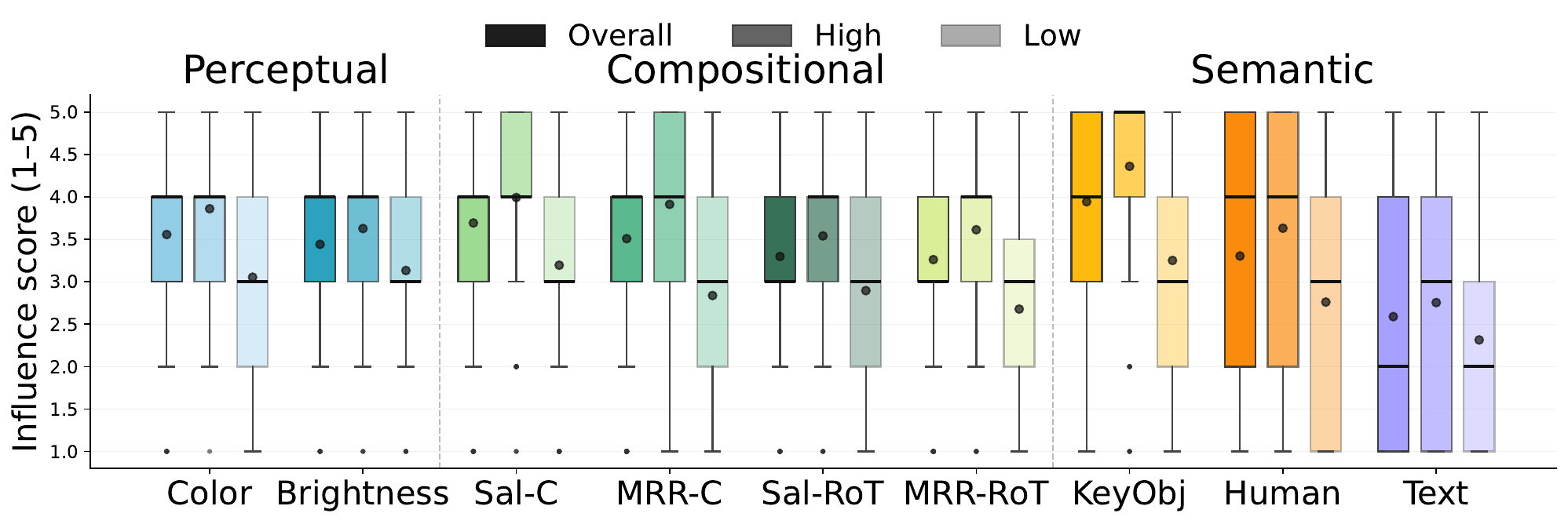}
    \caption{Factors influence ratings by 141 annotators. 
    }
    \label{fig:human_survey_toal}
\end{figure}

\paragraph{Result Details.}
As shown in Tab.~\ref{tab:human_survey_auc}, all nine factors show above-chance rank-based separation between High and Low items, with AUC values ranging from 0.649 to 0.805.
Fig.~\ref{fig:human_survey_toal} further visualizes the factor-wise rating distributions, showing consistently higher influence ratings for High than Low items.
Separately, under a substantially different annotation setting, new annotators' binary persuasiveness judgments align with the original GT labels at an average rate of 77.81\%, suggesting that the high-consensus labels remain reasonably stable.

\begin{figure*}[p]
    \centering
    \includegraphics[page=2,width=\textwidth,height=0.93\textheight,keepaspectratio]{image_example/human_survey_instructions.pdf}
    \caption{Instruction page 2 of the human survey interface.}
    \label{fig:survey_instruction_page2}
\end{figure*}

\clearpage

\section{Generalization Across Agreement Levels and Model Variants}
\label{app:generalization_results}

We further examine whether the main findings in \S\ref{sec:vpf_analysis} generalize beyond the original high-agreement subset and model set.
Specifically, we test progressively broader annotator-agreement levels and additional VLMs spanning different model scales, families, and reasoning variants.
Across these settings, we consistently observe the same three patterns: the recall-oriented bias, partial reproduction of human-aligned VPF patterns, and false-positive cue overgeneralization.

\subsection{Generalization Across Agreement Levels}
\label{app:agreement_generalization}

Our main analysis uses 562 image--message pairs with almost-perfect annotator agreement to reduce ambiguity in human persuasiveness judgments.
To examine whether the observed patterns depend on this high-agreement setting, we progressively broaden the evaluation set following the agreement levels (\emph{i.e.}, Tab.~\ref{tab:dataset_agreement}).
We consider (i) the original Almost Perfect subset ($N=562$), (ii) samples spanning Almost Perfect through Fair agreement ($N=834$), and (iii) samples spanning Almost Perfect through Slight agreement ($N=4{,}531$).
For the broader subsets, we retain items with a majority Low or High label and remove ties, thereby including substantially more diverse judgments while maintaining at least minimal annotator agreement.

\begin{table}[t]
\centering
\renewcommand{\arraystretch}{1.08}
\setlength{\tabcolsep}{4pt}

\begin{adjustbox}{max width=\columnwidth}
\begin{tabular}{llrrrr}
\toprule
\textbf{Agreement Range} & \textbf{Model}
& \textbf{Acc.} & \textbf{Prec.} & \textbf{Rec.} & \textbf{F1} \\
\midrule
\multirow{4}{*}{Almost Perfect $\sim$ Fair}
& GPT-5      & 67.51 & 47.19 & 96.71 & 63.43 \\
& GPT-5-mini & 61.03 & 42.68 & 98.35 & 59.53 \\
& Gemma3     & 66.79 & 46.43 & 90.95 & 61.47 \\
& Qwen2.5-VL & 53.48 & 37.44 & 88.89 & 52.68 \\
\midrule
\multirow{4}{*}{Almost Perfect $\sim$ Slight}
& GPT-5      & 60.67 & 46.01 & 93.99 & 61.78 \\
& GPT-5-mini & 53.96 & 42.03 & 95.30 & 58.33 \\
& Gemma3     & 59.19 & 44.71 & 87.47 & 59.17 \\
& Qwen2.5-VL & 49.44 & 39.03 & 88.19 & 54.12 \\
\bottomrule
\end{tabular}
\end{adjustbox}

\caption{
VLM performance across progressively broader annotator-agreement ranges.
Recall remains substantially higher than precision across all models as lower-agreement cases are included.
}
\label{tab:agreement_baseline}
\end{table}

\textbf{The recall-oriented bias persists across broader agreement levels.}
As shown in Tab.~\ref{tab:agreement_baseline}, all models continue to exhibit substantially higher recall than precision.
For the Almost Perfect--Fair subset, recall ranges from 88.89\% to 98.35\%, whereas precision remains between 37.44\% and 47.19\%.
The same pattern persists after expanding to 4,531 samples through Slight agreement, where recall remains 87.47--95.30\% compared with 39.03--46.01\% precision.
Thus, the tendency to over-predict High persuasiveness is not specific to the original high-consensus subset.

\input{latex/sections/appendix_agreement_vpfs_table}

\textbf{VPF-based findings likewise remain stable as agreement is relaxed.}
As shown in Tab.~\ref{tab:agreement_vpf_generalization}, VLM predictions continue to reproduce only part of the human-aligned VPF pattern.
Across both expanded agreement ranges, several persuasion-relevant factors---including color entropy, strategy-conditioned brightness, compositional saliency, and key object presence---remain associated with predicted High persuasiveness, although their strength and consistency vary across models.
This indicates that VLMs remain sensitive to meaningful persuasive cues without reproducing the full human factor pattern.

The false-positive analysis shows a similarly persistent pattern.
Among ground-truth Low items, many of the same persuasion-relevant cues are repeatedly associated with erroneous High predictions.
This includes perceptual, compositional, and semantic factors such as color entropy, strategy-conditioned brightness, saliency, key object presence, and human presence.

Overall, expanding the analysis from 562 high-consensus samples to as many as 4,531 samples does not qualitatively alter our main findings.
The recall-oriented bias, partial reproduction of human-aligned VPF patterns, and false-positive cue overgeneralization persist under progressively broader annotator agreement.

\subsection{Generalization Across Model Variants}
\label{app:model_generalization}

We next examine whether these findings are specific to the original model set.
We expand the evaluation along three dimensions: \textit{model scale}, using Gemma3-4B, Gemma3-12B, and Gemma3-27B; \textit{model family}, using InternVL3-8B-Instruct; and \textit{reasoning orientation}, comparing Qwen3-VL-8B-Instruct with Qwen3-VL-8B-Thinking.

\begin{table}[t]
\centering
\renewcommand{\arraystretch}{1.08}
\setlength{\tabcolsep}{4pt}

\begin{adjustbox}{max width=\columnwidth}
\begin{tabular}{lrrrr}
\toprule
\textbf{Model}
& \textbf{Acc.} & \textbf{Prec.} & \textbf{Rec.} & \textbf{F1} \\
\midrule
Gemma3-4B             & 60.32 & 43.28 & 81.46 & 56.53 \\
Gemma3-12B            & 64.77 & 47.16 & 93.26 & 62.64 \\
Gemma3-27B            & 63.88 & 46.33 & 88.76 & 60.89 \\
InternVL3-8B-Instruct & 61.39 & 43.93 & 79.21 & 56.51 \\
Qwen3-VL-8B-Instruct  & 68.15 & 49.84 & 87.08 & 63.39 \\
Qwen3-VL-8B-Thinking  & 61.18 & 44.91 & 96.63 & 61.32 \\
\bottomrule
\end{tabular}
\end{adjustbox}

\caption{
Performance across additional VLM scales, families, and reasoning variants.
The recall--precision gap persists across all evaluated model settings.
}
\label{tab:model_variant_baseline}
\end{table}

\textbf{The recall-oriented bias generalizes across model scale, family, and reasoning orientation.}
As shown in Tab.~\ref{tab:model_variant_baseline}, every additional model exhibits substantially higher recall than precision.
The pattern remains across the Gemma3 scale series and InternVL3, indicating that it is not specific to a particular model size or family.
It also persists across Qwen3-VL reasoning variants: Qwen3-VL-8B-Thinking reaches 96.63\% recall but only 44.91\% precision.
Thus, increased scale or explicit reasoning orientation alone does not eliminate the tendency to over-predict persuasiveness.

\input{latex/sections/appendix_diverse_model_table}

\textbf{The VPF-based diagnosis also persists across diverse model variants.}
As shown in Tab.~\ref{tab:model_variant_vpf_generalization}, the expanded models continue to partially reproduce the human-aligned VPF pattern.
Several cues associated with human persuasiveness judgments, such as representative color congruence, strategy-conditioned brightness, compositional saliency, and key object presence, also recur across VLM predictions.
However, their relative strength varies across model scales and families, and some cues exhibit associations that are weak or absent in human judgments.
This again suggests that VLMs capture meaningful components of the human factor structure but do not consistently weight or contextualize them in the same way.

False-positive cue overgeneralization likewise appears across the expanded model set.
Although the specific factors differ by model, persuasion-relevant perceptual, compositional, and semantic cues repeatedly distinguish false positives from true negatives.
In particular, cues such as strategy-conditioned brightness, saliency, key object presence, human presence, and text presence recur among erroneous High predictions.
The exact VPF profile is therefore model-dependent, but the broader failure mode remains consistent: VLMs tend to treat plausible persuasive cues as stronger evidence of persuasiveness than warranted by the human judgment.

Taken together, these results show that the central findings of our VPF analysis are not restricted to a particular annotator-agreement threshold or model configuration.
Across progressively broader agreement levels and diverse model scales, families, and reasoning variants, the recall-oriented bias, partial reproduction of human-aligned VPF patterns, and false-positive cue overgeneralization persist.

\section{Experiment Details}
\label{app:pr_sec}
All experiments were conducted on a single NVIDIA A6000 GPU. For computational efficiency, all models were loaded using 4-bit quantization, and most experimental single runs (\emph{i.e.,} one seed) completed within four hours. For inference, we employed the default generation configurations provided by each model's respective library; for the GPT models, no additional generation settings were specified beyond the standard API defaults. Detailed model versions and licenses are provided in Tab.~\ref{tab:model_version}.

\subsection{Binary Judgment}
We use the following baseline prompt for the binary persuasiveness decision.
The model receives one image (via the VLM input) and a text message \texttt{\{message\}}.
It must output \texttt{[[yes]]} if the image is persuasive for the given message, or \texttt{[[no]]} otherwise—no other text or reasoning is allowed.

\begin{Prompt}[caption={Baseline (binary judgment).},label={pr:binary}]
System Instruction: 
"You are an AI assistant with expertise in analyzing and evaluating the persuasiveness of images and messages based on general principles of communication and psychology."

User Prompt: 
"""
Below is an instruction that describes a task, paired with an input that provides further context. Write a response that appropriately completes the request.

### Instruction:
You will perform a task where you predict how persuasive certain individuals will find an image created from a message. Please classify the image as either persuasive or not persuasive based on the message and image. Respond with a single word: 'yes' if persuasive, 'no' if not persuasive.

### Input:
Message: {message}

please directly output your answer by strictly following this format: [[answer]], for example: [[yes]]

### Response:
"""
\end{Prompt}

\begin{table*}[t]
\centering
\resizebox{0.95\textwidth}{!}{
\begin{tabular}{l l l l}
\toprule
\textbf{Model} & \textbf{Version} & \textbf{License} & \textbf{Reasoning Effort} \\
\midrule
\raisebox{-0.6ex}{\includegraphics[width=1.3em]{icons/gpt.png}}\ GPT-5~\cite{openai2025gpt5} 
& \texttt{gpt-5-2025-08-07} 
& Proprietary & \makecell[l]{minimal (medium for KCC,\\ low for rationale step parsing)} \\
\raisebox{-0.6ex}{\includegraphics[width=1.3em]{icons/gpt.png}}\ GPT-5-mini 
& \texttt{gpt-5-mini-2025-08-07} 
& Proprietary & minimal (medium for KCC) \\
\raisebox{-0.6ex}{\includegraphics[width=1.3em]{icons/gemma-color.png}}\ Gemma-3~\cite{gemmateam2025gemma3technicalreport} 
& \texttt{gemma-3-27b-it} 
& Gemma Terms of Use & - \\
\raisebox{-0.6ex}{\includegraphics[width=1.1em]{icons/qwen-color.png}}\ Qwen2.5-VL~\cite{bai2025qwen25vltechnicalreport} 
& \texttt{Qwen2.5-VL-7B-Instruct} 
& Apache 2.0 & - \\
\raisebox{-0.6ex}{\includegraphics[width=1.3em]{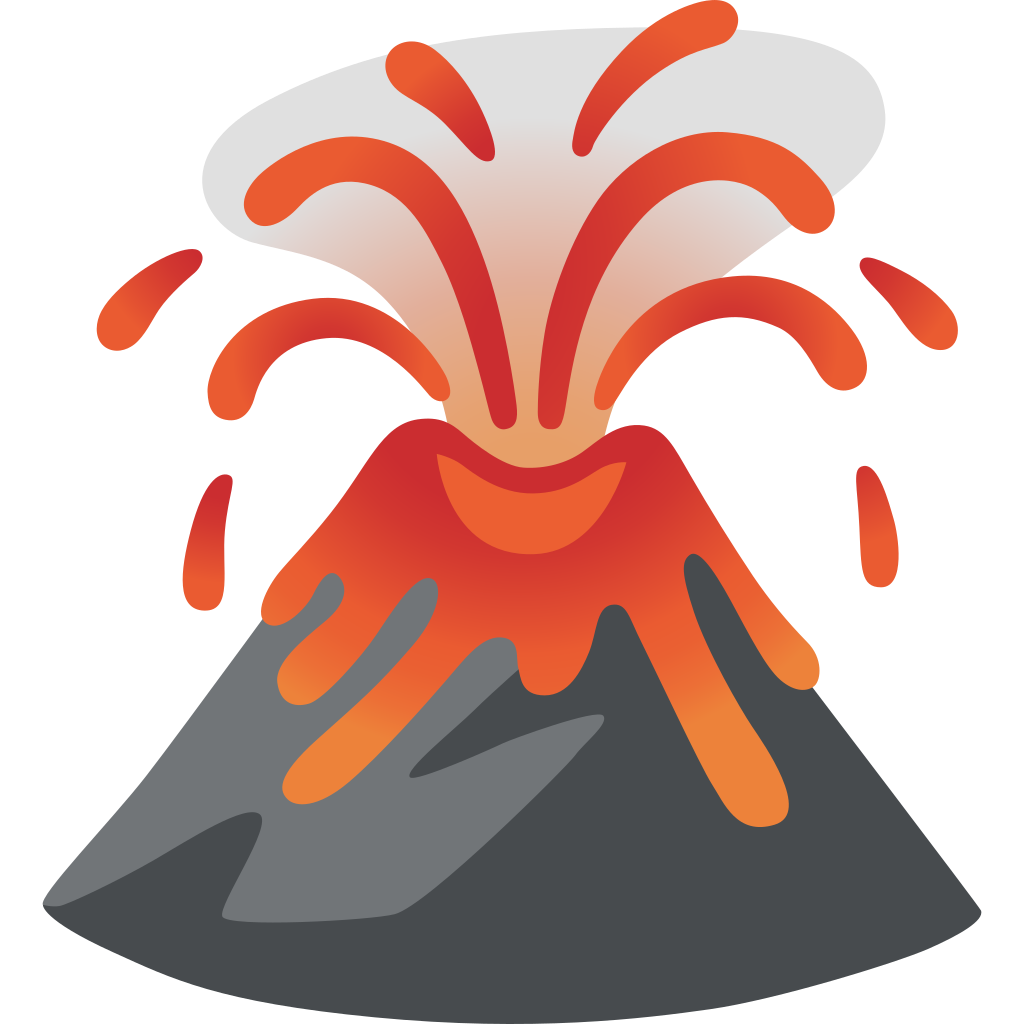}}\ LLaVA-1.5~\cite{liu2023llava} 
& \texttt{llava-1.5-13b-hf} 
& Llama 2 Community License & - \\
\raisebox{-0.6ex}{\includegraphics[width=1.3em]{icons/llava-color.png}}\ LLaVA-Next~\cite{liu2024llavanext} 
& \texttt{llava-v1.6-mistral-7b-hf} 
& Apache 2.0 & - \\
\bottomrule
\end{tabular}
}
\caption{Baseline models used for binary persuasiveness judgment.}
\label{tab:model_version}
\end{table*}

\subsection{Knowledge Injection and Reasoning Scaffold}
\label{app:knolwedge_injection_and_reasoning}

\begin{Prompt}[caption={Factor-aware ognitive injection prompt. Factor instruction details are in the Tab.~\ref{tab:cognitive_injection_prompts} and Tab.~\ref{tab:all_factors_prompt}.},label={pr:cognitive_injection}]
System Instruction: You are an AI assistant evaluating whether an image is persuasive for the given message. "{factor_instruction}"

User Prompt:
"""
Below is an instruction that describes a task, paired with an input that provides further context. Write a response that appropriately completes the request.

### Instruction:
You will perform a task where you predict how persuasive certain individuals will find an image created from a message. Please classify the image as either persuasive or not persuasive based on the message and image. Respond with a single word: 'yes' if persuasive, 'no' if not persuasive.

### Input:
Message: {message}

please directly output your answer by strictly following this format: [[answer]], for example: [[yes]]

### Response:
"""
\end{Prompt}

\begin{table*}[t]
\centering
\small
\renewcommand{\arraystretch}{1.12}
\begin{tabularx}{\textwidth}{
>{\raggedright\arraybackslash}p{0.18\textwidth}
>{\raggedright\arraybackslash}X}
\toprule
\textbf{Condition} & \textbf{System instruction} \\
\midrule

Color &
When making your judgment, explicitly consider the image’s overall color properties (dominant hues, saturation, and overall colorfulness). Use color only as supporting evidence when it plausibly affects persuasion for this message. \\
\addlinespace[4pt]
\midrule

Brightness &
When making your judgment, explicitly consider the image's overall brightness and contrast, including how light or dark it appears and how clearly key elements are visible. Use brightness only as supporting evidence when it plausibly affects persuasion for this message. \\
\addlinespace[4pt]
\midrule

Saliency-Center &
In addition to the image, you may be provided with a saliency cue indicating whether visual attention is concentrated near the image center. Saliency refers to where viewers are naturally drawn to look first or most. Treat this as a weak compositional cue, not a decisive rule. \\
\addlinespace[4pt]
\midrule

Saliency-RoT &
In addition to the image, you may be provided with a saliency cue indicating whether visual attention is concentrated near rule-of-thirds points, approximately at the 1/3 and 2/3 intersections of image width and height. Saliency refers to where viewers are naturally drawn to look first or most. Treat this as a weak compositional cue, not a decisive rule. \\
\addlinespace[4pt]
\midrule

MRR-Center &
In addition to the image, identify the message-relevant region, if any, that best corresponds to the message content. Then assess whether this region is located near the image center. The message-relevant region may be absent, ambiguous, or not visually present. If you cannot confidently identify such a region, ignore this consideration and judge normally. \\
\addlinespace[4pt]
\midrule

MRR-RoT &
In addition to the image, identify the message-relevant region, if any, that best corresponds to the message content. Then assess whether this region is located near rule-of-thirds points, approximately at the 1/3 and 2/3 intersections of image width and height. The message-relevant region may be absent, ambiguous, or not visually present. If you cannot confidently identify such a region, ignore this consideration and judge normally. \\
\addlinespace[4pt]
\midrule

Key Object &
Pay strong attention to whether the image depicts the message's key object(s); then judge whether and how much the presence, clarity, and prominence of those key object(s) should influence persuasiveness in this specific case. If there are no visible key object(s), proceed without using this cue. \\
\addlinespace[4pt]
\midrule

Human &
When making your judgment, consider whether people or humans are visually present, including faces, bodies, or crowds, but treat human presence as neutral by default. Do not assume that including people automatically increases persuasiveness. Only use human presence as supporting evidence if it clearly contributes to persuasion for this specific message, for example through social proof, demonstration, emotion, credibility, or context. \\
\addlinespace[4pt]
\midrule

Text &
When making your judgment, explicitly check whether the image contains visible text, such as captions, slogans, labels, overlays, signs, or dense paragraphs. Use text only as supporting evidence when it plausibly affects persuasion for this message. If there is no visible text, proceed without using this cue. \\

\bottomrule
\end{tabularx}
\caption{Factor-specific system instructions used for factor-aware cognitive injection.}
\label{tab:cognitive_injection_prompts}
\end{table*}

\begin{table*}[t]
\centering
\small
\renewcommand{\arraystretch}{1.12}
\begin{tabularx}{\textwidth}{
>{\raggedright\arraybackslash}p{0.18\textwidth}
>{\raggedright\arraybackslash}X}
\toprule
Condition & System instruction \\
\midrule

All Factors &
\begin{minipage}[t]{\linewidth}
You are an AI assistant evaluating whether an image is persuasive for the given message.

When making your judgment, jointly consider the following nine visual persuasion factors. Use them as supporting evidence, not as rigid rules. No single factor should dominate unless it clearly changes how persuasive the image is for the specific message.

\vspace{2pt}
1. Color: Consider the image's overall color properties, including dominant hues, saturation, and overall colorfulness. Use color as evidence when it reinforces or undermines the message's tone, credibility, urgency, or emotional framing.

\vspace{2pt}
2. Brightness: Consider the image's overall brightness and contrast, including whether key elements are clearly visible. Use brightness as evidence when lightness, darkness, or visibility plausibly affects persuasion.

\vspace{2pt}
3. Message-Relevant Region Centering: Identify the message-relevant visual region, if any, that best corresponds to the message content. Assess whether this region is located near the image center. If the relevant region is absent or ambiguous, ignore this cue.

\vspace{2pt}
4. Message-Relevant Region Rule of Thirds: Identify the message-relevant visual region, if any, and assess whether it is located near rule-of-thirds points, approximately at the 1/3 and 2/3 intersections of image width and height. If the relevant region is absent or ambiguous, ignore this cue.

\vspace{2pt}
5. Saliency Centering: Consider whether visual attention is concentrated near the image center. Treat center saliency as a weak compositional cue, not a decisive rule.

\vspace{2pt}
6. Saliency Rule of Thirds: Consider whether visual attention is concentrated near rule-of-thirds points. Treat rule-of-thirds saliency as a weak compositional cue, not a decisive rule.

\vspace{2pt}
7. Key Object Presence: Pay strong attention to whether the image clearly depicts the key object(s) implied by the message. Clear depiction of the core object can increase persuasiveness, while absence or ambiguity can weaken it.

\vspace{2pt}
8. Human Presence: Consider whether people or humans are visually present, including faces, bodies, or crowds. Treat human presence as neutral by default. Use it as evidence only when it contributes to persuasion through social proof, demonstration, emotion, credibility, or context.

\vspace{2pt}
9. Text Presence: Check whether the image contains visible text such as captions, slogans, labels, overlays, signs, or dense paragraphs. Use text as evidence when it helps clarify or support the message.

\vspace{2pt}
If a cue is not visible, absent, or ambiguous, simply proceed without using that cue. After considering all applicable factors together, decide whether the image is persuasive for the message.
\end{minipage}
\\

\bottomrule
\end{tabularx}
\caption{System instruction used for the All Factors condition.}
\label{tab:all_factors_prompt}
\end{table*}

\clearpage
\input{latex/sections/CI_original_result}
\clearpage

\begin{Prompt}[caption={Knowledge-conditioned chain.},label={pr:knowledge_chain}]
System Instruction: 
"You are an AI assistant with expertise in analyzing and evaluating the persuasiveness of images and messages based on general principles of communication and psychology."

User Prompt: 
"""
Below is an instruction that describes a task, paired with an input that provides further context. Write a response that appropriately completes the request.

### Instruction:
Decide whether an image (if present or implied) created from the message would be persuasive to likely viewers.

Before reasoning, note:
Key objects are supporting cues, not determining factors.
- They can provide additional hints for interpreting the message but are not required for persuasiveness.
- Their presence may support or reinforce the message when contextually aligned, but does not guarantee higher persuasiveness.
- Their absence does not necessarily reduce persuasiveness unless the message explicitly depends on them.

Use the following reasoning steps, and show your reasoning explicitly before giving the final answer:
1) Extract candidate key objects.
2) Analyze visual-message alignment.
3) Make a final judgment on whether the image effectively expresses or reinforces the message.

### Input:
Message: {message}

### Response:
Reasoning: <your reasoning here>
Answer: [[yes]] or [[no]]
"""
\end{Prompt}

\begin{Prompt}[caption={Aligned key object context, only used when a key object exists.},label={pr:aligned_key_context}]
System Instruction: 
"You are an AI assistant with expertise in analyzing and evaluating the persuasiveness of images and messages based on general principles of communication and psychology."

User Prompt: 
"""
Below is an instruction that describes a task, paired with an input that provides further context. Write a response that appropriately completes the request.

### Instruction:
You will perform a task where you predict how persuasive certain individuals will find an image created from a message. Respond with a single word: 'yes' if persuasive, 'no' if not persuasive.
IMPORTANT: The following key object(s) were algorithmically detected in the image and should be treated as verified evidence of presence. Use these objects only as evidence for what appears in the image; your final decision should still be based on overall persuasiveness.

### Input:
Message: {message}
Key Object: {key_object_json}

please directly output your answer by strictly following this format: [[answer]], for example: [[yes]]

### Response:
"""
\end{Prompt}

\subsection{Rationale Generator}
\label{app:rationale_gen}
We use a rationale-generation prompt, where the model generates rationales without access to the human persuasiveness label.
This setting allows us to examine whether models can explain the persuasive role of detected key objects without being conditioned on the ground-truth judgment.

\begin{Prompt}[caption={Key object rationale generation.},label={pr:rationale_gen_implicit}]
System Instruction: 
"You are an expert visual persuasion analyst. Given an image, a message, and detected key object tokens, explain concisely how the object(s) likely affect persuasiveness, focusing on message-image alignment, object salience/clarity, emotional/moral cues, and credibility."

User Prompt: 
"""
Analyze the following image-message pair.

### Instructions:
1) Provide a concise overall rationale for how persuasive (or not) the pair is likely to be.
2) For EACH key object in the list, write 1-2 short, specific sentences explaining whether and how that object increased or decreased persuasiveness.
3) Be concrete (what is visible + how it supports/undermines the message). Avoid speculation beyond the image.
4) OUTPUT **ONLY** a JSON object with this exact schema (no extra text):
{{
  "per_object": {{"<object1>": "<1-2 short sentences>", "<object2>": "<...>"}},
}}

### Inputs:
message: {message}
key_objects: {key_objects}
Return JSON only.

### Response:
"""
\end{Prompt}

\begin{Prompt}[caption={Key object rationale, only used when a key object exists. The \texttt{\{key\_objects\}} field contains the per-object rationales.},label={pr:key_object_rationale}]
System Instruction: 
"You are an AI assistant with expertise in analyzing and evaluating the persuasiveness of images and messages based on general principles of communication and psychology."

User Prompt: 
"""
Below is an instruction that describes a task, paired with an input that provides further context. Write a response that appropriately completes the request.

### Instruction:
You will perform a task where you predict how persuasive certain individuals will find an image created from a message. You are also given a verbatim JSON map of per-object rationales extracted from prior analysis. Use these rationales as-is as additional context. Please classify the image as either persuasive or not persuasive based on the message and image. Respond with a single word: 'yes' if persuasive, 'no' if not persuasive.

### Input:
Message: {message}
Key Objects: {key_objects}  # for reference

please directly output your answer by strictly following this format: [[answer]], for example: [[yes]]

### Response:
"""
\end{Prompt}

\subsection{Step-Wise Rationale Parsing Details}
\paragraph{Parsing Procedure Details.}
\label{app:stepwise_ablation}
For the step-wise rationale analysis, we used GPT-5 to parse each generated key object rationale into three functional spans.
We used reasoning effort set to \texttt{low}, rather than \texttt{minimal}, because we empirically found that \texttt{low} produced more reliable and stable span annotations while still preserving the original rationale text.
The parser was instructed to insert only step tags into the original rationale, without paraphrasing, deleting, or reordering any part of the text.
The exact prompt used for parsing is shown in Prompt~\ref{pr:s1s2s3_parsing}.

\begin{Prompt}[caption={Step-wise rationale parsing.},label={pr:s1s2s3_parsing}]
System Instruction:
"You are a strict text editor.
Your job is to insert ONLY the tags <S1>...</S1>, <S2>...</S2>, <S3>...</S3> into the given text.
You MUST NOT change, paraphrase, reorder, delete, or add any characters of the original text.
The only allowed edits are inserting these tags at character boundaries.

You must segment the text into exactly three contiguous spans:
- <S1> Factor Detection / Evidence (what is observed or stated as evidence)
- <S2> Evidence-to-Claim Mapping (how the evidence supports or conflicts with the message/claim)
- <S3> Persuasive Judgement / Deduction (final evaluative conclusion)

Segmentation policy (IMPORTANT):
- You may use these tags MULTIPLE TIMES.
- The order does NOT have to be S1 -> S2 -> S3. It can be in ANY ORDER depending on the text structure.

Hard constraints:
1) All original text must be covered by exactly one of S1/S2/S3. No text outside the tags.
2) Do NOT insert any other tokens, headings, bullet points, or explanations.
3) Output must be valid JSON only."

User Prompt:
"""
obj_key: {obj_key}

rationale_text (DO NOT EDIT ANY CHARACTERS, ONLY INSERT TAGS):
{rationale}

Return JSON only in this schema:
{
  "obj_key": "{obj_key}",
  "tagged_rationale": "<TAGGED_RATIONALE>"
}
"""
\end{Prompt}

\begin{figure*}[p]
    \centering
    \includegraphics[page=1,width=\textwidth,height=0.93\textheight,keepaspectratio]{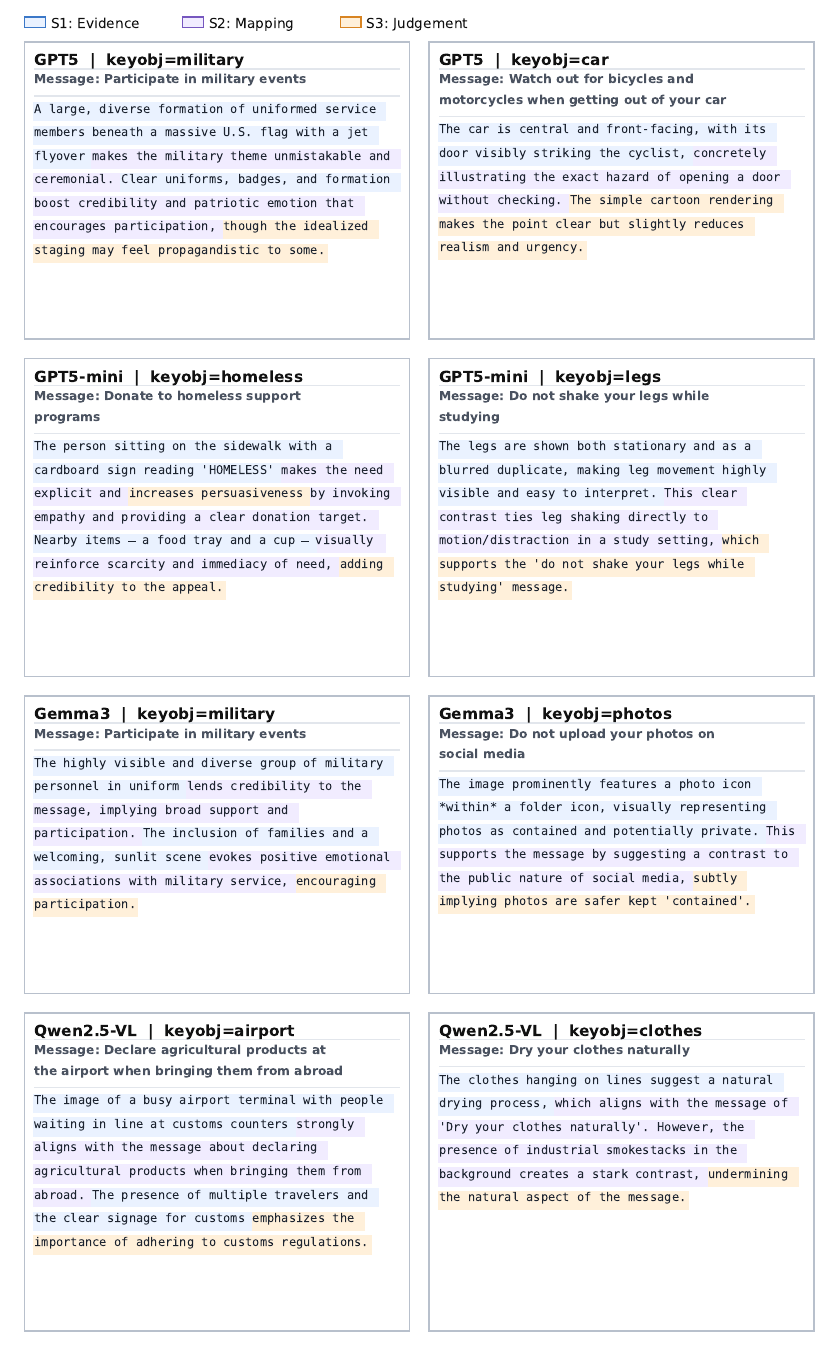}
    \caption{
    Examples of step-wise rationale parsing. 
    Each rationale is shown with text-preserving span annotations for \textbf{S1} evidence detection, \textbf{S2} evidence-to-claim mapping, and \textbf{S3} persuasive judgment.
    }
    \label{fig:s1s2s3_parsing_examples}
\end{figure*}

\paragraph{Parsing Examples.}
Fig.~\ref{fig:s1s2s3_parsing_examples} shows representative parsed rationales across models.
The examples illustrate that rationale functions are not always separated by sentence boundaries: a single sentence can contain both evidence and mapping, while longer rationales may alternate between evidence and interpretation before reaching the final judgment.

\begin{figure*}[t]
    \centering
    \includegraphics[width=\linewidth]{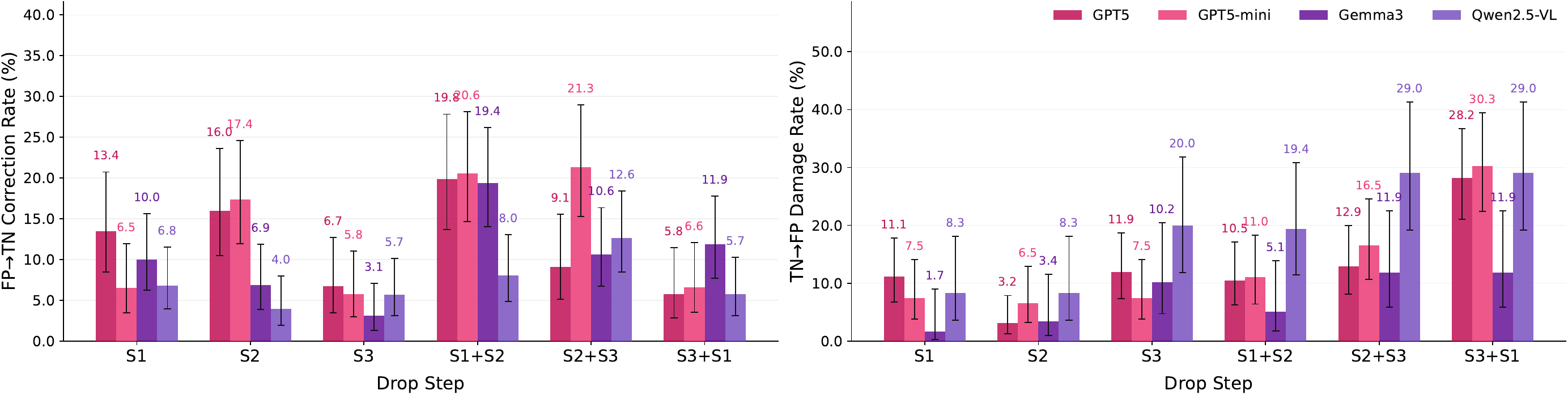}
    \caption{
    Correction--damage trade-off, comparing FP-to-TN corrections against TN-to-FP errors.
    }
    \label{fig:combined_fp_correction_damage_agnostic}
\end{figure*}

\begin{figure}[t]
    \centering
    \includegraphics[width=\linewidth]{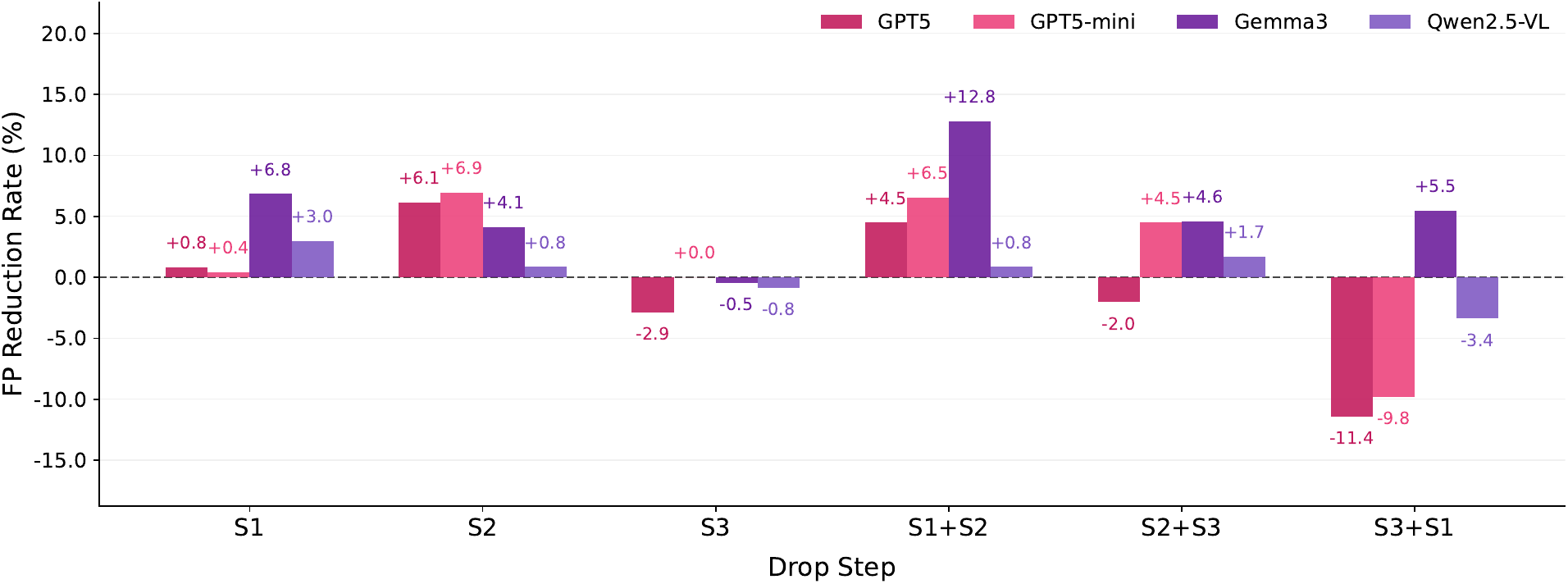}
    \caption{
    Reduction in false positive rate after removing each rationale step.
    }
    \label{fig:combined_fp_reduction_agnostic}
\end{figure}

\paragraph{Human Verification Study of the Rationale Segmentation.}
We independently validate the GPT-5-generated rationale segmentation using 25 qualified Amazon Mechanical Turk (MTurk) workers with more than 10{,}000 approved HITs and approval rates above 98\%. 
Each worker received \$1.50 for an approximately 10-minute task.
Following our validation survey protocol, we apply the same quality controls, including trap questions. 
Workers make a binary judgment for each rationale span to verify its assignment to $S1$, $S2$, or $S3$. 
As shown in Tab.~\ref{tab:rationale_segmentation_verify}, 92.04\% of the segmentations are verified as correct. 
We exclude incorrectly segmented instances from the reported step-wise ablation results.

\begin{table}[t]
\centering
\small
\begin{tabular}{lc}
\toprule
\textbf{Verification outcome} & \textbf{Rate (\%)} \\
\midrule
Parsed correctly   & 92.04 \\
Parsed incorrectly & 7.96  \\
\bottomrule
\end{tabular}
\caption{Human verification results for the GPT-5-generated rationale segmentation.}
\label{tab:rationale_segmentation_verify}
\end{table}

\paragraph{Robustness to Rationale Step Ordering.}
As a robustness check against incidental sentence-form effects, we test whether model predictions depend on the original $S1 \rightarrow S2 \rightarrow S3$ ordering. 
While preserving the same rationale content and length, we compare the original ordering with two shuffled variants: $S2 \rightarrow S1 \rightarrow S3$ and $S3 \rightarrow S2 \rightarrow S1$. 
We measure the proportion of items receiving the same prediction across all three orderings (three-way agreement) and the mean agreement between each pair of orderings. 
As shown in Tab.~\ref{tab:rationale_order_robustness}, predictions remain highly consistent across orderings, with 90.69--94.68\% three-way agreement and 93.79--96.45\% mean pairwise agreement. 
This indicates that model behavior is driven primarily by the rationale content rather than its ordering. 
Together with the human-verified segmentation results, this reduces the likelihood that the ablation findings arise from parsing errors or fragile prompt structure.

\begin{table}[t]
\centering
\small
\resizebox{\columnwidth}{!}{%
\begin{tabular}{lcc}
\toprule
\textbf{Model} & \textbf{Three-way agreement (\%)} 
& \textbf{Mean pairwise agreement (\%)} \\
\midrule
GPT-5       & 94.68 & 96.45 \\
GPT-5-mini  & 90.69 & 93.79 \\
Gemma3-27B  & 93.81 & 95.87 \\
Qwen2.5-VL  & 91.82 & 94.55 \\
\bottomrule
\end{tabular}%
}
\caption{Prediction agreement across rationale-step orderings.}
\label{tab:rationale_order_robustness}
\end{table}

\paragraph{False Positive Analysis.}
\label{app:step_wise_additional_result}
We further analyze step-wise ablation from the perspective of false positives. 
As shown in Fig.~\ref{fig:combined_fp_reduction_agnostic}, removing rationale components can reduce the tendency of models to incorrectly classify GT-Low items as persuasive, but the effect is not uniform across steps or models. 
In particular, ablating the evidence-to-claim mapping step, either alone or together with evidence detection, often yields the largest false-positive reduction, suggesting that models may overestimate persuasiveness when their rationales construct overly strong links between visual evidence and the target message. 
Fig.~\ref{fig:combined_fp_correction_damage_agnostic} further shows that this improvement should be interpreted as a correction--damage trade-off: some ablations successfully convert false positives into true negatives, while others also introduce new TN-to-FP errors. 
Overall, these results indicate that false positives are caused by intermediate reasoning steps that over-align visual cues with the persuasive claim.

%% file: latex/sections/app_cperceptual_table.tex
\begin{table*}[t]
\centering
\scriptsize
\renewcommand{\arraystretch}{1.18}
\setlength{\tabcolsep}{2pt}
\arrayrulecolor{gray!45}

\begin{adjustbox}{max width=\linewidth}
\begin{tabular}{
  >{\centering\arraybackslash}m{0.088\linewidth}
  >{\centering\arraybackslash}m{0.098\linewidth}
  >{\centering\arraybackslash}m{0.125\linewidth}
  >{\centering\arraybackslash}m{0.080\linewidth}
  >{\centering\arraybackslash}m{0.155\linewidth}
  >{\centering\arraybackslash}m{0.125\linewidth}
  >{\centering\arraybackslash}m{0.075\linewidth}
  >{\centering\arraybackslash}m{0.085\linewidth}
}
\toprule
\textbf{Image}
& \textbf{Clustered Palette}
& \textbf{Message}
& \textbf{Representative Color}
& \textbf{Premise}
& \textbf{GPT-5 Infer Color}
& \textbf{Brightness $L^*$}
& \textbf{Brightness Score} \\
\midrule

\includegraphics[width=0.95\linewidth]{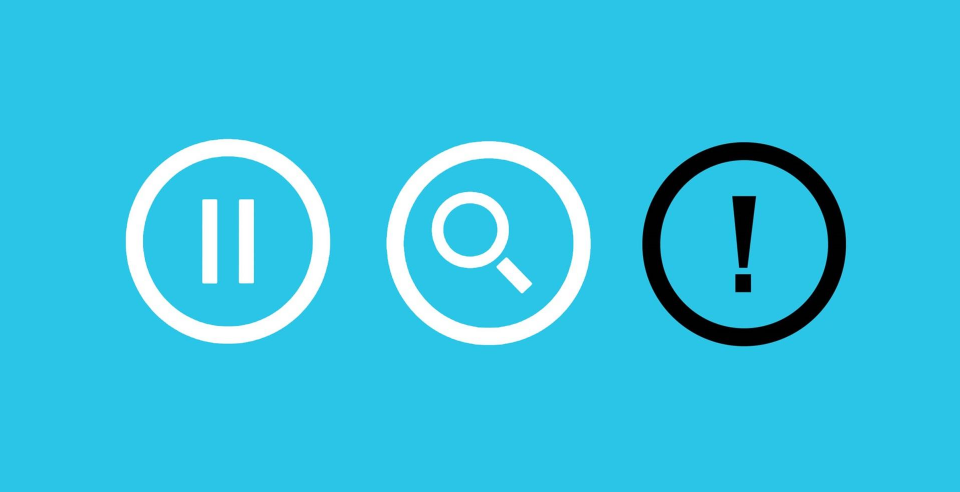}
&
\includegraphics[width=0.95\linewidth]{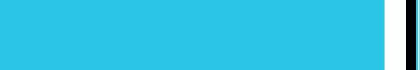}
&
\makecell[c]{Report fake news\\when you find it}
&
Blue
&
\makecell[c]{Positive effect:\\Safer online\\environment}
&
\makecell[c]{Top3:\\Blue, Green,\\Yellow}
&
73.29
&
0.733
\\
\midrule

\includegraphics[width=0.95\linewidth]{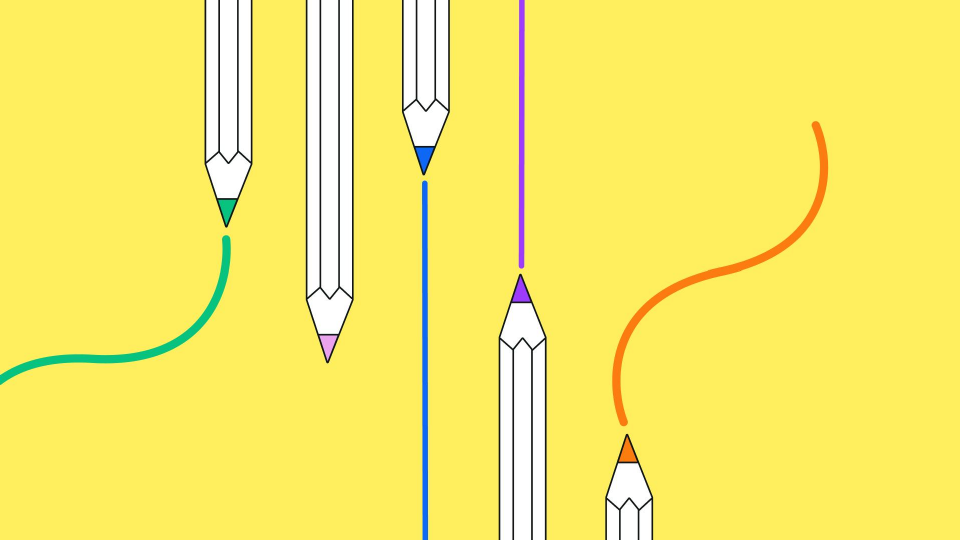}
&
\includegraphics[width=0.95\linewidth]{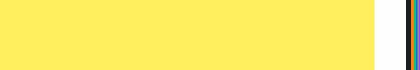}
&
\makecell[c]{Take an interest\\in the energy\\industry}
&
Yellow
&
\makecell[c]{Positive effect:\\Others can feel\\inspired by\\your passion}
&
\makecell[c]{Top3:\\Yellow, Blue,\\Green}
&
92.05
&
0.920
\\
\midrule

\includegraphics[width=0.95\linewidth]{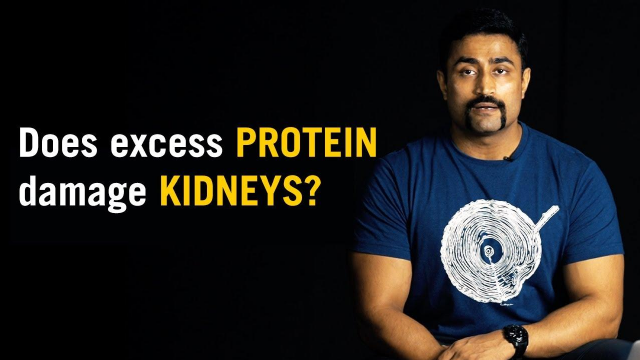}
&
\includegraphics[width=0.95\linewidth]{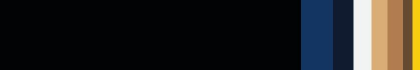}
&
\makecell[c]{Reduce protein\\shake intake}
&
Black
&
\makecell[c]{Negative effect:\\Kidney strain\\increases\\significantly}
&
\makecell[c]{Top3:\\Black, Red,\\Grey}
&
13.99
&
0.860
\\
\midrule

\includegraphics[width=0.95\linewidth]{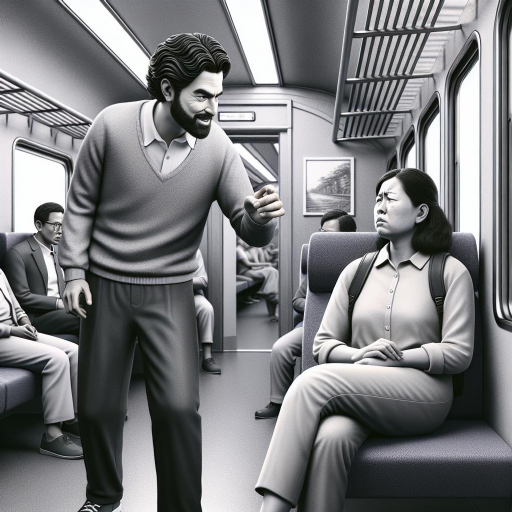}
&
\includegraphics[width=0.95\linewidth]{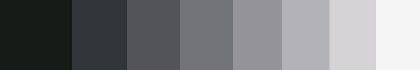}
&
\makecell[c]{Avoid conversations\\with strangers\\while traveling}
&
Black
&
\makecell[c]{Negative effect:\\You can be seen\\as disrespectful\\of privacy}
&
\makecell[c]{Top3:\\Black, Grey,\\Red}
&
50.21
&
0.498
\\
\midrule

\includegraphics[width=0.95\linewidth]{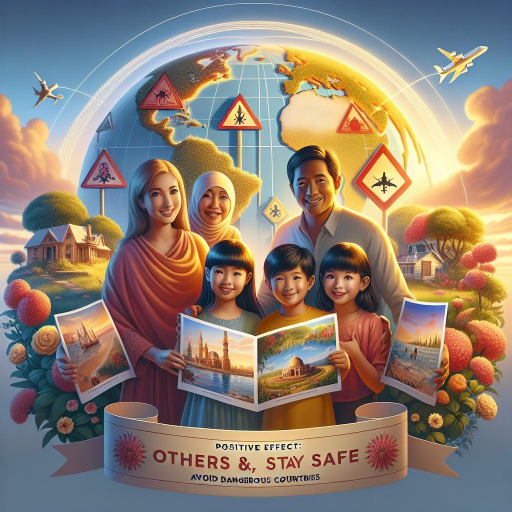}
&
\includegraphics[width=0.95\linewidth]{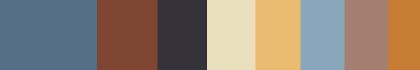}
&
\makecell[c]{Do not visit\\dangerous\\countries}
&
Blue
&
\makecell[c]{Positive effect:\\Others can\\avoid worry}
&
\makecell[c]{Top3:\\Yellow, Blue,\\White}
&
54.59
&
0.546
\\
\midrule

\includegraphics[width=0.95\linewidth]{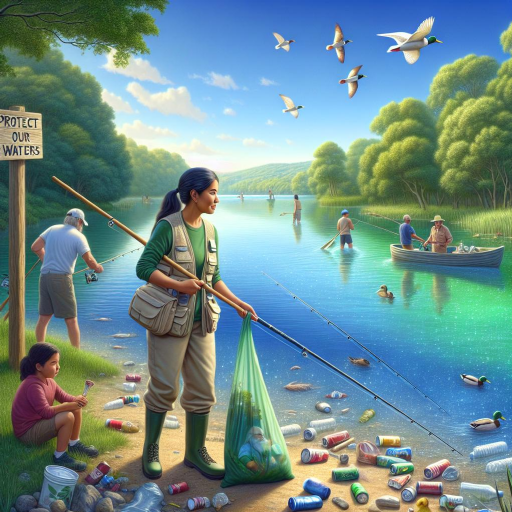}
&
\includegraphics[width=0.95\linewidth]{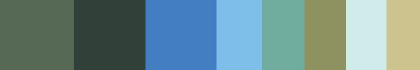}
&
\makecell[c]{Clean up trash\\after fishing}
&
Green
&
\makecell[c]{Positive effect:\\Others can see\\your care for\\the environment}
&
\makecell[c]{Top3:\\Green, Blue,\\Yellow}
&
56.68
&
0.567
\\

\bottomrule
\end{tabular}
\end{adjustbox}

\arrayrulecolor{black}
\caption{
Illustrative examples of representative color congruence and brightness alignment.
Representative Color denotes the color category of the top cluster extracted from the clustered palette.
GPT-5 Color denotes the color categories inferred to support the message.
Brightness $L^*$ is the original mean lightness value in CIELAB space, and Brightness Score is computed from normalized brightness and inferred strategy polarity.
}
\label{tab:rep_congruence_examples}
\end{table*}

%% file: latex/sections/appendix_agreement_vpfs_table.tex
\begin{table*}[t]
\centering
\renewcommand{\arraystretch}{1.12}
\setlength{\tabcolsep}{2.7pt}

\setlength{\abovecaptionskip}{5pt}
\setlength{\belowcaptionskip}{0pt}
\begin{adjustbox}{max width=\linewidth}
\begin{tabular}{l l l ccc ccc cc cc ccc}
\toprule
\multirow{3}{*}{\textbf{Agreement}}
& \multirow{3}{*}{\textbf{Model}}
& \multirow{3}{*}{\textbf{Row}}
& \multicolumn{6}{c}{\textbf{Perceptual (\%)}}
& \multicolumn{4}{c}{\textbf{Compositional (\%)}}
& \multicolumn{3}{c}{\textbf{Semantic (OR)}} \\
\cmidrule(lr){4-9}\cmidrule(lr){10-13}\cmidrule(lr){14-16}

&
&
& \multicolumn{3}{c}{Color}
& \multicolumn{3}{c}{Brightness}
& \multicolumn{2}{c}{CBI}
& \multicolumn{2}{c}{T3}
& \multirow{2}{*}{KeyObj}
& \multirow{2}{*}{Human}
& \multirow{2}{*}{Text} \\
\cmidrule(lr){4-6}\cmidrule(lr){7-9}\cmidrule(lr){10-11}\cmidrule(lr){12-13}

&
&
& Colorfulness & Entropy & \makecell{Rep.\\Congruence}
& Raw & Pos & Neg
& Saliency & MRR
& Saliency & MRR
& & & \\
\midrule

\rowcolor{gray!15!violet!10}
\multicolumn{16}{l}{\textbf{Panel A: VLM Predictions}} \\
\addlinespace[2pt]

\multirow{4}{*}{\makecell{Almost Perfect\\$\sim$ Fair}}
& GPT-5
& $\Delta_f^{(\hat{Y})}$
& -0.46\stat{}{n}
& 4.78\stat{\sym{***}}{s}
& 3.35\stat{}{n}
& -0.68\stat{}{n}
& -6.54\stat{\sym{***}}{s}
& 6.08\stat{\sym{***}}{s}
& 2.75\stat{\sym{***}}{s}
& -1.80\stat{}{n}
& 3.14\stat{\sym{***}}{s}
& 0.97\stat{}{n}
& 2.52\stat{\sym{***}}{s}
& 1.57\stat{\sym{***}}{s}
& 1.32\stat{}{n} \\
\addlinespace[1pt]

& GPT-5-mini
& $\Delta_f^{(\hat{Y})}$
& -1.66\stat{}{n}
& 4.89\stat{\sym{***}}{s}
& 2.60\stat{}{n}
& 1.20\stat{}{n}
& -6.25\stat{\sym{**}}{s}
& 9.15\stat{\sym{***}}{m}
& 3.48\stat{\sym{***}}{s}
& -0.29\stat{}{n}
& 3.52\stat{\sym{***}}{m}
& 0.75\stat{}{n}
& 2.07\stat{\sym{***}}{s}
& 1.78\stat{\sym{***}}{s}
& 1.14\stat{}{n} \\
\addlinespace[1pt]

& Gemma3
& $\Delta_f^{(\hat{Y})}$
& -2.53\stat{}{n}
& 4.15\stat{\sym{***}}{s}
& 5.98\stat{}{n}
& 0.53\stat{}{n}
& -6.74\stat{\sym{***}}{s}
& 9.39\stat{\sym{***}}{m}
& 3.80\stat{\sym{***}}{s}
& -1.12\stat{}{n}
& 3.56\stat{\sym{***}}{m}
& 1.68\stat{}{n}
& 1.74\stat{\sym{***}}{s}
& 1.78\stat{\sym{***}}{s}
& 1.41\stat{\sym{***}}{n} \\
\addlinespace[1pt]

& Qwen2.5-VL
& $\Delta_f^{(\hat{Y})}$
& -2.48\stat{}{n}
& 5.11\stat{\sym{***}}{s}
& 4.06\stat{}{n}
& 2.52\stat{\sym{*}}{n}
& -3.51\stat{}{s}
& 10.91\stat{\sym{***}}{m}
& 4.83\stat{\sym{***}}{s}
& 0.75\stat{}{n}
& 4.40\stat{\sym{***}}{m}
& 2.23\stat{}{n}
& 2.28\stat{\sym{***}}{s}
& 2.40\stat{\sym{***}}{s}
& 1.17\stat{}{n} \\
\midrule

\multirow{4}{*}{\makecell{Almost Perfect\\$\sim$ Slight}}
& GPT-5
& $\Delta_f^{(\hat{Y})}$
& -0.83\stat{}{n}
& 4.62\stat{\sym{***}}{s}
& 1.68\stat{}{n}
& 0.90\stat{}{n}
& -3.74\stat{\sym{***}}{s}
& 5.68\stat{\sym{***}}{s}
& 2.26\stat{\sym{***}}{s}
& -0.40\stat{}{n}
& 2.14\stat{\sym{***}}{s}
& 0.32\stat{}{n}
& 2.20\stat{\sym{***}}{s}
& 1.25\stat{\sym{**}}{n}
& 1.07\stat{}{n} \\
\addlinespace[1pt]

& GPT-5-mini
& $\Delta_f^{(\hat{Y})}$
& -0.81\stat{}{n}
& 4.48\stat{\sym{***}}{s}
& -1.36\stat{}{n}
& 0.66\stat{}{n}
& -3.73\stat{\sym{***}}{s}
& 6.13\stat{\sym{***}}{s}
& 3.01\stat{\sym{***}}{s}
& 0.03\stat{}{n}
& 2.59\stat{\sym{***}}{s}
& 0.27\stat{}{n}
& 1.92\stat{\sym{***}}{s}
& 1.34\stat{\sym{***}}{n}
& 1.13\stat{}{n} \\
\addlinespace[1pt]

& Gemma3
& $\Delta_f^{(\hat{Y})}$
& -1.59\stat{}{n}
& 5.33\stat{\sym{***}}{s}
& 1.62\stat{}{n}
& 1.61\stat{\sym{**}}{n}
& -4.72\stat{\sym{***}}{s}
& 8.91\stat{\sym{***}}{m}
& 3.41\stat{\sym{***}}{s}
& -0.28\stat{}{n}
& 3.06\stat{\sym{***}}{s}
& 1.18\stat{\sym{**}}{n}
& 1.48\stat{\sym{***}}{n}
& 1.86\stat{\sym{***}}{s}
& 1.23\stat{\sym{**}}{n} \\
\addlinespace[1pt]

& Qwen2.5-VL
& $\Delta_f^{(\hat{Y})}$
& -0.60\stat{}{n}
& 7.01\stat{\sym{***}}{m}
& 3.56\stat{\sym{*}}{n}
& 2.28\stat{\sym{**}}{n}
& -5.92\stat{\sym{***}}{s}
& 8.51\stat{\sym{***}}{m}
& 4.60\stat{\sym{***}}{s}
& 1.52\stat{}{n}
& 3.95\stat{\sym{***}}{m}
& 1.55\stat{\sym{***}}{n}
& 1.92\stat{\sym{***}}{s}
& 2.49\stat{\sym{***}}{s}
& 0.99\stat{}{n} \\

\addlinespace[3pt]

\rowcolor{gray!15!violet!10}
\multicolumn{16}{l}{\textbf{Panel B: VLM False Positives among Ground-Truth Low Items}} \\
\addlinespace[2pt]

\multirow{4}{*}{\makecell{Almost Perfect\\$\sim$ Fair}}
& GPT-5
& $\Delta_f^{(FP-TN)}$
& 1.74\stat{}{n}
& 4.12\stat{\sym{***}}{s}
& -3.81\stat{}{n}
& -1.75\stat{}{n}
& -7.15\stat{\sym{**}}{s}
& 3.23\stat{}{n}
& 2.51\stat{\sym{**}}{s}
& 0.05\stat{}{n}
& 2.61\stat{\sym{***}}{s}
& -0.46\stat{}{n}
& 1.69\stat{\sym{**}}{s}
& 1.53\stat{\sym{*}}{n}
& 1.88\stat{\sym{***}}{s} \\
\addlinespace[1pt]

& GPT-5-mini
& $\Delta_f^{(FP-TN)}$
& -0.14\stat{}{n}
& 3.80\stat{\sym{***}}{s}
& -2.71\stat{}{n}
& 0.75\stat{}{n}
& -6.49\stat{\sym{***}}{s}
& 7.33\stat{\sym{***}}{s}
& 3.36\stat{\sym{***}}{s}
& 1.76\stat{}{n}
& 3.13\stat{\sym{***}}{s}
& -0.49\stat{}{n}
& 1.45\stat{\sym{*}}{n}
& 1.92\stat{\sym{***}}{s}
& 1.51\stat{\sym{***}}{n} \\
\addlinespace[1pt]

& Gemma3
& $\Delta_f^{(FP-TN)}$
& -2.15\stat{}{n}
& 2.99\stat{\sym{***}}{s}
& -3.96\stat{}{n}
& -0.42\stat{}{n}
& -7.47\stat{\sym{***}}{s}
& 8.51\stat{\sym{***}}{s}
& 3.34\stat{\sym{***}}{s}
& 1.01\stat{}{n}
& 3.09\stat{\sym{***}}{s}
& 0.99\stat{}{n}
& 1.12\stat{}{n}
& 1.76\stat{\sym{**}}{s}
& 1.84\stat{\sym{***}}{s} \\
\addlinespace[1pt]

& Qwen2.5-VL
& $\Delta_f^{(FP-TN)}$
& -2.00\stat{}{n}
& 4.40\stat{\sym{***}}{s}
& -2.88\stat{}{n}
& 1.75\stat{}{n}
& -3.41\stat{}{s}
& 10.27\stat{\sym{***}}{m}
& 3.97\stat{\sym{***}}{s}
& 1.80\stat{}{n}
& 3.66\stat{\sym{***}}{m}
& 0.99\stat{}{n}
& 1.78\stat{\sym{***}}{s}
& 2.24\stat{\sym{***}}{s}
& 1.27\stat{}{n} \\
\midrule

\multirow{4}{*}{\makecell{Almost Perfect\\$\sim$ Slight}}
& GPT-5
& $\Delta_f^{(FP-TN)}$
& 0.45\stat{\sym{***}}{n}
& 3.84\stat{\sym{***}}{s}
& -2.26\stat{}{n}
& 0.37\stat{}{n}
& -3.70\stat{\sym{***}}{s}
& 3.27\stat{\sym{***}}{n}
& 1.87\stat{\sym{***}}{n}
& 0.02\stat{}{n}
& 1.59\stat{\sym{***}}{s}
& -0.20\stat{}{n}
& 1.88\stat{\sym{***}}{s}
& 1.21\stat{\sym{***}}{n}
& 1.35\stat{\sym{***}}{n} \\
\addlinespace[1pt]

& GPT-5-mini
& $\Delta_f^{(FP-TN)}$
& 0.39\stat{}{n}
& 3.60\stat{\sym{***}}{s}
& -4.55\stat{\sym{***}}{n}
& 0.23\stat{}{n}
& -3.48\stat{\sym{**}}{s}
& 4.30\stat{\sym{***}}{s}
& 2.77\stat{\sym{***}}{s}
& 0.44\stat{}{n}
& 2.17\stat{\sym{***}}{s}
& -0.12\stat{}{n}
& 1.68\stat{\sym{***}}{s}
& 1.32\stat{\sym{**}}{n}
& 1.30\stat{\sym{**}}{n} \\
\addlinespace[1pt]

& Gemma3
& $\Delta_f^{(FP-TN)}$
& -1.20\stat{}{n}
& 4.94\stat{\sym{***}}{s}
& -1.67\stat{}{n}
& 1.68\stat{\sym{***}}{n}
& -4.89\stat{\sym{***}}{s}
& 8.56\stat{\sym{***}}{m}
& 3.16\stat{\sym{***}}{s}
& -0.26\stat{}{n}
& 2.75\stat{\sym{***}}{s}
& 0.91\stat{\sym{*}}{n}
& 1.27\stat{\sym{**}}{n}
& 1.98\stat{\sym{***}}{s}
& 1.34\stat{\sym{***}}{n} \\
\addlinespace[1pt]

& Qwen2.5-VL
& $\Delta_f^{(FP-TN)}$
& -0.84\stat{}{n}
& 6.64\stat{\sym{***}}{s}
& 1.84\stat{}{n}
& 2.06\stat{\sym{***}}{n}
& -6.18\stat{\sym{***}}{s}
& 8.02\stat{\sym{***}}{s}
& 3.99\stat{\sym{***}}{s}
& 1.57\stat{}{n}
& 3.41\stat{\sym{***}}{m}
& 1.06\stat{\sym{***}}{n}
& 1.82\stat{\sym{***}}{s}
& 2.21\stat{\sym{***}}{s}
& 0.97\stat{}{n} \\

\bottomrule
\end{tabular}
\end{adjustbox}

\caption{
VPF-based analysis across progressively broader annotator-agreement ranges.
Panel A reports associations between VPFs and VLM-predicted persuasiveness, while Panel B compares false positives with true negatives among ground-truth Low items.
Across both Fair and Slight agreement ranges, VLMs continue to partially reproduce the VPF patterns observed in the main high-agreement analysis, while false positives repeatedly exhibit persuasion-relevant cues.
\textbf{These results indicate that partial human--VLM alignment and false-positive cue overgeneralization persist as the evaluation is expanded to substantially broader agreement levels.}
}
\label{tab:agreement_vpf_generalization}
\end{table*}

%% file: latex/sections/appendix_diverse_model_table.tex
\begin{table*}[t]
\centering
\renewcommand{\arraystretch}{1.12}
\setlength{\tabcolsep}{3pt}

\setlength{\abovecaptionskip}{5pt}
\setlength{\belowcaptionskip}{0pt}
\begin{adjustbox}{max width=\linewidth}
\begin{tabular}{l l ccc ccc cc cc ccc}
\toprule
\multirow{3}{*}{\textbf{Model}}
& \multirow{3}{*}{\textbf{Row}}
& \multicolumn{6}{c}{\textbf{Perceptual (\%)}}
& \multicolumn{4}{c}{\textbf{Compositional (\%)}}
& \multicolumn{3}{c}{\textbf{Semantic (OR)}} \\
\cmidrule(lr){3-8}\cmidrule(lr){9-12}\cmidrule(lr){13-15}

&
& \multicolumn{3}{c}{Color}
& \multicolumn{3}{c}{Brightness}
& \multicolumn{2}{c}{CBI}
& \multicolumn{2}{c}{T3}
& \multirow{2}{*}{KeyObj}
& \multirow{2}{*}{Human}
& \multirow{2}{*}{Text} \\
\cmidrule(lr){3-5}\cmidrule(lr){6-8}\cmidrule(lr){9-10}\cmidrule(lr){11-12}

&
& Colorfulness & Entropy & \makecell{Rep.\\Congruence}
& Raw & Pos & Neg
& Saliency & MRR
& Saliency & MRR
& & & \\
\midrule

\rowcolor{gray!15!violet!10}
\multicolumn{15}{l}{\textbf{Panel A: Human Ground Truth and Expanded VLM Predictions}} \\
\addlinespace[2pt]

\raisebox{-0.7ex}{\includegraphics[width=1.3em]{icons/audience.png}}\ Human
& $\Delta_f^{(Y)}$
& -3.94\stat{\sym{**}}{s}
& 4.57\stat{\sym{***}}{s}
& 15.09\stat{\sym{***}}{s}
& -0.34\stat{}{n}
& -5.11\stat{}{s}
& 5.91\stat{\sym{**}}{s}
& 2.65\stat{\sym{**}}{s}
& -2.83\stat{}{n}
& 3.19\stat{\sym{***}}{s}
& 3.31\stat{\sym{**}}{s}
& 3.36\stat{\sym{***}}{m}
& 1.25\stat{}{n}
& 0.60\stat{\sym{*}}{s} \\
\midrule

Gemma3-4B
& $\Delta_f^{(\hat{Y})}$
& -1.53\stat{}{n}
& 4.30\stat{\sym{***}}{s}
& 9.42\stat{\sym{*}}{n}
& 2.35\stat{}{n}
& -7.57\stat{\sym{**}}{s}
& 6.01\stat{\sym{**}}{s}
& 4.44\stat{\sym{***}}{s}
& -1.17\stat{}{n}
& 3.89\stat{\sym{***}}{m}
& 0.61\stat{}{n}
& 1.70\stat{\sym{**}}{s}
& 1.65\stat{\sym{**}}{s}
& 1.18\stat{}{n} \\
\addlinespace[1pt]

Gemma3-12B
& $\Delta_f^{(\hat{Y})}$
& -1.97\stat{}{n}
& 1.37\stat{}{n}
& 10.97\stat{\sym{***}}{s}
& -1.36\stat{}{n}
& -1.84\stat{}{n}
& 4.83\stat{\sym{***}}{s}
& 1.46\stat{}{n}
& -1.97\stat{}{n}
& 1.78\stat{\sym{***}}{s}
& 1.31\stat{}{n}
& 2.80\stat{\sym{***}}{m}
& 1.34\stat{}{n}
& 1.31\stat{}{n} \\
\addlinespace[1pt]

Gemma3-27B
& $\Delta_f^{(\hat{Y})}$
& -2.87\stat{}{n}
& 3.08\stat{\sym{**}}{s}
& 15.04\stat{\sym{***}}{s}
& -0.67\stat{}{n}
& -6.30\stat{\sym{*}}{s}
& 7.99\stat{\sym{***}}{m}
& 7.01\stat{\sym{***}}{m}
& 4.73\stat{\sym{*}}{s}
& 4.97\stat{\sym{***}}{m}
& 2.78\stat{\sym{***}}{s}
& 2.03\stat{\sym{***}}{s}
& 1.88\stat{\sym{**}}{s}
& 1.37\stat{}{n} \\
\addlinespace[1pt]

InternVL3-8B
& $\Delta_f^{(\hat{Y})}$
& 1.53\stat{}{n}
& 5.00\stat{\sym{***}}{s}
& 4.92\stat{}{n}
& -0.25\stat{}{n}
& -4.18\stat{}{s}
& 4.69\stat{\sym{*}}{s}
& 2.02\stat{\sym{**}}{n}
& -1.05\stat{}{n}
& 2.93\stat{\sym{***}}{s}
& 1.55\stat{}{n}
& 2.06\stat{\sym{***}}{s}
& 2.50\stat{\sym{***}}{s}
& 1.67\stat{\sym{*}}{s} \\
\addlinespace[1pt]

Qwen3-VL-8B-Instruct
& $\Delta_f^{(\hat{Y})}$
& -1.01\stat{}{n}
& 2.62\stat{\sym{**}}{s}
& 11.58\stat{\sym{**}}{s}
& -1.66\stat{}{n}
& -4.57\stat{}{s}
& 4.23\stat{}{s}
& 2.45\stat{\sym{**}}{s}
& -3.68\stat{}{n}
& 2.27\stat{\sym{***}}{s}
& 2.31\stat{}{n}
& 2.36\stat{\sym{***}}{s}
& 1.53\stat{\sym{***}}{n}
& 0.93\stat{}{n} \\
\addlinespace[1pt]

Qwen3-VL-8B-Thinking
& $\Delta_f^{(\hat{Y})}$
& -3.46\stat{\sym{***}}{s}
& 3.22\stat{\sym{**}}{s}
& 7.93\stat{}{n}
& -0.80\stat{}{n}
& -7.30\stat{}{s}
& 7.74\stat{\sym{***}}{s}
& 2.50\stat{\sym{***}}{s}
& -2.41\stat{}{n}
& 2.50\stat{\sym{***}}{s}
& 1.01\stat{}{n}
& 2.08\stat{\sym{***}}{s}
& 1.34\stat{}{n}
& 1.20\stat{}{n} \\

\addlinespace[3pt]

\rowcolor{gray!15!violet!10}
\multicolumn{15}{l}{\textbf{Panel B: VLM False Positives among Ground-Truth Low Items}} \\
\addlinespace[2pt]

Gemma3-4B
& $\Delta_f^{(FP-TN)}$
& -0.63\stat{}{n}
& 3.90\stat{}{s}
& 6.53\stat{}{n}
& 2.42\stat{}{n}
& -8.21\stat{}{s}
& 4.66\stat{}{s}
& 3.91\stat{\sym{***}}{s}
& -0.47\stat{}{n}
& 3.32\stat{\sym{***}}{m}
& -0.82\stat{}{n}
& 1.08\stat{}{n}
& 1.61\stat{\sym{*}}{s}
& 1.06\stat{}{n} \\
\addlinespace[1pt]

Gemma3-12B
& $\Delta_f^{(FP-TN)}$
& -0.40\stat{}{n}
& -1.18\stat{}{n}
& 5.00\stat{}{n}
& -1.83\stat{}{n}
& -0.24\stat{}{n}
& 2.38\stat{}{n}
& 0.34\stat{}{n}
& -1.29\stat{}{n}
& 0.43\stat{}{n}
& -0.64\stat{}{n}
& 2.12\stat{\sym{***}}{s}
& 1.08\stat{}{n}
& 1.66\stat{\sym{***}}{s} \\
\addlinespace[1pt]

Gemma3-27B
& $\Delta_f^{(FP-TN)}$
& -0.93\stat{}{n}
& 2.90\stat{}{s}
& 6.00\stat{}{n}
& -1.90\stat{}{n}
& -6.69\stat{\sym{***}}{s}
& 5.99\stat{\sym{***}}{s}
& 2.23\stat{\sym{*}}{s}
& 1.64\stat{}{n}
& 2.47\stat{\sym{**}}{s}
& 0.85\stat{}{n}
& 1.41\stat{}{n}
& 1.96\stat{\sym{**}}{s}
& 1.63\stat{\sym{***}}{s} \\
\addlinespace[1pt]

InternVL3-8B
& $\Delta_f^{(FP-TN)}$
& 3.77\stat{\sym{***}}{s}
& 4.95\stat{\sym{***}}{s}
& 0.10\stat{}{n}
& -0.99\stat{}{n}
& -4.61\stat{\sym{*}}{s}
& 4.07\stat{}{s}
& 1.02\stat{}{n}
& -0.96\stat{}{n}
& 2.30\stat{\sym{**}}{s}
& -0.16\stat{}{n}
& 1.38\stat{}{n}
& 2.17\stat{\sym{***}}{s}
& 1.61\stat{}{s} \\
\addlinespace[1pt]

Qwen3-VL-8B-Instruct
& $\Delta_f^{(FP-TN)}$
& 2.26\stat{}{n}
& 1.01\stat{}{n}
& 6.78\stat{}{n}
& -2.87\stat{}{n}
& -4.48\stat{}{s}
& 1.70\stat{}{n}
& 1.66\stat{}{n}
& -3.86\stat{}{n}
& 0.97\stat{}{n}
& 0.20\stat{}{n}
& 1.54\stat{\sym{*}}{s}
& 1.29\stat{}{n}
& 1.02\stat{}{n} \\
\addlinespace[1pt]

Qwen3-VL-8B-Thinking
& $\Delta_f^{(FP-TN)}$
& -2.22\stat{}{n}
& 1.62\stat{}{n}
& 1.94\stat{}{n}
& -1.29\stat{}{n}
& -6.21\stat{}{s}
& 5.66\stat{\sym{*}}{s}
& 1.25\stat{}{n}
& -2.31\stat{}{n}
& 1.15\stat{}{n}
& -1.13\stat{}{n}
& 1.32\stat{}{n}
& 1.15\stat{}{n}
& 1.47\stat{}{n} \\

\bottomrule
\end{tabular}
\end{adjustbox}

\caption{
Human--VLM VPF comparison across expanded model variants.
Panel A compares the human ground-truth VPF pattern with predictions from models spanning different scales, families, and reasoning variants.
Panel B analyzes false positives among ground-truth Low items.
Across these settings, VLMs continue to reproduce only part of the human-aligned VPF pattern, while persuasion-relevant cues recur among erroneous High predictions.
}
\label{tab:model_variant_vpf_generalization}
\end{table*}

%% file: latex/sections/CI_original_result.tex
\begin{table*}[t]
\centering
\renewcommand{\arraystretch}{1.08}
\setlength{\tabcolsep}{4.2pt}
\begin{adjustbox}{max width=\linewidth}
\small
\begin{tabular}{llcccc}
\toprule
\textbf{Model} & \textbf{Condition} & \textbf{Acc.} & \textbf{Prec. ($\Delta$)} & \textbf{Rec. ($\Delta$)} & \textbf{F1 ($\Delta$)} \\
\midrule

\rowcolor{gray!15}
\multicolumn{6}{l}{\textbf{Panel A. Baseline}} \\
\addlinespace[2pt]

\raisebox{-0.6ex}{\includegraphics[width=1.3em]{icons/gpt.png}}\ GPT-5
& Baseline & 69.57 & 51.01 {\scriptsize(0.00)} & 98.88 {\scriptsize(0.00)} & 67.30 {\scriptsize(0.00)} \\

\raisebox{-0.6ex}{\includegraphics[width=1.3em]{icons/gpt.png}}\ GPT-5-mini
& Baseline & 60.14 & 44.13 {\scriptsize(0.00)} & 97.19 {\scriptsize(0.00)} & 60.70 {\scriptsize(0.00)} \\

\raisebox{-0.6ex}{\includegraphics[width=1.3em]{icons/gemma-color.png}}\ Gemma3
& Baseline & 63.88 & 46.33 {\scriptsize(0.00)} & 88.76 {\scriptsize(0.00)} & 60.89 {\scriptsize(0.00)} \\

\raisebox{-0.6ex}{\includegraphics[width=1.1em]{icons/qwen-color.png}}\ Qwen2.5-VL
& Baseline & 54.98 & 40.51 {\scriptsize(0.00)} & 89.89 {\scriptsize(0.00)} & 55.85 {\scriptsize(0.00)} \\

\midrule

\rowcolor{blue!10}
\multicolumn{6}{l}{\textbf{Panel B. Factor-wise Cognitive Injection}} \\
\addlinespace[2pt]

\rowcolor{gray!5}
\multicolumn{6}{l}{\textit{GPT-5}} \\
\raisebox{-0.6ex}{\includegraphics[width=1.3em]{icons/gpt.png}}\ GPT-5
& CI(Color) & 65.66 & 47.96 \dDown{3.05} & 98.88 {\scriptsize(0.00)} & 64.59 \dDown{2.71} \\
\raisebox{-0.6ex}{\includegraphics[width=1.3em]{icons/gpt.png}}\ GPT-5
& CI(Brightness) & 67.44 & 49.30 \dDown{1.71} & 99.44 \dUp{0.56} & 65.92 \dDown{1.38} \\
\raisebox{-0.6ex}{\includegraphics[width=1.3em]{icons/gpt.png}}\ GPT-5
& CI(Sal-Center) & 68.51 & 50.14 \dDown{0.87} & 98.88 {\scriptsize(0.00)} & 66.54 \dDown{0.76} \\
\raisebox{-0.6ex}{\includegraphics[width=1.3em]{icons/gpt.png}}\ GPT-5
& CI(Sal-RoT) & 70.11 & 51.46 \dUp{0.45} & 98.88 {\scriptsize(0.00)} & 67.69 \dUp{0.39} \\
\raisebox{-0.6ex}{\includegraphics[width=1.3em]{icons/gpt.png}}\ GPT-5
& CI(Reg-Center) & 68.86 & 50.43 \dDown{0.58} & 98.31 \dDown{0.57} & 66.67 \dDown{0.63} \\
\raisebox{-0.6ex}{\includegraphics[width=1.3em]{icons/gpt.png}}\ GPT-5
& CI(Reg-RoT) & 69.22 & 50.72 \dDown{0.29} & 98.88 {\scriptsize(0.00)} & 67.05 \dDown{0.25} \\
\raisebox{-0.6ex}{\includegraphics[width=1.3em]{icons/gpt.png}}\ GPT-5
& CI(KeyObj) & 70.11 & 51.47 \dUp{0.46} & 98.31 \dDown{0.57} & 67.57 \dUp{0.27} \\
\raisebox{-0.6ex}{\includegraphics[width=1.3em]{icons/gpt.png}}\ GPT-5
& CI(Human) & 72.60 & 53.70 \dUp{2.69} & 97.75 \dDown{1.13} & 69.32 \dUp{2.02} \\
\raisebox{-0.6ex}{\includegraphics[width=1.3em]{icons/gpt.png}}\ GPT-5
& CI(Text) & 65.30 & 47.72 \dDown{3.29} & 100.00 \dUp{1.12} & 64.61 \dDown{2.69} \\

\addlinespace[3pt]

\rowcolor{gray!5}
\multicolumn{6}{l}{\textit{GPT-5-mini}} \\
\raisebox{-0.6ex}{\includegraphics[width=1.3em]{icons/gpt.png}}\ GPT-5-mini
& CI(Color) & 62.10 & 45.36 \dUp{1.23} & 96.07 \dDown{1.12} & 61.62 \dUp{0.92} \\
\raisebox{-0.6ex}{\includegraphics[width=1.3em]{icons/gpt.png}}\ GPT-5-mini
& CI(Brightness) & 64.06 & 46.79 \dUp{2.66} & 98.31 \dUp{1.12} & 63.41 \dUp{2.71} \\
\raisebox{-0.6ex}{\includegraphics[width=1.3em]{icons/gpt.png}}\ GPT-5-mini
& CI(Sal-Center) & 62.10 & 45.33 \dUp{1.20} & 95.51 \dDown{1.68} & 61.48 \dUp{0.78} \\
\raisebox{-0.6ex}{\includegraphics[width=1.3em]{icons/gpt.png}}\ GPT-5-mini
& CI(Sal-RoT) & 63.17 & 46.07 \dUp{1.94} & 95.51 \dDown{1.68} & 62.16 \dUp{1.46} \\
\raisebox{-0.6ex}{\includegraphics[width=1.3em]{icons/gpt.png}}\ GPT-5-mini
& CI(Reg-Center) & 61.03 & 44.70 \dUp{0.57} & 97.19 {\scriptsize(0.00)} & 61.24 \dUp{0.54} \\
\raisebox{-0.6ex}{\includegraphics[width=1.3em]{icons/gpt.png}}\ GPT-5-mini
& CI(Reg-RoT) & 61.39 & 44.91 \dUp{0.78} & 96.63 \dDown{0.56} & 61.32 \dUp{0.62} \\
\raisebox{-0.6ex}{\includegraphics[width=1.3em]{icons/gpt.png}}\ GPT-5-mini
& CI(KeyObj) & 61.03 & 44.70 \dUp{0.57} & 97.19 {\scriptsize(0.00)} & 61.24 \dUp{0.54} \\
\raisebox{-0.6ex}{\includegraphics[width=1.3em]{icons/gpt.png}}\ GPT-5-mini
& CI(Human) & 61.74 & 45.24 \dUp{1.11} & 98.88 \dUp{1.69} & 62.08 \dUp{1.38} \\
\raisebox{-0.6ex}{\includegraphics[width=1.3em]{icons/gpt.png}}\ GPT-5-mini
& CI(Text) & 62.28 & 45.33 \dUp{1.20} & 97.19 {\scriptsize(0.00)} & 62.01 \dUp{1.31} \\

\addlinespace[3pt]

\rowcolor{gray!5}
\multicolumn{6}{l}{\textit{Gemma3}} \\
\raisebox{-0.6ex}{\includegraphics[width=1.3em]{icons/gemma-color.png}}\ Gemma3
& CI(Color) & 70.46 & 52.00 \dUp{5.67} & 87.64 \dDown{1.12} & 65.27 \dUp{4.38} \\
\raisebox{-0.6ex}{\includegraphics[width=1.3em]{icons/gemma-color.png}}\ Gemma3
& CI(Brightness) & 68.15 & 49.84 \dUp{3.51} & 87.64 \dDown{1.12} & 63.54 \dUp{2.65} \\
\raisebox{-0.6ex}{\includegraphics[width=1.3em]{icons/gemma-color.png}}\ Gemma3
& CI(Sal-Center) & 68.86 & 50.50 \dUp{4.17} & 85.96 \dDown{2.80} & 63.62 \dUp{2.73} \\
\raisebox{-0.6ex}{\includegraphics[width=1.3em]{icons/gemma-color.png}}\ Gemma3
& CI(Sal-RoT) & 70.28 & 51.85 \dUp{5.52} & 86.52 \dDown{2.24} & 64.84 \dUp{3.95} \\
\raisebox{-0.6ex}{\includegraphics[width=1.3em]{icons/gemma-color.png}}\ Gemma3
& CI(Reg-Center) & 65.84 & 47.88 \dUp{1.55} & 88.76 {\scriptsize(0.00)} & 62.20 \dUp{1.31} \\
\raisebox{-0.6ex}{\includegraphics[width=1.3em]{icons/gemma-color.png}}\ Gemma3
& CI(Reg-RoT) & 66.01 & 48.07 \dUp{1.74} & 91.01 \dUp{2.25} & 62.91 \dUp{2.02} \\
\raisebox{-0.6ex}{\includegraphics[width=1.3em]{icons/gemma-color.png}}\ Gemma3
& CI(KeyObj) & 67.08 & 48.85 \dUp{2.52} & 83.71 \dDown{5.05} & 61.70 \dUp{0.81} \\
\raisebox{-0.6ex}{\includegraphics[width=1.3em]{icons/gemma-color.png}}\ Gemma3
& CI(Human) & 68.86 & 50.47 \dUp{4.14} & 89.89 \dUp{1.13} & 64.65 \dUp{3.76} \\
\raisebox{-0.6ex}{\includegraphics[width=1.3em]{icons/gemma-color.png}}\ Gemma3
& CI(Text) & 67.79 & 49.54 \dUp{3.21} & 89.89 \dUp{1.13} & 63.87 \dUp{2.98} \\

\addlinespace[3pt]

\rowcolor{gray!5}
\multicolumn{6}{l}{\textit{Qwen2.5-VL}} \\
\raisebox{-0.6ex}{\includegraphics[width=1.1em]{icons/qwen-color.png}}\ Qwen2.5-VL
& CI(Color) & 62.10 & 44.58 \dUp{4.07} & 80.90 \dDown{8.99} & 57.49 \dUp{1.64} \\
\raisebox{-0.6ex}{\includegraphics[width=1.1em]{icons/qwen-color.png}}\ Qwen2.5-VL
& CI(Brightness) & 61.74 & 44.64 \dUp{4.13} & 86.52 \dDown{3.37} & 58.89 \dUp{3.04} \\
\raisebox{-0.6ex}{\includegraphics[width=1.1em]{icons/qwen-color.png}}\ Qwen2.5-VL
& CI(Sal-Center) & 61.57 & 44.41 \dUp{3.90} & 84.83 \dDown{5.06} & 58.30 \dUp{2.45} \\
\raisebox{-0.6ex}{\includegraphics[width=1.1em]{icons/qwen-color.png}}\ Qwen2.5-VL
& CI(Sal-RoT) & 63.17 & 45.62 \dUp{5.11} & 84.83 \dDown{5.06} & 59.33 \dUp{3.48} \\
\raisebox{-0.6ex}{\includegraphics[width=1.1em]{icons/qwen-color.png}}\ Qwen2.5-VL
& CI(Reg-Center) & 65.48 & 47.47 \dUp{6.96} & 84.27 \dDown{5.62} & 60.73 \dUp{4.88} \\
\raisebox{-0.6ex}{\includegraphics[width=1.1em]{icons/qwen-color.png}}\ Qwen2.5-VL
& CI(Reg-RoT) & 63.88 & 46.06 \dUp{5.55} & 82.02 \dDown{7.87} & 58.99 \dUp{3.14} \\
\raisebox{-0.6ex}{\includegraphics[width=1.1em]{icons/qwen-color.png}}\ Qwen2.5-VL
& CI(KeyObj) & 58.90 & 42.82 \dUp{2.31} & 88.76 \dDown{1.13} & 57.77 \dUp{1.92} \\
\raisebox{-0.6ex}{\includegraphics[width=1.1em]{icons/qwen-color.png}}\ Qwen2.5-VL
& CI(Human) & 65.48 & 47.01 \dUp{6.50} & 70.79 \dDown{19.10} & 56.50 \dUp{0.65} \\
\raisebox{-0.6ex}{\includegraphics[width=1.1em]{icons/qwen-color.png}}\ Qwen2.5-VL
& CI(Text) & 65.84 & 47.71 \dUp{7.20} & 82.02 \dDown{7.87} & 60.33 \dUp{4.48} \\

\midrule

\rowcolor{green!10}
\multicolumn{6}{l}{\textbf{Panel C. All-factor Cognitive Injection}} \\
\addlinespace[2pt]

\raisebox{-0.6ex}{\includegraphics[width=1.3em]{icons/gpt.png}}\ GPT-5
& CI(All) & 69.93 & 51.35 \dUp{0.34} & 96.07 \dDown{2.81} & 66.93 \dDown{0.37} \\

\raisebox{-0.6ex}{\includegraphics[width=1.3em]{icons/gpt.png}}\ GPT-5-mini
& CI(All) & 60.50 & 44.33 \dUp{0.20} & 96.63 \dDown{0.56} & 60.78 \dUp{0.08} \\

\raisebox{-0.6ex}{\includegraphics[width=1.3em]{icons/gemma-color.png}}\ Gemma3
& CI(All) & 58.57 & 43.31 \dDown{3.02} & 94.94 \dUp{6.18} & 59.30 \dDown{1.59} \\

\raisebox{-0.6ex}{\includegraphics[width=1.1em]{icons/qwen-color.png}}\ Qwen2.5-VL
& CI(All) & 62.10 & 44.87 \dUp{4.36} & 85.96 \dDown{3.93} & 58.96 \dUp{3.11} \\

\bottomrule
\end{tabular}
\end{adjustbox}
\caption{
Results of factor-wise and all-factor cognitive injection on VLM persuasive judgment performance.
\textit{Notes.} $\Delta$ next to Precision, Recall, and F1 indicates the absolute percentage-point change from each model's own baseline Precision, Recall, and F1 scores, respectively.
}
\label{tab:cognitive_injection_results}
\end{table*}